\documentclass[11pt]{article}
\usepackage[margin=1in]{geometry}
\usepackage[utf8]{inputenc}
\DeclareUnicodeCharacter{03C0}{\ensuremath{\pi}}

\usepackage{mzpreamble}

\newcommand{\polylog}{\textrm{polylog}}

\renewcommand{\S}{\mathcal{S}}
\newcommand{\A}{\mathcal{A}}
\newcommand{\Phat}{\widehat{P}}
\newcommand{\Vhat}{\widehat{V}}
\newcommand{\Qhat}{\widehat{Q}}

\newcommand{\M}{\mathcal{M}}

\newcommand{\Vabsinact}{\Vhat^\star_{-sa}}
\newcommand{\VabsP}{\widetilde{V}^\star_{sa}}
\newcommand{\VabsPhat}{\Vhat^\star_{sa}}

\newcommand{\gbar}{\bar{\gamma}}

\newcommand{\pihat}{\widehat{\pi}}
\newcommand{\pistar}{\pi^\star}

\newcommand{\phihat}{\widehat{\phi}} 

\newcommand{\dbe}{\delta_{\mathrm{BE}}}
\newcommand{\eopt}{\varepsilon_{\mathrm{opt}}}
\newcommand{\epihat}{\varepsilon_{\pihat}}
\newcommand{\epistar}{\varepsilon_{\pistar}}

\newcommand{\Vs}{\overline{V}}
\newcommand{\rs}{\overline{r}}

\DeclareMathOperator{\diam}{diam}
\newcommand{\hitt}{\tau}

\newcommand{\F}{\mathcal{F}}
\newcommand{\Fsa}{\F_{-sa}}
\newcommand{\vmins}{\overline{v}_-}
\newcommand{\Ehit}{E_{\mathrm{hit}}}

\newcommand{\uhat}{\widehat{u}}

\newcommand{\PiSD}{\Pi_{\mathrm{SD}}}
\newcommand{\PiSR}{\Pi_{\mathrm{SR}}}

\newcommand{\emdps}{\widehat{\mathcal{M}}_n}
\newcommand{\distn}{\Delta_n(\S)}
\newcommand{\emps}{\widehat{\mathcal{P}}_n}

\DeclareMathOperator{\Clim}{\mathrm{Clim}}

\newcommand{\lfs}{\ell}

\newcommand{\ps}{p_\star}
\newcommand{\pz}{p_0}
\newcommand{\ts}{\tilde{s}}
\newcommand{\bs}{\bar{s}}
\newcommand{\tr}{\tilde{r}}

\newcommand{\kl}[2]{\mathsf{KL}(#1 \vert \vert #2)}
\newcommand{\klb}{\mathsf{kl}}
\newcommand{\Ber}{\mathrm{Ber}}
\newcommand{\Alg}{\textsc{Alg}}

\newcommand{\tmix}{\tau_{\mathrm{unif}}}

\usepackage{natbib}

\usepackage{algorithm}
\usepackage{algpseudocode}
\usepackage{float}

\usepackage{tikz}

\title{Bellman Residual Bracketing: Resolving the Sample Complexity of Discounted and Average-Reward RL}

\usepackage{authblk}
\author{Matthew Zurek}
\author{Yudong Chen}
\affil{Department of Computer Sciences, University of Wisconsin--Madison\\\vspace{0.8em}
\texttt{matthew.zurek@wisc.edu}\quad\texttt{yudongchen@cs.wisc.edu}}

\date{}

\begin{document}

\maketitle

\begin{abstract}
    We study the sample complexity of the empirical planning or plug-in approach for reinforcement learning (RL) with a generative model. We show the discounted plug-in approach finds an $\varepsilon$-optimal policy with $O(SA (1-\gamma)^{-3} \varepsilon^{-2} \log(SA/\delta))$ samples, where $S, A, \gamma, \delta$ are the state and action sizes, discount factor, and failure probability, respectively. In the average-reward setting, we show that discounted plug-in and direct (undiscounted) plug-in achieve $O(SAH\varepsilon^{-2}\log(SA/\delta))$ and $O(SAD\varepsilon^{-2}\log(SA/\delta))$ complexities, respectively, where $H$ upper-bounds the optimal bias span and $D$ is the diameter. Combining with our lower bounds for nonadaptive generative models, these results resolve the minimax sample complexity of discounted RL and average-reward RL---in both the span- and diameter-based classes---up to constant factors. Our main technical contribution is a novel Bellman residual bracketing technique which improves the efficiency of prior absorbing-MDP-based concentration analyses.
\end{abstract}

\section{Introduction}

Empirical planning---also known as the plug-in approach---is arguably the most basic approach to reinforcement learning with a generative model: estimate the transition kernel from samples and then solve the resulting empirical MDP.
The sample complexity of RL with a generative model is one of the most well-studied problems in the theory of reinforcement learning.
The plug-in approach is one of the most well-studied algorithms for this problem, and in the discounted setting it is known to enjoy optimal polynomial dependence on the problem parameters in a minimax sense. However, sharp bounds with the correct logarithmic factors are only known in 
restricted regimes or not at all for the average-reward plug-in approach.

Our first contribution is a sharp analysis of the discounted plug-in approach, showing that $O(SA(1-\gamma)^{-3} \varepsilon^{-2} \log(SA/\delta))$ samples suffice to learn an $\varepsilon$-optimal policy, and furthermore that all sufficiently near-optimal policies in the empirical MDP are $\varepsilon$-optimal. Combined with our matching lower bound for the nonadaptive sampling setting, this resolves the minimax sample complexity of discounted RL up to constant factors.
This analysis attains sharp logarithmic factors $\log(SA/\delta)$---independent of $\varepsilon$ and $\gamma$---and holds for the entire nontrivial range of $\varepsilon \in (0, \frac{1}{1-\gamma}]$.

These improvements are essential for our second main contribution, which is to resolve the sample complexity of computing an $\varepsilon$-optimal policy in both the span- and diameter-based minimax classes. For the span-based result we show that the discounted plug-in approach with appropriately tuned discount is optimal, and for the diameter-based class we show that the direct average-reward plug-in approach is also optimal.
In both cases the optimality is certified by our sharp $\Omega(SAD \varepsilon^{-2} \log(SA/\delta))$ lower bound, where $D$ is the diameter.
In particular, to the best of our knowledge these are the first results to achieve the optimal $\varepsilon^{-2}$ rate in the average-reward setting, as opposed to additional polylogarithmic dependence on $\varepsilon^{-1}$. This is also the first result to show that the vanilla average-reward plug-in approach, without anchoring or reward perturbation, achieves a clean diameter-based rate.

Our third main contribution is the novel concentration technique behind our analysis that enables sharp $\log(SA/\delta)$ logarithmic terms, which we call \emph{Bellman residual bracketing}. 
The seminal absorbing MDP technique \citep{agarwal_modelbased_2020} constructs a one-dimensional covering of the empirical optimal value function that is independent from one row of the empirical transition kernel, decoupling their probabilistic dependence and enabling the application of concentration inequalities, but with a cover size that grows polynomially in $\frac{1}{\varepsilon}$ or $\frac{1}{1-\gamma}$. We instead use monotonicity of a scalar Bellman residual associated with the absorbing MDPs to control a certain deviation term in terms of its values at two endpoints. 
We also develop a much simpler algebraic error decomposition than previous work.
Technical ideas developed for the analysis of RL with a generative model and in particular the plug-in approach have since been more widely influential, and we hope that this residual bracketing technique can also be broadly useful.

\section{Notation and Problem Setup}
\label{sec:setup}

We let $\N = \{1, 2, \dots\}$ be the set of natural numbers. 
For any finite set $X$ let $\Delta(X) := \{f : X \to [0,1] : \sum_{x \in X}f(x)=1\}$ denote the set of probability distributions over $X$.
For a vector $x$ and square matrix $X$ we define the infinity norm $\infnorm{x} = \max_i |x(i)|$, the span seminorm $\spannorm{x} = \max_{i} x(i) - \min_i x(i)$, and the infinity-to-infinity operator norm $\infinfnorm{X} = \max_i \sum_j |X_{ij}|$. We also let $x^{\circ 2}(i) = (x(i))^2$ be the elementwise square of $x$.
We also let $\zero, \one$ be the all-zero and all-one vectors (whose dimension will be clear from context). We define $e_i $ to be the $i$-th standard basis vector $e_i(j) = \ind(i = j)$.
For any sequence of real numbers $(a_n)_{n \in \N}$, their Cesaro limit is $\Clim_{n \to \infty} a_n := \lim_{k \to \infty} \frac{1}{k}\sum_{i=1}^k a_i$.

\paragraph{MDPs}
A Markov decision process (MDP) is a tuple $(\S, \A, P, r)$ consisting of a state space $\S$, action space $\A$, transition kernel $P : \S \times \A \to \Delta(\S)$, and a reward function $r : \S \times \A \to \R$.
We furthermore assume that $\S, \A$ are finite and that $r(s,a) \in [0,1]$ for all $s\in \S, a \in \A$. Let $S := |\S|$ and $A = |\A|$.
A stationary policy is a function $\pi : \S \to \Delta(\A)$, and we let $\PiSR$ denote the set of all such policies. Furthermore if $\pi(\cdot \mid s)$ is only supported on one action for all $s \in \S$, we say that the policy is a deterministic stationary policy, and we also treat it as a function $\S \to \A$. We let $\PiSD$ denote the set of all stationary deterministic policies.
Given a stationary policy and an MDP $(\S, \A, P, r)$, for a starting state $s_0 \in \S$, the process $S_0 = s_0, A_t \sim \pi(\cdot \mid S_t), S_{t+1} \sim P(\cdot \mid S_t, A_t)$ induces a Markov chain over state space $\S$ with transition matrix $P_\pi \in \R^{\S \times \S}$ where $(P_\pi)_{ss'} = \sum_{a \in \A} \pi(a \mid s) P(s' \mid s,a)$. We let $\E^\pi_{s_0}$ denote the expectation with respect to this distribution. We also define $r_\pi \in \R^\S$ by $r_\pi(s) = \sum_{a \in \A} \pi(a \mid s)r(s,a)$.
We also sometimes treat $P$ as a matrix in $\R^{SA \times S}$ and $r$ as a vector in $\R^{SA}$.
For $p \in \Delta(\S)$ and any vector $V \in \R^S$ we define the variance $\Var_p[V] = p(V^{\circ 2}) - (pV)^2$. We also define the variance vectors $\Var_{P_\pi}[V] \in \R^S, \Var_{P}[V] \in \R^{SA} $ by $\Var_{P_\pi}[V](s) = \Var_{P_\pi (s, \cdot)}[V]$ and $\Var_{P}[V](s,a) = \Var_{P_{sa}}[V]$.

\paragraph{Discounted MDPs}
We assume $\gamma \in (0,1)$ (and note that the problem is trivial when $\gamma = 0$ with known rewards).
For any $\pi \in \PiSR$ the discounted value function in an MDP $\M$ is $V^\pi_{ \M, \gamma} \in \R^S$ where $V^\pi_{ \M, \gamma}(s) = \E^\pi_{s}\sum_{t=0}^\infty \gamma^t r(S_t, A_t)$. We also have $V^\pi_{\M, \gamma} = (I - \gamma P_\pi)^{-1}r_\pi$. The optimal discounted value function is $V^\star_{\M, \gamma} \in \R^S$ where $V^\star_{\M, \gamma}(s) = \sup_{\pi \in \PiSR}V^\pi_{ \M, \gamma}(s)$, and it is well-known there exists $\pi \in \PiSD$ such that $V^\pi_{ \M, \gamma} = V^\star_{ \M, \gamma}$ \citep{puterman_markov_1994}. We also define the optimal Q-function $Q^\star \in \R^{SA}$ by $Q_{\M, \gamma}^\star = r + \gamma P V_{ \M, \gamma}^\star$.
Except for the average-reward result, we treat $\gamma$ as fixed so that we do not need to explicitly add it to the value functions.

\paragraph{Connective Properties}
We assume that $S \geq 2$, since otherwise both policy optimization and optimal value estimation are trivial by choosing the largest-reward action.
For a Markov chain over state space $\S$ with transition matrix $Q$ and starting state $s_0$ we write $\E^Q_{s_0}$ to denote the expectation with respect to its distribution.
For a hitting time in such a chain we write $\hitt_{s} := \inf \{t \geq 0 : S_t = s\}$ (which is $ \infty$ if $s$ is never reached as we take $\inf \emptyset = \infty$).
The diameter of an MDP transition kernel $P$ is defined $\diam(P) := \max_{s,s' \in \S : s \neq s'} \min_{\pi \in \PiSD} \E^{P_\pi}_{s}[\tau_{s'}]$.
We say that an MDP is communicating if $\diam(P) < \infty$. Note that $\diam(P) < \infty$ if and only if for any pair of states $s,s'$, there exists a policy which has a positive probability of eventually reaching $s'$ when being followed starting from state $s$ \citep[Exercise 38.4]{lattimore_bandit_2020}.
The assumption that $S \geq 2$ implies $\diam(P) \geq 1$.
We say that an MDP $(\S, \A, P, r)$ is weakly communicating if it contains a closed sub-MDP which is communicating (that is, there exists some $\S_c \subseteq \S$ such that $P(\cdot \mid s, a)$ is supported on $\S_c$ for all $s  \in \S_c, a \in \A$, and furthermore the restricted MDP $(\S_c, \A, P|_{\S_c \times \A}, r|_{\S_c \times \A})$ to $ \S_c \times \A$ has $\diam(P|_{\S_c \times \A}) < \infty$), as well as a possibly empty set of states which are transient under all stationary policies.
We say a Markov chain is unichain if it has one recurrent class (allowing for some states to be transient), and we say a policy is unichain if its induced chain $P_\pi$ is unichain. In this case we let $\mu^\pi \in \Delta(\S)$ denote its unique stationary distribution.

\paragraph{Average-Reward MDPs}
For any $\pi \in \PiSR$ we define its gain $\rho^\pi_\M \in \R^S$ in MDP $\M$ by $\rho^\pi(s) = \lim_{T \to \infty} \frac{1}{T}\E_s^\pi \sum_{t=0}^{T-1} r(S_t, A_t)$.
The optimal gain $\rho_\M^\star \in \R^\S$ is defined as $\rho^\star(s) = \sup_{\pi \in \PiSR} \rho^\pi(s)$, and there exists $\pi \in \PiSD$ such that $\rho^\pi_{\M} = \rho^\star_{\M}$ \citep{puterman_markov_1994}. 
We also let $P_\pi^\infty \in \R^{S \times S}$ denote the limiting matrix of $P_\pi$, that is, $P_\pi^\infty := \Clim_{T \to \infty} P_\pi^T$, and note that $\rho^\pi = P_\pi^\infty r_\pi$ and $P_\pi^\infty = P_\pi P_\pi^\infty = P_\pi^\infty P_\pi = P_\pi^\infty = P_\pi^\infty P_\pi^\infty$ \citep[Appendix A]{puterman_markov_1994}.
We define $\pi \in \PiSR$ to be bias-optimal if $\lim_{\gamma \uparrow 1}\infnorm{V_{\M, \gamma}^\star - V_{\M, \gamma}^\pi} = 0$, and we define the optimal bias function $h^\star_{\M} \in \R^{S}$ as $h_{\M}^\star = \lim_{\gamma \uparrow 1} \left( V_{\M, \gamma}^\star - \frac{\rho_{\M}^\star}{1-\gamma} \right)$ (which always exists; see \citealt{puterman_markov_1994}).

\subsection{Sampling Model}
\label{sec:sampling_model}
Throughout all of our sample complexity upper bounds, we assume a common sampling model which we now describe. Let $\M := (\S, \A, P, r)$ be the true MDP, and we assume $r$ is known and $P$ is unknown.
We suppose that, for each $s \in \S, a \in \A$, for $n \in \N$, we observe $n$ samples drawn independently from $P(\cdot \mid s, a)$ and also independently from other rows, that is,
\begin{equation*}
    S^{(1)}_{sa}, S^{(2)}_{sa},  \dots, S^{(n)}_{sa} \stackrel{\text{i.i.d.}}{\sim} P(\cdot \mid s, a).
\end{equation*}
In particular, all failure probabilities within our main results are with respect to the probability space induced by this model. 
We define the empirical transition kernel $\Phat$ by
\begin{equation*}
    \Phat(s' \mid s, a) := \frac{1}{n} \sum_{i=1}^n \ind(S^{(i)}_{sa} = s')
\end{equation*}
where $\ind$ is the indicator function. We refer to $\widehat{\M} := (\S, \A, \Phat, r)$ as the empirical MDP.
We also define simplified notation for the true and empirical MDPs, where for any $\pi \in \PiSR$, we define $V^\pi := V^\pi_{\M, \gamma}, V^\star := V^\star_{\M, \gamma}, Q^\star := Q^\star_{\M, \gamma}, \Vhat^\pi := V^\pi_{\widehat{\M}, \gamma}, \Vhat^\star := V^\star_{\widehat{\M}, \gamma}, \Qhat^\star := Q^\star_{\widehat{\M}, \gamma}$, and also for the average-reward results $\rho^\pi := \rho_{\M}^\pi, \rho^\star := \rho_{\M}^\star , h^\star := h_{\M}^\star, \widehat{\rho}^\pi := \rho^\pi_{\widehat{\M}}, \widehat{\rho}^\star :=  \rho^\star_{\widehat{\M}}$.

A generative model is an oracle which, given any $s \in \S, a \in \A$, returns a fresh independent sample $S' \sim P(\cdot \mid s,a)$ \citep{kearns_finitesample_1998}. The sampling model that we consider can then be simulated by drawing $n$ samples from each $s,a$. This is a non-adaptive or oblivious use of the generative model, since the number of samples to be drawn from each $s,a$ is a fixed function of only $s,a$ and possibly external randomness.
This contrasts the adaptive sampling model where each generative model query may depend on the outcome of all previous queries.

\section{Related Work}

\begin{table}
{
\renewcommand{\arraystretch}{1.4} 
\centering
\begin{tabular}{ccccc}
\toprule
Algorithm & Complexity & $\varepsilon$ Range & Reference & Notes \\ \midrule
Plug-in & $\frac{SA H_\gamma^3\log(SA/\delta)}{\varepsilon^2}$ & $(0, 1]$ & \cite{azar_sample_2012} & $Q^\star$ est.  \\
Plug-in & $\frac{SA H_\gamma^3\log(SA/\delta)}{\varepsilon^2}$ & $(0, \sqrt{\frac{H_\gamma}{S}}]$ & \cite{azar_minimax_2013} &  \\
VRQVI & $\frac{SAH_\gamma^3 \log(SAH_\gamma/\delta \varepsilon) \log(H_\gamma/\varepsilon)}{\varepsilon^2}$ & $(0, 1]$ & \cite{sidford_nearoptimal_2018} \\
Plug-in & $\frac{SA H_\gamma^3\log(SAH_\gamma/\delta)}{\varepsilon^2}$ & $(0, \sqrt{H_\gamma}]$ & \cite{agarwal_modelbased_2020} &  \\
Perturbed plug-in \hspace{-1em} & $\frac{SA H_\gamma^3\log(SAH_\gamma/\varepsilon\delta)}{\varepsilon^2}$ & $(0, H_\gamma]$ & \cite{li_breaking_2020} &  \\
Plug-in & $\frac{SA H_\gamma^3\log^3(SAH_\gamma/\varepsilon\delta)}{\varepsilon^2}$ & $(0, H_\gamma]$ & \cite{zurek_plugin_2025} &  \\
\textbf{Plug-in} & $\frac{SA H_\gamma^3\log(SA/\delta)}{\varepsilon^2}$ & $(0, H_\gamma]$ & \textbf{Theorem \ref{thm:main_theorem_with_horizon}} &  \\ \midrule
Lower bound & $\frac{SAH_\gamma^2\log(1/\delta)}{\varepsilon^2}$ & & \cite{even-dar_action_2006} & Adaptive \\
Lower bound & $\frac{SA H_\gamma^3\log(1/\delta)}{\varepsilon^2}$ & & \cite{feng_how_2020} & Adaptive \\
\textbf{Lower bound} & $\frac{SA H_\gamma^3\log(SA/\delta)}{\varepsilon^2}$ & & \textbf{Theorem \ref{thm:lower_bound_discounted}} &  \\ 
\bottomrule
\end{tabular}
\caption{\textbf{Upper and lower bounds on the sample complexity of solving discounted MDPs} to suboptimality $\varepsilon$. Here $H_\gamma := \frac{1}{1-\gamma}$ and $\delta$ is the failure probability. All complexity bounds ignore constant factors.
All algorithms can be implemented under the oblivious sampling model, while the lower bounds of \cite{even-dar_action_2006} and \cite{feng_how_2020} hold against the larger class of adaptive sampling algorithms.
}
\label{table:DMDPs}
}
\end{table}

\paragraph{Discounted MDPs}
The study of the sample complexity of computing an $\varepsilon$-optimal discounted policy from a generative model has received extensive attention. Table \ref{table:DMDPs} summarizes some more recent works which are optimal up to logarithmic factors in some regime, but these are a small subset of the influential works on this problem
(e.g., \citealt{ghavamzadeh_speedy_2011, wang_primaldual_2017, wainwright_variancereduced_2019, sidford_variance_2023, jin_truncated_2024}).
\citet{azar_sample_2012} bounds the complexity of estimating the optimal Q-function $Q^\star$ to $\infnorm{\cdot}$ error $\varepsilon$, rather than computing an $\varepsilon$-optimal policy. 
\citet{li_breaking_2024} are the first to obtain an optimal (up to logs) sample complexity bound for the full range of suboptimality $\varepsilon \in (0, H_\gamma]$, and their algorithms are closely based upon the plug-in approach but either require the reward to be randomly perturbed or a certain conservative planning oracle. \citet{zurek_plugin_2025} achieve this for the unmodified plug-in approach without reward perturbation.

\paragraph{Discounted lower bounds}
\citet{azar_minimax_2013} claim a lower bound of $\Omega(SAH_\gamma^3 \varepsilon^{-2} \log(SA/\delta))$ for $Q^\star$ estimation under adaptive sampling, but due to proof gaps that we discuss in Appendix \ref{sec:discounted_lower_bounds}, we provide Theorem \ref{thm:lower_bound_discounted}, which covers policy optimization under nonadaptive sampling.
\citet{sidford_nearoptimal_2018} argues that a $Q^\star$-estimation lower bound could also imply a lower bound for policy optimization through a reduction using a policy evaluation oracle (that is, an algorithm for policy optimization combined with an evaluation oracle yields an algorithm for $Q^\star$ estimation), but there are several caveats:
This reduction requires adaptive generative model use.
Also, their policy evaluation subroutine utilizes $\widetilde{O}(SA H_\gamma^3 \varepsilon^{-2})$ samples, and even our new sharper evaluation complexity does not fully match adaptive complexity lower bounds for policy optimization \citep{feng_how_2020}.
Therefore this reduction yields lower bounds for some regimes of $A$ but does not provide a sharp-logarithm lower bound nor apply to the nonadaptive setting. 
\cite{strehl_reinforcement_2009, lattimore_pac_2012} prove lower bounds for a related but different setting of sample complexity of exploration.

\begin{table}

\renewcommand{\arraystretch}{1.4} 
\centering
\begin{tabular}{ccccc}
\toprule
Algorithm & Complexity  & Reference & Notes \\ \midrule
Discounted reduction & $\frac{SA H\log(SA/\delta \varepsilon)}{\varepsilon^3}$ & \cite{wang_sampleoptimal_2022} & \\
Monotone Q-learning & $\big(\frac{SAH^2}{\varepsilon^2}\! +\! \frac{S^2 A H }{\varepsilon}\big){\scriptstyle\log(SAH/\varepsilon\delta) \log(1/\varepsilon)}$ & \cite{zhang_sharper_2023} \\
Discounted reduction & $\frac{SA H\log(SAH/\delta \varepsilon)}{\varepsilon^2}$ & \cite{zurek_spanbased_2024} & \\
Anchored plug-in & $\frac{SA D\log^3(SAD/\delta \varepsilon)}{\varepsilon^2}$ & \cite{zurek_plugin_2025} & \\
SAVIA & $\frac{SAH^2 \log(SAH/\delta \varepsilon) \log^3(H/\varepsilon)}{\varepsilon^2} $ & \cite{lee_nearoptimal_2025} & \\
\textbf{Discounted reduction}\!\!\! & $\frac{SA H\log(SA/\delta)}{\varepsilon^2}$ & \textbf{Theorem \ref{thm:main_average_reward_reduction}} & \\
\textbf{Plug-in} & $\frac{SA D\log(SA/\delta)}{\varepsilon^2}$ & \textbf{Theorem \ref{thm:main_average_reward_plug_in}} & \\\midrule
Lower bound & $\frac{SA H}{\varepsilon^2}$ & \cite{jin_tight_2021} &  \\
Lower bound & $\frac{SAD\log(1/\delta)}{\varepsilon^2}$ & \cite{wang_sampleoptimal_2022} & Adaptive\\
\textbf{Lower bound} & $\frac{SA D \log(SA/\delta)}{\varepsilon^2}$ & \textbf{Theorem \ref{thm:avg_rew_lower_bound}} &  \\
\bottomrule
\end{tabular}
\caption{\textbf{Upper and lower bounds on the sample complexity of solving average-reward MDPs} to suboptimality $\varepsilon$. Here $H = \spannorm{h^\star}+1$ and $D$ is the diameter. All algorithms can be implemented under the oblivious sampling model. We always have $\spannorm{h^\star} \leq D$. All complexity bounds ignore constant factors.
}
\label{table:AMDPs}
\end{table}

\paragraph{Average-Reward MDPs}
The sample complexity of average-reward MDPs under a generative model has also been extensively studied.
Table \ref{table:AMDPs} again provides a subset of works that use similar algorithmic approaches or also study the span- or diameter-based minimax sample complexity. In particular, for all discounted reduction rows of Table \ref{table:AMDPs}, the discounted algorithm is a plug-in approach variant contained in Table \ref{table:DMDPs}.
We refer to \cite{jin_efficiently_2020, neu_dealing_2024, bravo_stochastic_2026} for alternative algorithms, and
we also refer to the table within \cite{lee_nearoptimal_2025} for a more exhaustive list.
We also omit the well-studied uniformly mixing setting. Within this setting,
notably \citet{jin_tight_2021} are the first to study the discounted reduction approach, and \citet{wang_optimal_2024} give a minimax-optimal sample complexity bound up to additional logarithmic factors---their bound contains $\log(SA \tmix \varepsilon^{-1} \delta^{-1})$, where $\tmix$ is a uniform mixing time parameter.
Our Theorem \ref{thm:main_average_reward_reduction} gives an improvement for the uniformly mixing setting since always $\spannorm{h^\star} \leq O(\tmix)$ \citep{wang_sampleoptimal_2022, zurek_plugin_2025}.
The lower bound of \citet{jin_tight_2021} actually features the uniform mixing time, but their bound implies the stated $H$-based bound.
In general $\spannorm{h^\star} \leq D$ \citep{bartlett_regal_2012, lattimore_bandit_2020}, so upper bounds in terms of $H$ are stronger, and lower bounds in terms of $D$ are stronger.
While we focus on characterizing the minimax sample complexity, another significant area of research is what rates are achievable without any prior knowledge of complexity parameters. The works \citet{tuynman_finding_2024, zurek_spanagnostic_2025a} respectively show that, up to log factors, the optimal diameter- and span-based rates are achievable without prior knowledge, and we refer to them for more details on this direction. Our Theorem \ref{thm:main_average_reward_plug_in} achieves the optimal diameter-based rate, without additional log factors or sample size restrictions, and without prior knowledge of $D$.
Closely related to our Theorem \ref{thm:main_average_reward_plug_in}, \citet{zurek_plugin_2025} give an analysis of the average-reward plug-in approach, but the bound depends on a certain empirical discount factor threshold which is not bounded in terms of the span or diameter parameters. They also obtain a prior-knowledge-free diameter-based rate for an average-reward plug-in approach based on solving an anchored version of the empirical MDP, rather than the true empirical MDP.
The lower bound result of \citet{wang_sampleoptimal_2022} holds for learning algorithms which output deterministic stationary policies.

\section{Main Results}

In this section we present our main results for discounted and average-reward MDPs.

\subsection{Discounted MDPs}
\begin{theorem}
    \label{thm:main_theorem_with_horizon}
    Fix $\delta \in (0,1)$.
    With probability at least $1 - \delta$,
    for all stationary randomized policies $\pihat \in \PiSR$ satisfying
    \begin{equation}
        \infnorm{\Vhat^\star - \Vhat^{\pihat}} \leq \frac{1}{(1-\gamma)n}, \label{eq:main_theorem_with_horizon_subopt_cond}
    \end{equation}
    we have
    \begin{equation}
        V^{\pihat} \geq V^\star - \eta \one \qquad \text{and}\qquad \infnorm{\Qhat^\star - Q^\star} \leq \eta \label{eq:main_theorem_with_horizon_conclusions}
    \end{equation}
    where
    \begin{equation*}
        \eta = 37 \sqrt{\frac{\log(16SA/\delta)}{(1-\gamma)^3n}}.
    \end{equation*}
\end{theorem}
Equivalently, Theorem \ref{thm:main_theorem_with_horizon} shows that $SA\left\lceil\frac{37^2 \log(16SA/\delta)}{(1-\gamma)^3\varepsilon^2}\right\rceil$ samples suffice to learn an $\varepsilon$-optimal policy and an $\varepsilon$-error estimate of $Q^\star$, for any $\varepsilon \in (0, \frac{1}{1-\gamma}]$.
Theorem \ref{thm:main_theorem_with_horizon} applies simultaneously to all nearly (discounted) optimal policies for the empirical MDP $(\S, \A, \Phat, r)$ satisfying condition~\eqref{eq:main_theorem_with_horizon_subopt_cond}.
Both the exact value of $\Qhat^\star$ and an exactly optimal policy for the empirical MDP can be computed by $O(\frac{SA \log(1/(1-\gamma))}{1-\gamma})$ steps of policy iteration \citep{scherrer_improved_2016}; however, we emphasize that Theorem \ref{thm:main_theorem_with_horizon} is algorithm-agnostic, and also applies directly to the empirical MDP rather than requiring reward perturbation \citep{li_breaking_2024}.

This exactly matches the lower bound Theorem \ref{thm:lower_bound_discounted} for the nonadaptive sampling setting up to constant factors for the entire range of $\varepsilon \in (0, \frac{1}{1-\gamma}]$. 
As shown in Table \ref{table:DMDPs}, comparing to previous analyses of the plug-in approach for this same task, this result is the first to simultaneously hold for the entire range $(0, \frac{1}{1-\gamma}]$ and achieve the optimal logarithmic term $\log(SA/\delta)$, in particular attaining the optimal $O(\varepsilon^{-2})$ complexity when $S, A, \frac{1}{1-\gamma}, \delta$ are fixed, rather than $O(\polylog(\varepsilon^{-1}) \varepsilon^{-2})$.
The fact that the guarantee holds for the entire range of $\varepsilon$ and simultaneously the logarithmic factors do not contain $\frac{1}{1-\gamma}$ is of particular utility for the average-reward setting discussed in the next subsection, where standard discounted reduction techniques \citep{jin_tight_2021,wang_sampleoptimal_2022} choose $\frac{1}{1-\gamma}$ to grow with $\varepsilon^{-1}$.

\subsection{Average-Reward MDPs}
\begin{theorem}
\label{thm:main_average_reward_reduction}
    Suppose that $P$ is weakly communicating.
    Fix $\delta \in (0,1)$. Suppose that $H > 0, \gbar \in (0,1)$ satisfy that $\spannorm{h^\star} \leq H$ and
    \begin{equation*}
        \frac{1}{1-\gbar} = \sqrt{\frac{n H}{\log(8SA/\delta)}}.
    \end{equation*}
    Then with probability at least $1 - \delta$, for all stationary randomized policies $\pihat \in \PiSR$ satisfying
    \begin{equation*}
        \infnorm{\Vhat_{\gbar}^\star - \Vhat_{\gbar}^{\pihat}} \leq \frac{1}{(1-\gbar)n} = \sqrt{\frac{H}{n \log(8SA/\delta)}},
    \end{equation*}
    we have
    \begin{equation*}
        \rho^{\pihat} \geq \rho^\star - \eta \one \qquad \text{and}\qquad\infnorm{(1-\gbar)\Vhat_{\gbar}^\star - \rho^\star} \leq \eta,
    \end{equation*}
    where
    \begin{equation*}
        \eta = 802 \sqrt{\frac{H\log(8SA/\delta)}{n}}.
    \end{equation*}
\end{theorem}

This exactly matches the lower bound Theorem \ref{thm:avg_rew_lower_bound} shown in the next subsection, up to constant factors, thus resolving the minimax sample complexity over the class of MDPs with span bounded by $H$.
Notably this result in particular is the first to attain the optimal $n^{-1/2}$ rate. Previous discounted reduction results that are based upon a discounted sample complexity that contains logarithmic factors of the effective horizon $\frac{1}{1-\gamma}$ instead have $\sqrt{\log(n) / n}$ dependence on $n$, since the reduction requires a horizon growing at least $\sqrt{n}$ \citep{jin_tight_2021, wang_sampleoptimal_2022}. (Such a reduction horizon that grows with $n$ is also the reason that the earlier results of \cite{azar_minimax_2013} cannot be used.) Instead, our Theorem \ref{thm:main_average_reward_reduction} is built upon Theorem \ref{thm:main_first_span_based_bd}, which is a refinement of the discounted Theorem \ref{thm:main_theorem_with_horizon} that replaces one effective horizon factor with the optimal value function span while attaining sharp $\log(SA/\delta)$ logarithmic terms, refining the results of \citet{zurek_spanbased_2024}.

The choice of $\gbar$ used in Theorem \ref{thm:main_average_reward_reduction} requires prior knowledge of $\spannorm{h^\star}$. While this has been a common assumption (e.g., \cite{zhang_regret_2019, wang_sampleoptimal_2022, wang_optimal_2024, zurek_spanbased_2024}), recent work has shown that it is removable \citep{boone_achieving_2024, zurek_spanagnostic_2025a}, albeit sometimes with a price \citep{zamir_optimal_2026}.
While the superior guarantees of Theorems \ref{thm:main_theorem_with_horizon} or \ref{thm:main_average_reward_reduction} could be utilized within the algorithms of \cite{zurek_spanagnostic_2025a} to obtain algorithms agnostic to $\spannorm{h^\star}$ that almost match the sample complexity of Theorem \ref{thm:main_average_reward_reduction}, their span calibration algorithm uses a statistical test to choose among $O(\log n)$ different discount factor choices; a union bound would thus lead to an additional $\log \log n$ term and it is unclear whether this is improvable.
We did not attempt to optimize constants in the analysis, and adjusting the choice of $\frac{1}{1-\gbar}$ by a constant factor may also yield an improved constant.

Next we present a theorem for the (average-reward) plug-in approach, which directly computes a bias optimal policy for $\Phat$ without discounting, and thus does not require such prior knowledge.
\begin{theorem}
\label{thm:main_average_reward_plug_in}
    Suppose $P$ is communicating and $\diam(P) \leq D$. Fix $\delta \in (0,1)$.
    Then with probability at least $1 - \delta$, for all deterministic policies $\pihat \in \PiSD$ that are bias-optimal for $\Phat$,
    we have
    \begin{equation*}
        \rho^{\pihat} \geq \rho^\star - \eta \one \qquad \text{and}\qquad\infnorm{\widehat{\rho}^{\star} - \rho^\star} \leq \eta,
    \end{equation*}
    where
    \begin{equation*}
        \eta = 54 \sqrt{\frac{D\log(24 S^3 A/\delta)}{n}}.
    \end{equation*}
\end{theorem}
By the lower-bound Theorem \ref{thm:avg_rew_lower_bound}, this shows the plug-in approach is minimax-optimal over the class of MDPs with diameter bounded by $D$ up to constant factors, assuming that the generative model is used obliviously.
The previous analysis of the plug-in approach incurs logarithmic dependencies on $n$, again related to the fact that previous analyses of the discounted plug-in approach incurred logarithmic dependencies on $\frac{1}{1-\gamma}$ and this quantity generally needs to grow with $n$ to approximate gain optimality.
Compared to Theorem \ref{thm:main_average_reward_reduction}, the direct average-reward plug-in approach does not require knowledge of $\spannorm{h^\star}$, but Theorem \ref{thm:main_average_reward_plug_in} attains a worse rate since generally $\spannorm{h^\star} \leq D$ \citep{bartlett_regal_2012, lattimore_bandit_2020}.
By the example instance shown in \cite{zurek_plugin_2025}, the dependence on $D$ cannot generally be replaced with $\spannorm{h^\star}$ for the plug-in approach, at least not for the entire regime where $n \ll D$.

We remark that Theorem \ref{thm:main_average_reward_plug_in} can be trivially extended to the larger class of weakly communicating MDPs: since such MDPs contain a communicating sub-MDP for which the theorem can be applied, the quantity replacing $D$ will be the diameter of this communicating sub-MDP \citep{fruit_optimal_2019}. Policy optimization outside this sub-MDP is trivial since all such states must be transient under all policies. See Theorem \ref{thm:average_reward_plug_in_weakly_communicating}.

\subsection{Sample Complexity Lower Bounds}
Finally we show matching (up to constants) sample complexity lower bounds for both discounted and average-reward MDPs. In this subsection we consider a slightly more general sampling model than the one described in subsection \ref{sec:sampling_model}, where we allow the number of samples for each state-action pair to vary. We let $n : \S \times \A \to \N$ be a sample size function and suppose that, for each $s \in \S, a \in \A$, we observe $n(s,a)$ samples drawn from $P(\cdot \mid s, a)$ independently (from each other and other rows), that is,
\begin{equation*}
    S^{(1)}_{sa}, S^{(2)}_{sa},  \dots, S^{(n(s,a))}_{sa} \stackrel{\text{i.i.d.}}{\sim} P(\cdot \mid s, a).
\end{equation*}
We use $\P_\M$ (where $\M = (\S, \A, P, r)$) to denote the probability measure induced by this process.
We also define $\Alg$ to be a function from realizations of the $(S^{(i)}_{sa})_{1 \leq i \leq n(sa), s\in \S, a\in \A}$ to the set of stationary randomized policies $\PiSR$.
First we present our lower bound for the average-reward setting.

\begin{theorem}
    \label{thm:avg_rew_lower_bound}
    Let $S, A$ be even positive natural numbers such that $A \geq 4, S \geq 2$. Also let $\varepsilon \in (0,1/4]$ and $D$ be a natural number such that $ D \geq \log_{A/2}(S/2)$. Then there exists a family of $m := SA/4$ MDPs $\M_1, \dots, \M_m = (\S, \A, P^{(1)}, r), \dots,  (\S, \A, P^{(m)}, r)$ such that
    \begin{enumerate}
        \item $|\S| = S, |\A| = A,$ and for all $i = 1, \dots, m$, $\diam(P^{(i)}) \leq 5D$.
        \item If there exists $\delta \in (0, \frac{1}{4}]$ such that
        \begin{equation*}
            \text{for all }i \in \{1, \dots, m\}, \quad\P_{\M_i} \left( \rho^{\Alg}_{\M_i} \geq \rho^\star_{\M_i} - \frac{\varepsilon}{32} \one \right) \geq 1 - \delta,
        \end{equation*}
        then
        \begin{equation*}
            \sum_{s \in \S, a \in \A} n(s,a) \geq  SA \frac{1}{8} \frac{D}{\varepsilon^2}\log\left( \frac{SA}{8\delta} \right).
        \end{equation*}
    \end{enumerate}
\end{theorem}
Since the diameter always upper-bounds the optimal bias span, that is $\spannorm{h^\star_{\M_i}} \leq \diam (P^{(i)})$ \citep{bartlett_regal_2012, lattimore_bandit_2020}, Theorem \ref{thm:avg_rew_lower_bound} also implies sample complexity lower bounds over classes of MDPs with bounded spans.
(The instances $(\M_i)_i$ themselves can also easily be seen to have optimal bias function spans on the order of $D$; see subsection \ref{sec:lb_construction}.)
Theorem \ref{thm:avg_rew_lower_bound} is the first to capture the logarithmic dependence on $SA$, and, together with the matching (up to constants) upper bound Theorem \ref{thm:main_average_reward_reduction}, implies that the minimax sample complexity of computing an $\varepsilon$-optimal policy is $\Theta(SA\spannorm{h^\star} \varepsilon^{-2} \log(SA/\delta))$ when $\spannorm{h^\star} \geq 1$.

The proof of Theorem \ref{thm:avg_rew_lower_bound} is mostly standard and uses similar constructions to those of the lower bound of \cite{wang_sampleoptimal_2022}, but features sharper information-theoretical arguments.
The theorem notably holds even when $\Alg$ may output randomized policies, whereas prior arguments \citep{wang_sampleoptimal_2022} only apply to algorithms outputting deterministic stationary policies. This requires a more nontrivial decoder construction (that is, a mapping from output policy to MDP parameter within the information-theoretical testing lower bound); see subsection \ref{sec:lb_decoder}. We also note that the condition $ D \geq \log_{A/2}(S/2)$ is used to ensure that the MDPs are communicating; \citet{lattimore_bandit_2020} show that always $\diam(P) \geq \log_A(S) - 2$, so a condition of this form is necessary.

Next we present our lower bound for the discounted setting. Whereas the previous hard instance family depends on $D$ and have diameter $O(D)$, the next family depends on the discount factor $\gamma$.

\begin{theorem}
\label{thm:lower_bound_discounted}
    Let $S,A$ be positive natural numbers such that $S \geq 2, A \geq 3$. Also let $\gamma \in [\frac{1}{2}, 1)$ and $\varepsilon \in (0, \frac{1}{2}(1-\gamma)^{-1}]$. Then there exists a family of $m := (S-1)(A-1)$ MDPs $\M_1, \dots, \M_m = (\S, \A, P^{(1)}, r), \dots, (\S, \A, P^{(m)}, r)$ with the following properties:
    \begin{enumerate}
        \item $|\S| = S$ and $|\A| = A$.
        \item If there exists $\delta \in (0, \frac{1}{4}]$ such that for all $i \in \{1, \dots, m\}$ we have
        \begin{equation*}
            \P_{\M_i} \left( V_{\M_i}^{\Alg} \geq V^\star_{\M_i} - \frac{\varepsilon}{27} \one \right) \geq 1 - \delta,
        \end{equation*}
        then
        \begin{equation*}
            \sum_{s \in \S , a \in \A} n(s,a)  \geq  \frac{SA}{96} \frac{1}{(1-\gamma)^3 \varepsilon^2} \log\left(\frac{SA}{8\delta}\right).
        \end{equation*}
    \end{enumerate}
\end{theorem}
To the best of our knowledge Theorem \ref{thm:lower_bound_discounted} is the first to capture the logarithmic dependence on $SA$. (See Appendix \ref{sec:discounted_lower_bounds} for discussion of why the result of \citealt{azar_minimax_2013} cannot be used.)
This sample complexity matches Theorem \ref{thm:main_theorem_with_horizon} up to constant factors, establishing a minimax sample complexity of $\Theta( SA (1-\gamma)^{-3} \varepsilon^{-2} \log(SA/\delta))$.
The proof is relatively simpler than the average-reward case and builds upon ideas of \citet{azar_minimax_2013}.

\section{Proof Ideas}
In this section we highlight the ideas behind the proofs of our upper bound results, which we believe are of substantial interest.
We mainly focus on the proof of Theorem \ref{thm:main_theorem_with_horizon}, and then sketch the improvements needed for the subsequent theorems.

\subsection{Error Decomposition}
The most significant new idea is the novel leave-one-out probabilistic argument which is discussed in subsection \ref{sec:pf_sketch_loo}, but first we briefly mention some algebraic improvements.
The key finding is that the
conclusions~\eqref{eq:main_theorem_with_horizon_conclusions} of Theorem \ref{thm:main_theorem_with_horizon}, namely bounds on both the suboptimality of a near-empirically-optimal policy $\pihat$ and on the $Q^\star$ estimation error $\infnorm{\Qhat^\star - Q^\star}$, ultimately follow from just the two ``Bernstein-like'' inequalities
\begin{align}
    \left|(\Phat_{\pistar} - P_{\pistar})V^{\pistar} \right| &\leq \sqrt{\frac{\Var_{\Phat_{\pistar} }[V^{\pistar}] \lfs}{n}} + \frac{\lfs}{(1-\gamma)n}\one, \label{eq:pf_sketch_true_bernstein}\\
    \left|(\Phat_{\pihat} - P_{\pihat})\Vhat^\star \right| &\leq \sqrt{\frac{\Var_{P_{\pihat} }[\Vhat^\star] \lfs}{n}} + \frac{\lfs}{(1-\gamma)n}\one\label{eq:pf_sketch_emp_bernstein}
\end{align}
where $\lfs = C\log(cSA/\delta)$ for some constants $C, c \geq 1$.
Note these are elementwise inequalities.

This is in significant contrast to the arguments of \citet{li_breaking_2024} and \citet{zurek_plugin_2025}, the only previous analyses that are able to handle the entire range of error parameter $\varepsilon$ up to $\frac{1}{1-\gamma}$, which both employ multi-step decompositions of the true/empirical errors $\tinfnorm{\Vhat^\pi - V^\pi}$ that require concentration of many auxiliary quantities. Since their arguments are technical and ultimately not necessary to understand our approach, we refer to those papers for details.

Our own arguments, which again instead reduce Theorem \ref{thm:main_theorem_with_horizon} to only the two concentration inequalities~\eqref{eq:pf_sketch_true_bernstein} and~\eqref{eq:pf_sketch_emp_bernstein}, are relatively shorter and simpler and can be found in their entirety in Section \ref{sec:error_decomposition}, with the main steps in Lemmas \ref{lem:error_to_var_param} and \ref{lem:var_lemma}. While they are less novel and intricate than the concentration techniques explained in the next subsection, to the best of our knowledge our results would not be possible without them.
We therefore hope that they may find broader use in the analysis of model-based RL because of their simplicity and efficiency.

We briefly describe one high-level key idea. Previous nearly-minimax sample complexity analyses use a certain ``Bellman total-variance'' technique that bounds certain intrinsic variance parameters in a sharp way by utilizing the Bellman equation \citep{sobel_variance_1982, azar_minimax_2013}. However, this bound directly applies to ``matched'' weighted variance parameters of the form $\infnorm{(I - \gamma P_\pi)^{-1} \Var_{P_\pi}[V^\pi]}$ for some policy $\pi$, whereas (due to the asymmetry within the resolvent identity $\Vhat^\pi - V^\pi = \gamma (I - \gamma P_\pi)^{-1} (\Phat_\pi - P_\pi) \Vhat^\pi$) the variance quantity appearing more naturally within previous analyses is a ``mismatched'' quantity of the form $\infnorm{(I - \gamma P_\pi)^{-1} \Var_{P_\pi}[\Vhat^\pi]}$, involving both true and empirical MDP quantities. Our Lemma \ref{lem:var_lemma} gives a direct bound on this mismatched quantity using only one Bernstein-like inequality as input, rather than relating it to the matched parameter (e.g., \citealt{agarwal_modelbased_2020}) or performing a multiple-step decomposition (e.g., \citealt{li_breaking_2024, zurek_plugin_2025}).

\subsection{Absorbing MDPs}
\label{sec:pf_sketch_loo}

Now we discuss how to establish the two key Bernstein-like bounds~\eqref{eq:pf_sketch_true_bernstein} and~\eqref{eq:pf_sketch_emp_bernstein}. The first follows directly from an empirical Bernstein inequality~\citep{maurer_empirical_2009}. The second is much more involved due to the dependence between $\Phat$ and $\Vhat^\star$. \citet{agarwal_modelbased_2020} introduce absorbing MDPs, an instance of the general leave-one-out technique, to handle this challenge. Since our approach builds upon this idea, we first describe it in more detail.

For a single fixed transition row $(s,a)$, while we cannot directly apply Bernstein's inequality to the inner product $(\Phat_{sa} - P_{sa}) \Vhat^\star $, instead it would suffice to find some $\Vhat^{(s,u)}$\footnote{In principle the construction could also depend on $a$.} which is independent of the samples that form $\Phat_{sa}$ (but can depend on all the other rows of $\Phat$) such that $\Vhat^{(s,u)} \approx \Vhat^\star$, because then
\begin{equation}
    ( \Phat_{sa} - P_{sa}) \Vhat^\star  = ( \Phat_{sa} - P_{sa}) \Vhat^{(s,u)}  + ( \Phat_{sa} - P_{sa})( \Vhat^\star - \Vhat^{(s,u)} ), \label{eq:pf_sketch_covering_0}
\end{equation}
so Bernstein's inequality can be applied to the first term, and the second term can be bounded by $2 \infnorm{\Vhat^\star - \Vhat^{(s,u)}}$. \citet{agarwal_modelbased_2020} construct such a $\Vhat^{(s,u)}$ using the idea of absorbing MDPs: from the empirical MDP $(\S, \A, \Phat, r)$, the state $s$ is made absorbing with reward $(1-\gamma)u$, so that $\Vhat^{(s,u)}(s) = u$. Since all the rows of $\Phat$ corresponding to state-actions in state $s$ are deleted, such a $\Vhat^{(s,u)}$ is also independent from $\Phat_{sa}$.
However, there is not a single deterministic choice of $u \in [0, 1/(1-\gamma)]$ which will ensure $\Vhat^\star - \Vhat^{(s,u)}$ is small; the ``ideal'' choice is $u = \Vhat^\star(s)$ in the sense that then $\Vhat^{(s,\Vhat^\star(s))} = \Vhat^\star$, but this is dependent on $\Phat_{sa}$.
We can thus try to control an interval of values $[L,R] \subseteq [0, 1/(1-\gamma)]$ of $u$, and as long as $\Vhat^\star(s) \in [L, R]$ then we have
\begin{equation}
    \left|( \Phat_{sa} - P_{sa}) \Vhat^\star \right| \leq \sup_{u \in [L,R]}\left|( \Phat_{sa} - P_{sa}) \Vhat^{(s,u)} \right|. \label{eq:pf_sketch_covering_1}
\end{equation}
One can crudely bound~\eqref{eq:pf_sketch_covering_1} by building a net for the entire interval $[0,1/(1-\gamma)]$, applying Bernstein's inequality for each point in the net, and taking a union bound. Following steps like~\eqref{eq:pf_sketch_covering_0} and using that $u \mapsto \Vhat^{(s,u)}$ is $1$-Lipschitz to bound $\tinfnorm{\Vhat^{(s,u)}-\Vhat^\star}$ \citep{agarwal_modelbased_2020}, this approach works, but the covering size needs to grow with $n$ to ensure $\tinfnorm{\Vhat^{(s,u)}-\Vhat^\star}$ is lower-order. Nonetheless this style of covering is still used in many related absorbing MDP constructions (e.g. \citealt{li_breaking_2024, zurek_plugin_2025,  zurek_optimal_2025}).

\citet{agarwal_modelbased_2020} refine this approach by using a smaller interval centered around the true $V^\star(s)$. One can relatively more easily show with high probability that $|\Vhat^\star(s) - V^\star(s)| \leq \frac{C}{(1-\gamma)^2} \sqrt{\frac{\log(eSA/\delta)}{n}} =: \Delta$, which is suboptimal in its dependence on the effective horizon but otherwise unimprovable (the proof uses Hoeffding's inequality, instead of the Bernstein-style bounds and total-variance arguments; see \citealt[Lemma 5]{agarwal_modelbased_2020}, \citealt[Lemma 4]{azar_minimax_2013}). Then one can construct a net only for the interval $[V^\star(s) - \Delta, V^\star(s) + \Delta]$, and the net size does not need to grow with $n$ since $\Delta$ already decreases as $n^{-1/2}$, but it does need to be polynomial in $\frac{1}{1-\gamma}$.
This is still insufficient for our theorems, which remove both the sample size and horizon dependencies.


\subsection{Bellman Residual Bracketing Concentration Analysis}
Now we describe our leave-one-out constructions and analysis, which are very closely inspired by the above ideas, but also make some substantial refinements.

First we perform an easy simplification. Consider the ``$(s,a)$-inactive MDP'' $\widehat{\mathcal{M}}_{-sa}$ which is equal to the empirical MDP, except that in state $s$, we delete action $a$ (or equivalently replace it with a copy of another action; see Definition~\ref{def:active_and_inactive_MDPs}). This MDP is clearly independent of $\Phat_{sa}$. Consequently we can always include its optimal value function $\Vabsinact$ in our cover for $\Vhat^\star$, and whenever it happens to be the case that $a$ is inactive (not taken by some optimal policy) we will have $\Vhat^\star = \Vabsinact$. The reason this will be helpful is that for the remaining case we know $(s,a)$ is active. (This is only valid when $A > 1$; if $A = 1$ then $a$ is always active.)

Thus we let $\VabsPhat$ be the optimal value function of the empirical MDP where state-action $(s,a)$ is forced to be active. We also let $\VabsP$ be the optimal value function of the same empirical MDP with $(s,a)$ forced active, except that we also replace its transition row $\Phat_{sa}$ by $P_{sa}$; we write this MDP as $\widetilde{\mathcal{M}}_{saP_{sa}}$. 
This value function $\VabsP$ is also clearly independent from $\Phat_{sa}$. 
Furthermore both of these value functions are realizable as absorbing MDPs for particular values of $u$, because they already satisfy the same Bellman equations as $\Vhat^{(s,u)}$ in all states other than state $s$. Thus define $u_0, \uhat$ so that $\Vhat^{(s,u_0)} = \VabsP$ and $\Vhat^{(s,\uhat)} = \VabsPhat$.
We further define two functions which measure the Bellman residual of $\Vhat^{(s,u)}$ at $s$ as a function of $u$, for both systems of Bellman equations:
\begin{equation}
    \phihat(u) := \frac{u - r(s,a)}{\gamma} - \Phat_{sa} \Vhat^{(s,u)}, \qquad 
    \phi(u) := \frac{u - r(s,a)}{\gamma} - P_{sa} \Vhat^{(s,u)}.
\end{equation}
By definition of $u_0$ and $\uhat$ we have that $\phi(u_0) = 0$ and $\phihat(\uhat) = 0$. Also for any $u$ we have
\begin{equation}
    \phi(u) - \phihat(u) = (\Phat_{sa} - P_{sa})\Vhat^{(s,u)}, \label{eq:pf_sketch_phi_minus_phihat}
\end{equation}
so combining all these definitions we have
\begin{equation}
    |\phi(\uhat)| = |\phi(\uhat) - \phihat(\uhat)|= \left|(\Phat_{sa} - P_{sa})\Vhat^{(s,\uhat)} \right| = \left|(\Phat_{sa} - P_{sa})\VabsPhat \right|. \label{eq:pf_sketch_residual_concentration_reln}
\end{equation}
Equation~\eqref{eq:pf_sketch_residual_concentration_reln} reveals that the term we hope to bound, $\left|(\Phat_{sa} - P_{sa})\VabsPhat \right|$, is exactly a certain (coordinate of a) Bellman residual of $\Vhat^{(s,u)}$ for the MDP $\widetilde{\mathcal{M}}_{saP_{sa}}$.
The reason this connection will be useful is that these Bellman-residual-measuring functions $\phi, \phihat$ can be shown to be strictly increasing, which is naturally an extremely useful property for controlling their fluctuations. To see why $\phi$ is strictly increasing, since $u \mapsto \Vhat^{(s,u)}$ is 1-Lipschitz (with respect to $\infnorm{\cdot}$), $\frac{u}{\gamma}$ grows more quickly in $u$ than $P_{sa} \Vhat^{(s,u)}$. (See Lemma \ref{lem:residual_properties}.)
We also note that this connection~\eqref{eq:pf_sketch_residual_concentration_reln} makes key use of the fact that $(s,a)$ is active.

Now we describe how the (strict) monotonicity of $\phi$ and $\phihat$ can be used to control $|\phi(\uhat)|$. By the monotonicity of $\phi$, if we can find some $u_-, u_+$ such that $u_- \leq \uhat \leq u_+$, then $\phi(u_-) \leq \phi(\uhat) \leq \phi(u_+)$. To obtain such endpoints $u_-, u_+$, by the strict monotonicity of $\phihat$, it would suffice for $\phihat(u_-) \leq \phihat(\uhat) \leq \phihat(u_+)$. Also by definition $\phihat(\uhat) =0 $, so in light of~\eqref{eq:pf_sketch_phi_minus_phihat}, we will try to choose $u_-$ to be sufficiently far below $u_0$ so that $\phihat(u_-) \leq 0$ with high probability: we have
\begin{equation*}
    \phihat(u_-) = \phi(u_-) + \left(\phihat(u_-) -  \phi(u_-)\right) \leq \phi(u_-) + \left|(\Phat_{sa} - P_{sa})\Vhat^{(s, u_-)} \right|
\end{equation*}
by~\eqref{eq:pf_sketch_phi_minus_phihat}, so $\phi(u_-)$ should be negative enough that its magnitude is larger than the second term in the above display, which can be bounded by Bernstein's inequality.
Analogously $u_+$ should have $\phihat(u_+) \geq 0$ with high probability. At the same time they should be as close to the center $u_0$ as possible to keep the interval small.

There is still more important technique involved to construct such endpoints $u_-, u_+$ and also to get the exact form of bound on $|\phi(\uhat)|$ that we desire.
We ultimately want the bound on $|\phi(\uhat)|$ to have the form
\begin{equation*}
    |\phi(\uhat)| \leq \sqrt{\frac{C\Var_{P_{sa}} [\Vhat^\star] \log(e/\delta)}{n}} + \frac{C \log(e /\delta)}{(1-\gamma)n}
\end{equation*}
that is, involving $\Vhat^\star = \Vhat^{(s,\uhat)}$ in the variance on the RHS. However, if we naively defined $u_-$ to satisfy
\begin{equation*}
    \phi(u_-) = - \left( \sqrt{\frac{C\Var_{P_{sa}} [\Vhat^\star] \log(e/\delta)}{n}} + \frac{C \log(e /\delta)}{(1-\gamma)n} \right)
\end{equation*}
(and likewise for $u_+$ but using a positive value) then (i) this would not be a valid definition for $u_-$ because it depends on $\Phat_{sa}$ through $\uhat$, and (ii) even if we could obtain a Bernstein-like inequality for $\phi(u_-) - \phihat(u_-) = (\Phat_{sa} - P_{sa})V^{(s, u_-)}$, we wouldn't immediately be able to conclude that $\phihat(u_-) < 0$ with high probability, due to the mismatch between $\Var_{P_{sa}}[\Vhat^{(s,\uhat)}]$ and $\Var_{P_{sa}}[\Vhat^{(s,u_-)}]$.
This suggests that we need a way to relate these different variance quantities.
The naive way of doing so, by using $1$-Lipschitzness of $u \mapsto \Vhat^{(s,u)}$ and bounding
\begin{equation*}
    \sqrt{\Var_{P_{sa}}[\Vhat^{(s, u)}]} \leq \sqrt{\Var_{P_{sa}}[\Vhat^{(s, v)}]} + \sqrt{\Var_{P_{sa}}[\Vhat^{(s, u)}-\Vhat^{(s, v)}]} \leq \sqrt{\Var_{P_{sa}}[\Vhat^{(s, v)}]} + |u - v|,
\end{equation*}
is actually insufficient: we will eventually need to upper-bound $|u - v|$ in terms of $|\phi(u) - \phi(v)|$, and we only generally have
\begin{equation*}
    |u - v| \leq \frac{\gamma}{1-\gamma}|\phi(u) - \phi(v)|
\end{equation*}
(see Equation~\eqref{eq:residual_properties_1}), so the variance parameter conversion will eventually only work when $n$ is at least order $\frac{C \log(e/\delta)}{(1-\gamma)^2}$. On the other hand we can also show $\gamma |\phi(u) - \phi(v)| \leq |u - v|$, suggesting that it could be sharper to instead bound the variance difference in terms of the residual difference $|\phi(u) - \phi(v)|$ directly. Indeed Lemma \ref{lem:residual_properties} shows 
\begin{equation}
    \sqrt{\Var_{P_{sa}}[\Vhat^{(s,u)} - \Vhat^{(s,v)}]} \leq \frac{\gamma}{2\sqrt{1-\gamma}} |\phi(u) - \phi(v)|, \label{eq:pf_sketch_variance_transfer}
\end{equation}
improving the horizon dependence.
Letting
\begin{equation*}
    B := \sqrt{\frac{C\Var_{P_{sa}} [\Vhat^{(s,u_0)}] \log(e/\delta)}{n}} + \frac{C \log(e /\delta)}{(1-\gamma)n} \leq \sqrt{\frac{C\Var_{P_{sa}} [\Vhat^{\star}] \log(e/\delta)}{n}} + \frac{C' \log(e /\delta)}{(1-\gamma)n}
\end{equation*}
and defining\footnote{Actually we only make this definition when the image of $[0,1/(1-\gamma)] $ under $\phi$ includes $-B, B$; if not we can simply set $u_- =0 $ and/or $u_+ = 1/(1-\gamma)$, and then the properties $u_- \leq \uhat$ and/or $\uhat \leq u_+$ are immediate.} $u_-, u_+$ so that $\phi(u_-) = -B, \phi(u_+) = B,$
we can then show that
\begin{equation*}
    -B \leq \phi(u_-) \leq \phi(\uhat) \leq \phi(u_+) \leq B
\end{equation*}
as desired.

\subsection{Average-Reward and Plug-in Results}

Now we discuss the proofs of our two results, based on discounted reduction and the plug-in approach, respectively, for solving average-reward MDPs.

Having established our improved results for the discounted plug-in approach, the discounted-reduction-based Theorem \ref{thm:main_average_reward_reduction} then follows from established techniques.
More specifically,
the main discounted Theorem \ref{thm:main_theorem_with_horizon} actually follows from a refinement (Theorem \ref{thm:main_intermediate_horizon_version}) which replaces one factor of $\frac{1}{1-\gamma}$ by $\tspannorm{V^\star} +\tspannorm{\Vhat^\star}$, similarly to the results of \cite{zurek_spanbased_2024, zurek_plugin_2025}.
Furthermore using the approximate discounted optimality of the empirical policy to argue $ \tspannorm{\Vhat^\star} \approx \tspannorm{V^\star}$ as well as the bound $\tspannorm{V^\star} \leq 2 \tspannorm{h^\star}$ \citep{wei_modelfree_2020}, we obtain the optimal span-based rate.

Now we discuss our average-reward plug-in result, Theorem \ref{thm:main_average_reward_plug_in}.
A natural and attractive idea is to view the average-reward plug-in approach as the limit of the discounted plug-in approach as $\gamma \to 1$. 
Specifically, there is some sufficiently large $\gbar < 1$ for which both the gain is close to the normalized discounted value function and also bias-optimal policies are nearly discounted-optimal, simultaneously for the true MDP and also for all (finitely many) possible empirical MDPs.
The detailed choice of $\gbar$ is provided by Lemma \ref{lem:bias_optimal_planning_discount_properties}.
Since the associated effective horizon $\frac{1}{1-\gbar}$ is generally very large and not bounded in terms of the complexity parameters which define the usual minimax classes for the study of average-reward MDPs, the logarithmic factors of $\frac{1}{1-\gbar}$ contained in previous discounted guarantees obstructed this discounted limit approach from working.

However, while the logarithmic term in our developed concentration bound is sufficient for this purpose, the $n^{-1}$ Bernstein term needs to be refined to replace the $\frac{1}{1-\gamma}$ scaling with something bounded as $\gamma \uparrow 1$.
This is immediate for the true optimal value Bernstein-like inequality, because we can immediately use $\spannorm{V^{\pistar}}$ instead of its conservative upper-bound $\frac{1}{1-\gamma}$, and $\tspannorm{V^{\pistar}} \leq 2 \tspannorm{h^\star}$ \citep{wei_modelfree_2020}.
For the empirical optimal value concentration inequality, we will eventually establish that
\begin{equation*}
    \left| (\Phat_{sa} - P_{sa}) \Vhat^\star \right| \leq \sqrt{\frac{\Var_{P_{sa}}[\Vhat^\star] \ell}{n}} + \frac{D \ell}{n}.
\end{equation*}
This is again much more challenging and interesting than the true optimal concentration bound. The first step that breaks from the above sketch is the variance transfer inequality~\eqref{eq:pf_sketch_variance_transfer}, which still introduces a dependence on $\frac{1}{1-\gamma}$.
Instead we can develop a bound which directly depends on the span seminorms of the involved value functions:
\begin{equation*}
    \Var_{P_{sa}} [\Vhat^{(s,u)} - \Vhat^{(s,v)}] \leq |\phi(u) - \phi(v)| \left(\tspannorm{\Vhat^{(s,u)}} + \tspannorm{\Vhat^{(s,v)}} \right).
\end{equation*}
See Lemma \ref{lem:variance_residual_bound_span_version}.
However, it is still nontrivial to control the spans of all involved absorbing MDP value functions, and this requires more care in choosing the center $u_0$ and the endpoints $u_-, u_+$.

Firstly, we can show that the empirical MDP has diameter $O(D)$ under this condition.
This follows the same idea from \citet{zurek_plugin_2025} and applies our (improved) 
discounted results on $S$ auxiliary MDPs where, for each $s \in \S$, reward is given only for reaching state $s$; achieving nontrivial reward on such an MDP implies that reaching the particular state is possible in the empirical MDP.

Now we consider how to restrict the interval of absorbing MDP parameters so that the span of $\Vhat^{(s,u)}$ is controlled by $O(D)$. 
This established diameter bound is useful because by standard arguments (see Lemma \ref{lem:optimal_value_fn_difference_diameter}) it implies
\begin{equation*}
    \Vhat^{(s,u)}(s') \geq \Vhat^{(s,u)}(s) - \diam(\Phat) \geq u - O(D).
\end{equation*}
Therefore we will have $\tspannorm{\Vhat^{(s,u)}} \leq O(D)$ whenever $u$ is such that state $s$ is the highest-value state or within $O(D)$ of the highest value. (See Lemma \ref{lem:bounded_span_interval_monotone}.)
Using the simple facts that $u \mapsto \Vhat^{(s,u)}$ is monotone and 1-Lipschitz (see Lemma \ref{lem:absorbing_mdp_properties_monotonicity}), the set of all $u$ such that $\Vhat^{(s,u)} \leq u + O(D)$ (that is, the value $u$ at state $s$ is at least as large as all other states minus $O(D)$) is actually an interval, because if it contains any $v \in [0, 1/(1-\gamma)]$, then it also contains all larger $u \geq v$. We let $\vmins$ be the lowest point of this interval.
Thus we will choose the endpoints to be at least $\vmins$, and we will still have $\uhat$ in the interval even if the lower endpoint has residual $ > -B$, because the fact that the diameter is $O(D)$ separately ensures $\tspannorm{\Vhat^{(s,\uhat)}} = \tspannorm{\Vhat^\star} \leq O(D)$.

Finally we briefly mention the most important details that were omitted for clarity. We must actually consider different cases depending on where $\vmins$ falls and also potentially adjust the center point, because we may not have $\tspannorm{\Vhat^{(s, u_0)}} \leq O(D)$. Consequently we do not use the radius $B$, which is based on the variance with respect to $\Vhat^{(s, u_0)}$, but instead one based on the modified center point $v_0 = \max\{u_0, \vmins\}$.

\section{Error Decomposition}
\label{sec:error_decomposition}
In this section we decompose the desired error bounds within our main theorems into several steps, ultimately reducing their proofs to a few concentration inequalities. Proving the needed concentration inequalities will be the subject of the next section. All of the arguments in this section are deterministic and may treat the needed concentration bounds as inputs.

First we give standard bounds of the suboptimality and value errors in terms of optimal-policy and empirical-optimal-policy evaluation errors.
\begin{lemma}
    \label{lem:deterministic_error_decomp}
    Let $\pistar \in \PiSD$ be a deterministic discounted-optimal policy.
    Suppose that $\infnorm{\Vhat^{\pihat} - \Vhat^\star} \!\leq \eopt$. Let $\epistar = \infnorm{\Vhat^{\pistar} - V^{\pistar}}$ and $\epihat = \infnorm{\Vhat^{\pihat} - V^{\pihat}}$. Then
    \begin{equation*}
        V^{\pihat} \geq V^{\star} - (\epistar + \eopt + \epihat)\one \quad \text{and} \quad \infnorm{\Vhat^\star - V^\star} \leq \eopt + \epihat + \epistar.
    \end{equation*}
\end{lemma}
\begin{proof}
    Using the assumptions, we have the elementwise bounds
    \begin{align*}
        V^{\pihat} & \geq \Vhat^{\pihat} - \epihat \one \\
        & \geq \Vhat^\star - (\eopt + \epihat)\one \\
        & \geq \Vhat^{\pistar} - (\eopt + \epihat)\one \\
        & \geq V^{\pistar} - (\epistar + \eopt + \epihat)\one.
    \end{align*}
    Additionally, since $V^\star = V^{\pistar} \geq V^{\pihat}$, the same inequality steps from above imply that
    \begin{equation*}
        V^\star \geq V^{\pihat} \geq \Vhat^\star - (\eopt + \epihat)\one
    \end{equation*}
    and also
    \begin{equation*}
        \Vhat^\star \geq V^{\pistar} - \epistar \one = V^\star - \epistar \one,
    \end{equation*}
    so
    \begin{equation*}
        \infnorm{\Vhat^\star - V^\star} \leq \max\{\eopt + \epihat, \epistar\} \leq \eopt + \epihat + \epistar.
    \end{equation*}
\end{proof}

Next, assuming a Bernstein-like concentration inequality, we bound a policy evaluation error term by a certain variance parameter.
\begin{lemma}
    \label{lem:error_to_var_param}
    Let $P, \Phat$ be two MDP transition kernels with common state and action spaces $\S$ and $\A$, respectively. Let $\pi \in \PiSR$ be a policy and suppose that $V \in \R^S$ satisfies $\infnorm{V^\pi - V} \leq \varepsilon$. Also suppose that $M \geq \spannorm{V}$ and
    \begin{equation}
        \left|(\Phat_\pi - P_\pi)V \right| \leq \sqrt{\frac{\alpha \Var_{\Phat_\pi}[V]}{n}} + \frac{\alpha M}{n} \one\label{eq:err_to_var_lemma_bernstein_ineq}
    \end{equation}
    for some scalar $\alpha > 0$.
    Then
    \begin{equation*}
        \infnorm{\Vhat^\pi - V^\pi} \leq \frac{\gamma \sqrt{\alpha}}{\sqrt{n}\sqrt{1-\gamma}} \infnorm{ \sqrt{ (I - \gamma \Phat_\pi)^{-1} \Var_{\Phat_\pi}[V]}} + \gamma\frac{ 2 n\varepsilon + \alpha M}{n(1-\gamma)}.
    \end{equation*}
\end{lemma}
\begin{proof}
    By the standard simulation lemma identity we have
    \begin{align*}
        \Vhat^\pi - V^\pi = \gamma(I - \gamma \Phat_\pi)^{-1} (\Phat_\pi - P_\pi) V^\pi.
    \end{align*}
    By standard calculations we can bound elementwise
    \begin{align*}
        \left|\Vhat^\pi - V^\pi \right| &= \left| \gamma(I - \gamma \Phat_\pi)^{-1} (\Phat_\pi - P_\pi) V^\pi \right| \leq \gamma (I - \gamma \Phat_\pi)^{-1} \left|(\Phat_\pi - P_\pi) V^\pi \right|
    \end{align*}
    since each entry of $(I - \gamma \Phat_\pi)^{-1}$ is non-negative. 
    From the assumption that $\infnorm{V^\pi - V} \leq \varepsilon$, we have that
    \begin{equation*}
        \left|(\Phat_\pi - P_\pi) V^\pi \right| \leq \left|(\Phat_\pi - P_\pi) V \right| + \left|(\Phat_\pi - P_\pi) (V^\pi - V) \right| \leq \left|(\Phat_\pi - P_\pi) V \right| + 2\varepsilon \one
    \end{equation*}
    where the final step is by Holder's inequality. With this we can further bound
    \begin{align*}
        \gamma (I - \gamma \Phat_\pi)^{-1} \left|(\Phat_\pi - P_\pi) V^\pi \right|
        & \leq \gamma (I - \gamma \Phat_\pi)^{-1} \left|(\Phat_\pi - P_\pi) V \right| + \frac{\gamma 2 \varepsilon}{1-\gamma}\one.
    \end{align*}
    Combining all steps with assumption~\eqref{eq:err_to_var_lemma_bernstein_ineq} we obtain
    \begin{equation*}
        \left|\Vhat^\pi - V^\pi \right| \leq \gamma (I - \gamma \Phat_\pi)^{-1} \sqrt{\frac{\alpha \Var_{\Phat_\pi}[V]}{n}} + \gamma\frac{ 2 n\varepsilon + \alpha M}{n(1-\gamma)}\one.
    \end{equation*}
    The matrix $(I - \gamma \Phat_\pi)^{-1}$ has all-nonnegative entries and each row sums to $\frac{1}{1-\gamma}$, so each row of $(1-\gamma)(I - \gamma \Phat_\pi)^{-1}$ is a probability distribution. Thus by the concavity of $\sqrt{\cdot}$ and Jensen's inequality applied to each row, we have elementwise
    \begin{equation*}
        (I - \gamma \Phat_\pi)^{-1} \sqrt{\Var_{\Phat_\pi}[V]} \leq \frac{1}{1-\gamma} \sqrt{(1-\gamma) (I - \gamma \Phat_\pi)^{-1}\Var_{\Phat_\pi}[V]} = \frac{1}{\sqrt{1-\gamma}} \sqrt{(I - \gamma \Phat_\pi)^{-1}\Var_{\Phat_\pi}[V]}.
    \end{equation*}
    Combining with the previous display we obtain
    \begin{equation*}
        \left|\Vhat^\pi - V^\pi \right| \leq \frac{\gamma}{\sqrt{1-\gamma}}  \sqrt{\frac{\alpha (I - \gamma \Phat_\pi)^{-1} \Var_{\Phat_\pi}[V]}{n}} + \gamma\frac{ 2 n\varepsilon + \alpha M}{n(1-\gamma)}\one.
    \end{equation*}
    Taking $\infnorm{\cdot}$ of both sides yields the desired conclusion.
\end{proof}

Next we show how this variance parameter can itself be bounded using only the same Bernstein-like concentration inequality.
\begin{lemma}
\label{lem:var_lemma}
    Let $P, \Phat$ be two MDP transition kernels with common state and action spaces $\S$ and $\A$, respectively. Let $\pi \in \PiSR$ be a policy and suppose $V \in \R^S$ satisfies $\infnorm{V - r_\pi - \gamma P_\pi V} \leq \varepsilon $ and $V \geq V^\pi$. Also suppose that $M \in \R$ has $M \geq \spannorm{V}$ and
    \begin{equation}
        \left|(\Phat_\pi - P_\pi)V \right| \leq \sqrt{\frac{\alpha \Var_{\Phat_\pi}[V]}{n}} + \frac{\alpha M}{n} \one\label{eq:var_lemma_bernstein_ineq}
    \end{equation}
    for some scalar $\alpha > 0$. Then
    \begin{equation*}
        \infnorm{(I - \gamma \Phat_\pi)^{-1} \Var_{\Phat_\pi}[V]} \leq 8\frac{\spannorm{V} M}{\gamma^2} \frac{\alpha}{n(1-\gamma)} + \frac{4\spannorm{V} (\spannorm{r_\pi}+\varepsilon)}{\gamma^2(1-\gamma)} .
    \end{equation*}
\end{lemma}
\begin{proof}
    To obtain a bound depending on $\spannorm{V}$ rather than $\infnorm{V}$, defining $v := \min_{s \in \S}V(s)$, $\Vs := V - v \one$, and $\rs := r_\pi - (1-\gamma)v \one$, we note that
    \begin{align}
        \infnorm{\Vs - \rs - \gamma P_\pi \Vs} &= \infnorm{V - v\one - r_\pi + (1-\gamma)v\one - \gamma P_\pi (V - v \one)} \nonumber\\
        &= \infnorm{V - r_\pi - \gamma P_\pi V} \leq \varepsilon \label{eq:var_lemma_shifted_approx_bellman_eqn}
    \end{align}
    since $P_\pi \one = \one$.

    Next we bound $\Var_{\Phat_\pi}[V]$ elementwise. Define $\dbe = \rs + \gamma P_\pi \Vs - \Vs$, which by~\eqref{eq:var_lemma_shifted_approx_bellman_eqn} has $\infnorm{\dbe} \leq \varepsilon$. Then we calculate that
    \begin{align*}
        \Var_{\Phat_\pi}[V]
        &= \Var_{\Phat_\pi}[\Vs] \\
        &= \Phat_\pi \Vs^{\circ 2} - (\Phat_\pi \Vs)^{\circ 2} \\
        &= \Phat_\pi \Vs^{\circ 2} - \frac{1}{\gamma^2}(\gamma \Phat_\pi \Vs)^{\circ 2} \\
        &= \Phat_\pi \Vs^{\circ 2} - \frac{1}{\gamma^2}\left(\gamma P_\pi \Vs + \gamma (\Phat_\pi - P_\pi)\Vs\right)^{\circ 2} \\
        &= \Phat_\pi \Vs^{\circ 2} - \frac{1}{\gamma^2}\left(\gamma P_\pi \Vs + \gamma (\Phat_\pi - P_\pi)\Vs\right)^{\circ 2} \\
        &= \Phat_\pi \Vs^{\circ 2} - \frac{1}{\gamma^2}\left(\Vs - \rs + \dbe + \gamma (\Phat_\pi - P_\pi)\Vs\right)^{\circ 2}
    \end{align*}
    where the first equality is because the variance is invariant to constant shifts.
    Expanding the square on the RHS of the above display, we obtain
    \begin{align}
        &\Var_{\Phat_\pi}[V] \nonumber\\
        &= \Phat_\pi \Vs^{\circ 2} - \frac{1}{\gamma^2}\left[\Vs^{\circ 2} + \left(- \rs + \dbe + \gamma (\Phat_\pi - P_\pi)\Vs\right)^{\circ 2} + 2\Vs \circ \left(- \rs + \dbe + \gamma (\Phat_\pi - P_\pi)\Vs\right) \right] \nonumber\\
        &\leq \frac{1}{\gamma^2} (\gamma^2 \Phat_\pi - I) \Vs^{\circ 2} - \frac{2}{\gamma^2}\Vs \circ \left(- \rs + \dbe + \gamma (\Phat_\pi - P_\pi)\Vs\right) \label{eq:var_lemma_variance_calculation_-1}
    \end{align}
    where the inequality comes from dropping the non-positive term. Further bounding the rightmost term of~\eqref{eq:var_lemma_variance_calculation_-1}, we have
    \begin{align}
        -\Vs \circ \left(- \rs + \dbe + \gamma (\Phat_\pi - P_\pi)\Vs\right)
        &= \Vs \circ \left(r_\pi - (1-\gamma)v\one - \dbe - \gamma (\Phat_\pi - P_\pi)\Vs\right)\nonumber\\
        &\stackrel{(i)}{\leq} \Vs \circ \left(r_\pi - (1-\gamma)\min_{s \in \S}V^\pi(s)\one - \dbe - \gamma (\Phat_\pi - P_\pi)\Vs\right)\nonumber\\
        &\stackrel{(ii)}{\leq} \Vs \circ \left(\spannorm{r_\pi}\one - \dbe - \gamma (\Phat_\pi - P_\pi)\Vs\right)\nonumber \\
        &\leq \Vs \circ \left(\spannorm{r_\pi} + \infnorm{\dbe} \right)\one + \gamma \Vs \circ \left|(\Phat_\pi - P_\pi)\Vs\right| \nonumber \\
        &\leq \infnorm{\Vs} \left(\spannorm{r_\pi} + \infnorm{\dbe} \right)\one + \gamma  \infnorm{\Vs} \left|(\Phat_\pi - P_\pi)\Vs\right| \nonumber \\
        &\stackrel{(iii)}{=} \infnorm{\Vs} \left(\spannorm{r_\pi} + \infnorm{\dbe} \right)\one + \gamma  \infnorm{\Vs} \left|(\Phat_\pi - P_\pi)V\right|\label{eq:var_lemma_variance_calculation_0}
    \end{align}
    where all inequality steps use that $\Vs \geq \zero$, and furthermore $(i)$ uses that $V \geq V^\pi$ so $v \geq \min_{s \in \S}V^\pi(s) $, $(ii)$ uses that $(1-\gamma)\min_{s \in \S}V^\pi(s) \geq \min_{s \in \S} r_\pi(s)$, and $(iii)$ uses that $(\Phat_\pi - P_\pi) \Vs = (\Phat_\pi - P_\pi) V$ since $\Phat_\pi \one = P_\pi \one = \one$.
    Combining~\eqref{eq:var_lemma_variance_calculation_0} with~\eqref{eq:var_lemma_variance_calculation_-1} we obtain that
    \begin{align}
        \Var_{\Phat_\pi}[V]
        &\leq \frac{1}{\gamma^2} (\gamma^2 \Phat_\pi - I) \Vs^{\circ 2} + \frac{2\infnorm{\Vs}}{\gamma^2}  \left(\spannorm{r_\pi} + \infnorm{\dbe} \right)\one + \frac{2\infnorm{\Vs}}{\gamma}   \left|(\Phat_\pi - P_\pi)V\right|. \label{eq:var_lemma_variance_calculation_1}
    \end{align}

    Since $(I - \gamma \Phat_\pi)^{-1}$ has all entries nonnegative and thus preserves elementwise inequalities after multiplication, inequality~\eqref{eq:var_lemma_variance_calculation_1} implies
    \begin{align}
        (I - \gamma \Phat_\pi)^{-1} \Var_{\Phat_\pi}[V]
        &\leq \frac{1}{\gamma^2}(I - \gamma \Phat_\pi)^{-1} (\gamma^2 \Phat_\pi - I) \Vs^{\circ 2} + \frac{2\infnorm{\Vs}\left(\spannorm{r_\pi} + \varepsilon \right)}{(1-\gamma)\gamma^2}\one \nonumber\\
        & \quad + \frac{2\infnorm{\Vs}}{\gamma} (I - \gamma \Phat_\pi)^{-1}\left|(\Phat_\pi - P_\pi)V \right|\label{eq:var_lemma_variance_calculation_2}
    \end{align}
    (simplifying the second term on the RHS by using that $(I - \gamma \Phat_\pi)^{-1} \one = \frac{1}{1-\gamma}\one$ and that $\infnorm{\dbe}\leq \varepsilon$).
    Now we consider the first and third terms on the RHS of~\eqref{eq:var_lemma_variance_calculation_2}. 
    
    By a standard calculation, the first term can be shown to be non-positive, which we now show for completeness. We have
    \begin{align*}
        (I - \gamma \Phat_\pi)^{-1} (\gamma^2 \Phat_\pi - I)
        &= \sum_{t=0}^\infty \gamma^t \Phat_\pi^t (\gamma^2 \Phat_\pi - I)
        = \sum_{t=0}^\infty \gamma^{t+2} \Phat_\pi^{t+1} - \sum_{t=0}^\infty \gamma^t \Phat_\pi^t \\
        &= \sum_{t=0}^\infty \gamma^{t+2} \Phat_\pi^{t+1} - I - \sum_{t=0}^\infty \gamma^{t+1} \Phat_\pi^{t+1} \\
        & = -I + \sum_{t=0}^\infty (\gamma^{t+2} - \gamma^{t+1}) \Phat_\pi^{t+1} \leq -I \leq 0
    \end{align*}
    where the penultimate inequality is because $\gamma^{t+2} \leq \gamma^{t+1}$.
    Consequently $(I - \gamma \Phat_\pi)^{-1} (\gamma^2 \Phat_\pi - I) \Vs^{\circ 2} \leq \zero$ since $\Vs^{\circ 2} \geq \zero$.

    For the third term of~\eqref{eq:var_lemma_variance_calculation_2}, by~\eqref{eq:var_lemma_bernstein_ineq} we have
    \begin{align*}
        &\frac{2\infnorm{\Vs}}{\gamma} (I - \gamma \Phat_\pi)^{-1}\left|(\Phat_\pi - P_\pi)V \right| \\
        & \leq \frac{2\infnorm{\Vs}}{\gamma} (I - \gamma \Phat_\pi)^{-1} \left[\sqrt{\frac{\alpha \Var_{\Phat_\pi}[V]}{n}} + \frac{\alpha M}{n} \one \right] \\
        &= \frac{2\infnorm{\Vs}}{\gamma} (I - \gamma \Phat_\pi)^{-1}\sqrt{\frac{\alpha \Var_{\Phat_\pi}[V]}{n}} + \frac{2\infnorm{\Vs}}{\gamma(1-\gamma)}\frac{\alpha M}{n}\one \\
        & \leq \frac{2\infnorm{\Vs}}{\gamma}\sqrt{\frac{\alpha}{n(1-\gamma)}} \sqrt{(I - \gamma \Phat_\pi)^{-1}\Var_{\Phat_\pi}[V]} + \frac{2\infnorm{\Vs}}{\gamma(1-\gamma)}\frac{\alpha M}{n}\one.
    \end{align*}

    Combining these three bounds, and also observing that $\infnorm{\Vs} = \spannorm{V}$, we obtain that
    \begin{align}
        (I - \gamma \Phat_\pi)^{-1} \Var_{\Phat_\pi}[V]
        & \leq \frac{2\spannorm{V} (\spannorm{r_\pi}+\varepsilon)}{\gamma^2(1-\gamma)}\one + \frac{2\spannorm{V}}{\gamma(1-\gamma)}\frac{\alpha M}{n}\one \nonumber \\
        &\quad + \frac{2\spannorm{V}}{\gamma}\sqrt{\frac{\alpha}{n(1-\gamma)}} \sqrt{(I - \gamma \Phat_\pi)^{-1}\Var_{\Phat_\pi}[V]}. \nonumber
    \end{align}
    Since all entries of the vector $(I - \gamma \Phat_\pi)^{-1} \Var_{\Phat_\pi}[V]$ are non-negative, this implies
    \begin{align}
        \infnorm{(I - \gamma \Phat_\pi)^{-1} \Var_{\Phat_\pi}[V]}
        & \leq \frac{2\spannorm{V} (\spannorm{r_\pi}+\varepsilon)}{\gamma^2(1-\gamma)} + \frac{2\spannorm{V}}{\gamma(1-\gamma)}\frac{\alpha M}{n} \nonumber \\
        &\quad + \frac{2\spannorm{V}}{\gamma}\sqrt{\frac{\alpha}{n(1-\gamma)}} \sqrt{\infnorm{(I - \gamma \Phat_\pi)^{-1}\Var_{\Phat_\pi}[V]}}. \nonumber
    \end{align}
    Letting $x := \sqrt{\infnorm{(I - \gamma \Phat_\pi)^{-1} \Var_{\Phat_\pi}[V]}}$, we can rearrange the above to obtain the quadratic inequality in $x$
    \begin{align}
        x^2 - \frac{2\spannorm{V}}{\gamma}\sqrt{\frac{\alpha}{n(1-\gamma)}} x - \frac{2\spannorm{V} (\spannorm{r_\pi}+\varepsilon)}{\gamma^2(1-\gamma)} - \frac{2\spannorm{V}}{\gamma(1-\gamma)}\frac{\alpha M}{n} \leq 0.
    \end{align}
    Using the quadratic formula, this implies that
    \begin{align}
        x &\leq \frac{1}{2}\frac{2\spannorm{V}}{\gamma}\sqrt{\frac{\alpha}{n(1-\gamma)}} \nonumber\\
        &+ \frac{1}{2}\sqrt{\left(\frac{2\spannorm{V}}{\gamma}\sqrt{\frac{\alpha}{n(1-\gamma)}} \right)^2 + 4\left( \frac{2\spannorm{V} (\spannorm{r_\pi}+\varepsilon)}{\gamma^2(1-\gamma)} + \frac{2\spannorm{V}}{\gamma(1-\gamma)}\frac{\alpha M}{n}\right)} \nonumber .
    \end{align}
    Squaring both sides and using that
    $(\frac{a+b}{2})^2 = \frac{a^2 + b^2 + 2ab}{4} \leq \frac{a^2 + b^2}{2}$ by AM-GM, we have
    \begin{align*}
        x^2 &\leq \frac{1}{2} \frac{4 \spannorm{V}^2}{\gamma^2} \frac{\alpha}{n(1-\gamma)} \\
        &+ \frac{1}{2}\left( \left(\frac{2\spannorm{V}}{\gamma}\sqrt{\frac{\alpha}{n(1-\gamma)}} \right)^2 + 4\left( \frac{2\spannorm{V} (\spannorm{r_\pi}+\varepsilon)}{\gamma^2(1-\gamma)} + \frac{2\spannorm{V}}{\gamma(1-\gamma)}\frac{\alpha M}{n}\right)\right) \\
        &\leq 8\frac{M\spannorm{V}}{\gamma^2} \frac{\alpha}{n(1-\gamma)} + \frac{4\spannorm{V} (\spannorm{r_\pi}+\varepsilon)}{\gamma^2(1-\gamma)} 
    \end{align*}
    where in the second inequality we used that $\gamma \leq 1/\gamma$ and $\spannorm{V} \leq M$ (to help combine terms).
\end{proof}

Finally, the following lemma can be used to bound the error of a Q-function estimate produced from a value function estimate.
\begin{lemma}
\label{lem:Q_est_from_V_est}
    Suppose that $\widetilde{V} \in \R^S$ satisfies
    \begin{equation*}
        \infnorm{\widetilde{V} - V^\star} \leq \varepsilon \qquad\text{and}\qquad\left| (\Phat - P)V^\star\right| \leq \sqrt{\frac{\alpha \spannorm{V^\star}^2}{n}} \one
    \end{equation*}
    for some $\alpha, \varepsilon > 0$. Then
    \begin{equation*}
        \infnorm{r + \gamma \Phat \widetilde{V} - Q^\star} \leq \gamma \varepsilon + \gamma \sqrt{\frac{\alpha \spannorm{V^\star}}{(1-\gamma)^2n}}.
    \end{equation*}
\end{lemma}
\begin{proof}
    We calculate
    \begin{align*}
        \infnorm{r + \gamma \Phat \widetilde{V} - Q^\star} &= \gamma\infnorm{ \Phat \widetilde{V} - P V^\star} \\
        &\leq \gamma\infnorm{\Phat (\widetilde{V} - V^\star)} + \gamma\infnorm{(\Phat - P)V^\star} \\
        & \leq \gamma \varepsilon + \gamma \sqrt{\frac{\alpha \spannorm{V^\star}^2}{n}} \\
        & \leq \gamma \varepsilon + \gamma \sqrt{\frac{\alpha \spannorm{V^\star}}{(1-\gamma)^2n}} 
    \end{align*}
    using that $Q^\star = r + \gamma P V^\star$, $\tinfinfnorm{\Phat} \leq 1$, the assumptions, and then the coarse bound that $\spannorm{V^\star} \leq \frac{1}{1-\gamma} \leq \frac{1}{(1-\gamma)^2}$ in the final inequality.
\end{proof}

\section{Concentration Inequalities}
In this section we show that the concentration inequalities needed within the arguments of the previous section hold with high probability.

\subsection{True optimal value case}
We start with the much simpler bounds involving the true optimal value function. The below inequality is more general since it is stated for an arbitrary fixed function $V \in \R^S$, but later it will only be applied with $V = V^\star$.

\begin{lemma}
\label{lem:fixed_pol_empirical_bernstein}
    Fix a deterministic policy $\pi$ and $V \in \R^S$. Suppose $n \geq 6$ and let $\delta \in (0,1)$. Then with probability at least $1 - \delta$,
    \begin{equation*}
        \left|(\Phat_{\pi} - P_{\pi})V \right| \leq \sqrt{\frac{3\Var_{\Phat_{\pi}}[V]\log(4S/\delta)}{n}} + \frac{3\spannorm{V}\log(4S/\delta)}{n}.
    \end{equation*}
\end{lemma}
\begin{proof}
    First we fix $s \in \S, \delta' \in (0,1)$ and observe that by the empirical Bernstein bound of Lemma \ref{lem:empirical_bernstein}, with probability at least $1-\delta'$,
    \begin{equation}
        \left|(\Phat_{s\pi(s)} - P_{s\pi(s)})V \right| \leq \sqrt{\frac{3\Var_{\Phat_{s\pi(s)}}[V]\log(4/\delta')}{n}} + \frac{3\spannorm{V}\log(4/\delta')}{n}.
    \end{equation}
    By a union bound over all $s \in \S$, with probability at least $1-S\delta'$ we have that
    \begin{equation}
        \left|(\Phat_\pi - P_\pi)V \right| \leq \sqrt{\frac{3\Var_{\Phat_\pi}[V]\log(4/\delta')}{n}} + \frac{3\spannorm{V}\log(4/\delta')}{n}\one.
    \end{equation}
    We conclude by setting $\delta' = \delta/S$.
\end{proof}

\subsection{Empirical optimal value case}
We make use of the absorbing MDP construction first developed in \cite{agarwal_modelbased_2020}.
\begin{definition}
\label{def:absorbing_mdp}
    Recall the empirical MDP $\widehat{\mathcal{M}} = (\S, \A, \Phat, r)$. Fix $s \in \S$ and $u \in [0, 1/(1-\gamma)]$. Define the $(s,u)$-absorbing MDP $\widehat{\mathcal{M}}_{s,u} = (\S, \A, \Phat^{(s)}, r^{(s,u)})$ where, for any $s' \in \S, a' \in \A$,
    \begin{equation*}
        \Phat^{(s)}_{s'a'} = \begin{cases}
            \Phat_{s'a'} & s' \neq s \\
            e_{s}^\top & s' =s
        \end{cases},
        \qquad
        r^{(s,u)}(s'a') = \begin{cases}
            r(s'a') & s' \neq s \\
            (1-\gamma)u & s' =s
        \end{cases}.
    \end{equation*}
    We define $\Vhat^{(s,u)} := V^\star_{\widehat{\M}_{s,u}}$ to be the optimal discounted value function of the $(s,u)$-absorbing MDP $\widehat{\mathcal{M}}_{s,u}$.
\end{definition}

\begin{lemma}
\label{lem:absorbing_mdp_properties_1}
    \begin{enumerate}
        \item For any $s \in \S, u \in [0, 1/(1-\gamma)]$, $\Vhat^{(s,u)}(s) = u$.
        \item For any $s \in \S$, letting $\hat{u}^\star = \Vhat^\star(s)$, we have $\Vhat^{(s,\hat{u}^\star)} = \Vhat^\star$.
        \item For any $s \in \S$ and any $u, u' \in [0, 1/(1-\gamma)]$,
        \begin{equation*}
            \infnorm{\Vhat^{(s,u)} - \Vhat^{(s,u')}} \leq |u - u'|.
        \end{equation*}
    \end{enumerate}
\end{lemma}
\begin{proof}
    These three statements are proved in the second paragraph of \cite[Section 4.2]{agarwal_modelbased_2020}, \cite[Lemma 7]{agarwal_modelbased_2020}, and \cite[Lemma 8]{agarwal_modelbased_2020}, respectively.
\end{proof}

The following simple lemma shows the monotonicity of the absorbing MDPs as the reward parameter is increased.
\begin{lemma}
    \label{lem:absorbing_mdp_properties_monotonicity}
    Suppose $u, v \in [0, 1/(1-\gamma)]$ satisfy $u \leq v$. Then for any $s \in \S$,
    \begin{equation*}
        \Vhat^{(s,u)} \leq \Vhat^{(s,v)}.
    \end{equation*}
\end{lemma}
\begin{proof}
    Let $\pi_u$ be an optimal policy for the $(s,u)$-absorbing MDP $\widehat{\mathcal{M}}_{s,u}$. Since $\widehat{\mathcal{M}}_{s,u}$ and $\widehat{\mathcal{M}}_{s,v}$ have the same transition kernel $\Phat^{(s)}$ and differ only in their reward functions, which satisfy $r^{(s,u)} \leq r^{(s,v)}$ elementwise and thus $r^{(s,u)}_{\pi_u} \leq r^{(s,v)}_{\pi_u}$, we have
    \begin{equation*}
        \Vhat^{(s,u)} = V^{\pi_u}_{\widehat{\mathcal{M}}_{s,u}} = (I - \gamma \Phat^{(s)}_{\pi_u})^{-1}r^{(s,u)}_{\pi_u} \leq (I - \gamma \Phat^{(s)}_{\pi_u})^{-1}r^{(s,v)}_{\pi_u} = V^{\pi_u}_{\widehat{\mathcal{M}}_{s,v}} \leq V^{\star}_{\widehat{\mathcal{M}}_{s,v}} = \Vhat^{(s,v)}.
    \end{equation*}
\end{proof}

The main result of this subsection is the following.
\begin{lemma}
\label{lem:absorbing_bernstein_ineq}
    Fix $s\in \S, a\in \A$ and let $\delta \in (0,1)$. Suppose $n \geq \frac{16\gamma\log(4/\delta)}{1-\gamma}$. Then with probability at least $1 - \delta$,
    \begin{equation*}
        \left|(\Phat_{sa} - P_{sa})\Vhat^\star \right| \leq \sqrt{\frac{32\Var_{P_{sa}}[\Vhat^\star]\log(4/\delta)}{n}} + \frac{4\log(4/\delta)}{3(1-\gamma)n}.
    \end{equation*}
\end{lemma}
Before proving Lemma \ref{lem:absorbing_bernstein_ineq} we will develop some useful intermediate constructions and lemmas.
We treat $s \in \S, a \in \A$ as fixed for the remainder of this subsection to simplify notation and statements; all of the subsequent results and definitions implicitly depend on the identity of $s,a$.

\begin{definition}
\label{def:active_and_inactive_MDPs}
    For any distribution $q \in \Delta(\S)$, we define the $(s,a, q)$-active-MDP $\widetilde{\mathcal{M}}_{saq} = (\S, \A, \widetilde{P}^{(s,q)}, \widetilde{r}^{(s,a)})$ where
    \begin{equation*}
        \widetilde{P}^{(s,q)}_{s'a'} = \begin{cases}
            \Phat_{s'a'} & s' \neq s \\
            q & s' = s
        \end{cases},
        \qquad
        \widetilde{r}^{(s,a)}(s',a') = \begin{cases}
            r(s',a') & s' \neq s \\
            r(s,a) & s' = s
        \end{cases}.
    \end{equation*}
    We define the abbreviations $\VabsP := V^\star_{\widetilde{\mathcal{M}}_{saP_{sa}}}$ and $\VabsPhat := V^\star_{\widetilde{\mathcal{M}}_{sa\Phat_{sa}}}$.
    If $A > 1$, then we also define the $(s,a)$-inactive MDP $\widehat{\mathcal{M}}_{-sa} = (\S, \A, \Phat^{(-sa)}, r^{(-sa)})$ where
    \begin{equation*}
        \Phat^{(-sa)}_{s'a'} = \begin{cases}
            \Phat_{s'a'} & (s',a') \neq (s,a) \\
            \Phat_{s, \bar{a}} & (s'a') = (s,a)
        \end{cases},
        \qquad
        r^{(-sa)}(s'a') = \begin{cases}
            r(s',a') & (s',a') \neq (s,a) \\
            r(s, \bar{a}) & (s'a') = (s,a)
        \end{cases}
    \end{equation*}
    where $\bar{a}$ is the first action in $\A$ (with respect to an arbitrary ordering) that is not equal to $a$.
    We define the abbreviation $\Vabsinact := V^\star_{\widehat{\mathcal{M}}_{-sa}}$.
\end{definition}
In words, the $(s,a, q)$-active MDP forces all actions in state $s$ to have next-state transition distribution $q$ and reward $r(s,a)$.
When $q = \Phat_{sa}$ this can be understood as forcing action $a$ to be active in state $s$ of the empirical MDP. When instead $q = P_{sa}$, this yields an MDP which is still independent of $\Phat_{sa}$ but still leverages the information $(\Phat_{s'a'})_{(s'a') \neq (s,a)}$ and still forces $a$ to be active. 
The $(s,a)$-inactive MDP replaces $\Phat_{sa}$ and $r(s,a)$ with a redundant copy of another action from that state, effectively forcing action $a$ to be inactive in state $s$, and thus ensuring independence from $\Phat_{sa}$.

Action $a$ in state $s$ must either be active or inactive in the empirical MDP. Clearly in the inactive case we have $\Vhat^\star = \Vabsinact$, that is, the optimal value function of the $(s,a)$-inactive MDP $\widehat{\mathcal{M}}_{-sa}$ is equal to $\Vhat^\star$. While the event that $\Vhat^\star = \Vabsinact$ is of course not independent of $\Phat_{sa}$, we can simply add $\Vabsinact$ to our covering of value functions to handle the $(s,a)$-inactive case. Similarly, we will also show a concentration inequality for the value function $\VabsPhat$ which forces $a$ to be active in state $s$, which will handle the case that $a$ is indeed active.

Finally, we note that for the case that $A=1$, to avoid extra casework we can simply define $\Vabsinact$ to be an arbitrary fixed vector, and all of the following arguments will still hold since $\Vabsinact$ is only needed to be equal to $\Vhat^\star$ on the event that $(s,a)$ is inactive, which is impossible when $A=1$. For concreteness, we set $\Vabsinact := \zero$ in the case that $A=1$.

\begin{lemma}
\label{lem:active_inactive_split}
    If there exists $a^\star \neq a$ such that $a^\star \in \argmax_{a' \in \A} r(s,a') + \gamma \Phat_{sa'}\Vhat^\star$, then $\Vhat^\star = \Vabsinact.$ Otherwise if $a \in \argmax_{a' \in \A} r(s,a') + \gamma \Phat_{sa'}\Vhat^\star$, then $\Vhat^\star = \VabsPhat$. Consequently, $\Vhat^\star \in \{\Vabsinact, \VabsPhat\}$ almost surely.
\end{lemma}
\begin{proof}
    Under the first condition that $a^\star \in \argmax_{a' \in \A} r(s,a') + \gamma \Phat_{sa'}\Vhat^\star$, the Bellman optimality equation for the empirical optimal value function $\Vhat^\star$ in state $s$ yields that
    \begin{equation*}
        \Vhat^\star(s) = \max_{a' \in \A} r(s,a') + \gamma \Phat_{sa'} \Vhat^\star = \max_{a' \in \A \setminus \{a\}} r(s,a') + \gamma \Phat_{sa'} \Vhat^\star.
    \end{equation*}
    Since action $a$ is redundant with action $\bar{a}$ in state $s$ of the $(s,a)$-inactive MDP, the Bellman optimality equation for $\Vabsinact$ in state $s$ has the identical form
    \begin{equation*}
        \Vabsinact(s) = \max_{a' \in \A} r^{(-sa)}(s,a') + \gamma \Phat^{(-sa)}_{sa'} \Vabsinact = \max_{a' \in \A \setminus \{a\}}  r(s,a') + \gamma \Phat_{sa'} \Vabsinact.
    \end{equation*}
    Furthermore, since all rows of the transition kernel $\Phat^{(-sa)}$ and the reward function $r^{(-sa)}$ of the $(s,a)$-inactive MDP are identical to those of the empirical MDP $\Phat, r$ except for state action $(s,a)$, the Bellman optimality equations satisfied by $\Vhat^\star$ and $\Vabsinact$ in all other states $s' \neq s$ are identical. Therefore $\Vhat^\star$ and $\Vabsinact$ satisfy the same Bellman optimality equations under the condition that $a^\star \in \argmax_{a' \in \A} r(s,a') + \gamma \Phat_{sa'}\Vhat^\star$, and so under this condition they must be equal.

    Next we show the second part of the lemma. Suppose that $a \in \argmax_{a' \in \A} r(s,a') + \gamma \Phat_{sa'}\Vhat^\star$. Then we have
    \begin{equation*}
        \Vhat^\star(s) = r(s,a) + \gamma \Phat_{sa}\Vhat^\star
    \end{equation*}
    in state $s$, and for all $s' \neq s$, we have
    \begin{equation*}
        \Vhat^\star(s') = \max_{a' \in \A} r(s',a') + \gamma \Phat_{s'a'}\Vhat^\star,
    \end{equation*}
    which are exactly the optimality equations of the $(s,a,\Phat_{sa})$-active MDP $\widetilde{\M}_{sa \Phat_{sa}}$. Since they have the unique solution $\VabsPhat$, we must have $\Vhat^\star = \VabsPhat$.
\end{proof}

\begin{lemma}
\label{lem:concentration_inactive_case}
    Fix $\delta \in (0,1)$. Then with probability at least $1 - \delta$, we have
    \begin{equation*}
        \left|(\Phat_{sa} - P_{sa})\Vabsinact \right| \leq \sqrt{\frac{2\Var_{P_{sa}}[\Vabsinact]\log(2/\delta)}{n}} + \frac{\spannorm{\Vabsinact}\log(2/\delta)}{3n}. 
    \end{equation*}
\end{lemma}
\begin{proof}
    First we condition on $(\Phat_{s'a'})_{(s'a') \neq (s,a)}$. Since the random variables $S^{(1)}_{sa}, \dots, S^{(n)}_{sa}$ are independent from $(\Phat_{s'a'})_{(s'a') \neq (s,a)}$, and $\Vabsinact$ is $(\Phat_{s'a'})_{(s'a') \neq (s,a)}$-measurable, we have by the Bernstein inequality (Lemma \ref{lem:bernstein}) that with conditional probability at least $1-\delta$,
    \begin{equation*}
        \left|(\Phat_{sa} - P_{sa})\Vabsinact \right| \leq \sqrt{\frac{2\Var_{P_{sa}}[\Vabsinact]\log(2/\delta)}{n}} + \frac{\spannorm{\Vabsinact}\log(2/\delta)}{3n}. 
    \end{equation*}
    The unconditional statement then follows from the tower property.
\end{proof}

To handle the active case, we will show the following Bernstein-like inequality involving $\VabsPhat = V^\star_{\widetilde{\mathcal{M}}_{sa\Phat_{sa}}}$.
\begin{lemma}
    \label{lem:active_case_empirical_bernstein}
    Fix $\delta \in (0,1)$ and suppose that $n \geq \frac{16\gamma\log(2/\delta)}{1-\gamma}$. Then with probability at least $1 - \delta$,
    \begin{equation*}
        \left|(\Phat_{sa} - P_{sa})\VabsPhat \right| \leq \sqrt{\frac{32\Var_{P_{sa}}[\VabsPhat]\log(2/\delta)}{n}} + \frac{4\log(2/\delta)}{3n(1-\gamma)}.
    \end{equation*}
\end{lemma}
Given Lemmas \ref{lem:active_inactive_split}, \ref{lem:concentration_inactive_case}, and \ref{lem:active_case_empirical_bernstein}, we can immediately prove Lemma \ref{lem:absorbing_bernstein_ineq}.
\begin{proof}[Proof of Lemma \ref{lem:absorbing_bernstein_ineq}]
    We apply Lemmas \ref{lem:concentration_inactive_case} and \ref{lem:active_case_empirical_bernstein} with failure probability $\delta/2$, so that by the union bound their conclusions both hold with probability at least $1 - \delta$. By Lemma \ref{lem:active_inactive_split}, $\Vhat^\star \in \{\Vabsinact, \VabsPhat\}$, and so if $\Vhat^\star = \Vabsinact$ then
    \begin{align*}
        \left| (\Phat_{sa} - P_{sa})\Vhat^\star \right| &= \left| (\Phat_{sa} - P_{sa})\Vabsinact \right| \leq  \sqrt{\frac{2\Var_{P_{sa}}[\Vabsinact]\log(4/\delta)}{n}} + \frac{\spannorm{\Vabsinact}\log(4/\delta)}{3n} \\
        &=  \sqrt{\frac{2\Var_{P_{sa}}[\Vhat^\star]\log(4/\delta)}{n}} + \frac{\spannorm{\Vhat^\star}\log(4/\delta)}{3n} \\
        &\leq  \sqrt{\frac{32\Var_{P_{sa}}[\Vhat^\star]\log(4/\delta)}{n}} + \frac{4\log(4/\delta)}{3(1-\gamma)n}
    \end{align*}
    since $\spannorm{\Vhat^\star} \leq \frac{1}{1-\gamma}$.
    Similarly if $\Vhat^\star = \VabsPhat$ then
    \begin{align*}
        \left| (\Phat_{sa} - P_{sa})\Vhat^\star \right| &= \left| (\Phat_{sa} - P_{sa})\VabsPhat \right| \leq  \sqrt{\frac{32\Var_{P_{sa}}[\VabsPhat]\log(4/\delta)}{n}} + \frac{4\log(4/\delta)}{3(1-\gamma)n} \\
        &=  \sqrt{\frac{32\Var_{P_{sa}}[\Vhat^\star]\log(4/\delta)}{n}} + \frac{4\log(4/\delta)}{3(1-\gamma)n}.
    \end{align*}
\end{proof}

The remainder of this section therefore is dedicated to showing Lemma \ref{lem:active_case_empirical_bernstein}.
\begin{definition}
    We define the true and empirical residual functions $\phi, \phihat : [0, 1/(1-\gamma)] \to \R$ by
    \begin{align}
        \phi(u) &:= \frac{u - r(s,a)}{\gamma} - P_{sa}\Vhat^{(s,u)}, \\
        \phihat(u) &:= \frac{u - r(s,a)}{\gamma} - \Phat_{sa}\Vhat^{(s,u)}.
    \end{align}
\end{definition}

\begin{lemma}
    \label{lem:residual_properties}
    The functions $\phi$ and $\phihat$ are strictly increasing.
    Also, supposing that $v, u \in [0, 1/(1-\gamma)]$ satisfy $u \leq v$, then
    \begin{equation}
        \Var_{P_{sa}}\left[\Vhat^{(s,v)} - \Vhat^{(s,u)}\right] \leq \frac{\gamma^2}{4(1-\gamma)} \left|\phi(v) - \phi(u) \right|^2. \label{eq:absorbing_mdp_val_residual_bd_2}
    \end{equation}
\end{lemma}
\begin{proof}
    First we show that $\phi$ is strictly increasing. Letting $ u \leq v$, we have
    \begin{equation}
        \phi(v) - \phi(u) = \frac{v - u}{\gamma} - P_{sa}(\Vhat^{(s,v)} - \Vhat^{(s,u)}) \geq \frac{v - u}{\gamma} - (v-u)P_{sa}\one = \frac{1-\gamma}{\gamma}(v-u) \label{eq:residual_properties_1}
    \end{equation}
    where the inequality step is because elementwise $\Vhat^{(s,v)} - \Vhat^{(s,u)} \leq (v-u)\one$, since as a consequence of Lemma \ref{lem:absorbing_mdp_properties_1} we have
    \begin{equation*}
        \Vhat^{(s,v)} \leq \Vhat^{(s,u)} + \infnorm{\Vhat^{(s,v)} - \Vhat^{(s,u)}}\one \leq \Vhat^{(s,u)} + |v - u|\one = \Vhat^{(s,u)} + (v - u)\one.
    \end{equation*}
    In particular if $v > u$ then $\phi(v) > \phi(u)$, so $\phi$ is strictly increasing. The same proof but with $\Phat_{sa}$ in place of $P_{sa}$ shows that $\phihat$ is also strictly increasing.

    The monotonicity shown in Lemma \ref{lem:absorbing_mdp_properties_monotonicity} implies $\Vhat^{(s,v)} - \Vhat^{(s,u)} \geq \zero$, and combining with the upper bound on $\Vhat^{(s,v)} - \Vhat^{(s,u)}$ shown above, we have
    \begin{equation}
        \zero \leq \Vhat^{(s,v)} - \Vhat^{(s,u)} \leq (v-u)\one. \label{eq:absorbing_MDP_diff_sandwich_bounds}
    \end{equation}
    Combining these bounds with Lemma \ref{lem:bhatia_davis} we have
    \begin{equation}
         \Var_{P_{sa}}\left[\Vhat^{(s,v)} - \Vhat^{(s,u)}\right] \leq \left( (v-u) - P_{sa}(\Vhat^{(s,v)} - \Vhat^{(s,u)}) \right) P_{sa}(\Vhat^{(s,v)} - \Vhat^{(s,u)}).  \label{eq:residual_properties_2}
    \end{equation}
    Next, notice that by AM-GM we have
    \begin{align*}
        \phi(v) - \phi(u) &= \frac{v - u}{\gamma} - P_{sa}(\Vhat^{(s,v)} - \Vhat^{(s,u)}) \\
        &= \frac{1}{\gamma}\left[\left((v-u) - P_{sa}(\Vhat^{(s,v)} - \Vhat^{(s,u)}) \right) + (1-\gamma) P_{sa}(\Vhat^{(s,v)} - \Vhat^{(s,u)}) \right] \\
        & \geq \frac{2}{\gamma} \sqrt{\left((v-u) - P_{sa}(\Vhat^{(s,v)} - \Vhat^{(s,u)}) \right)(1-\gamma) P_{sa}(\Vhat^{(s,v)} - \Vhat^{(s,u)})}.
    \end{align*}
    Combining with~\eqref{eq:residual_properties_2} we thus have that
    \begin{equation*}
        \Var_{P_{sa}}\left[\Vhat^{(s,v)} - \Vhat^{(s,u)}\right] \leq \frac{\gamma^2}{4(1-\gamma)} (\phi(v) - \phi(u))^2
    \end{equation*}
    as desired.
\end{proof}

\begin{lemma}
\label{lem:u_0_and_uhat_properties}
    \begin{enumerate}
        \item Letting $u_0 := \VabsP(s)$, then $u_0 \in [0, 1/(1-\gamma)]$, $\phi(u_0) = 0$, and $\Vhat^{(s, u_0)} = \VabsP$.
        \item Letting $\widehat{u} := \VabsPhat(s)$, then $\widehat{u} \in [0, 1/(1-\gamma)]$, $\phihat(\widehat{u}) = 0$, and $\Vhat^{(s, \widehat{u})} = \VabsPhat$.
    \end{enumerate}
\end{lemma}
\begin{proof}
    Clearly as an entry of a value function, we have $u_0 \in [0,1/(1-\gamma)]$. To show that $u_0$ satisfies $\phi(u_0) = 0$, it suffices to show that $\VabsP = \Vhat^{(s, u_0)}$, since then we will have
    \begin{equation*}
        \phi(u_0) = \frac{\VabsP(s) - r(sa)}{\gamma} - P_{sa} \Vhat^{(s, u_0)} = \frac{\VabsP(s) - r(sa)}{\gamma} - P_{sa} \VabsP = 0
    \end{equation*}
    where the final equality is the Bellman optimality equation for $\widetilde{\mathcal{M}}_{saP_{sa}}$ in state $s$. 
    By Lemma \ref{lem:absorbing_mdp_properties_1}, we have $\Vhat^{(s, u_0)}(s) = u_0 = \VabsP(s)$. For all $s' \neq s$, the Bellman optimality equations for the two MDPs $\widetilde{\mathcal{M}}_{saP_{sa}}$ and $\widehat{\mathcal{M}}_{s, u_0}$ corresponding to state $s'$ are identical, and so by uniqueness of solutions to discounted Bellman equations we have that $\VabsP = \Vhat^{(s, u_0)}$ as desired. We have thus also shown that $\phi(u_0) = 0$.

    The proof of the second statement of the lemma is completely analogous, with $\Phat_{sa}$, $\phihat$, $\widehat{u}$ in place of $P_{sa}$, $\phi$, $u_0$.
\end{proof}

Next, fix $\delta' \in (0,1)$ and fix the (conditionally measurable) quantity
\begin{equation*}
    B := 2 \left( \sqrt{\frac{2\Var_{P_{sa}}[\Vhat^{(s, u_0)}]\log(1/\delta')}{n}} + \frac{\log(1/\delta')}{3n(1-\gamma)} \right).
\end{equation*}
Define $u_-, u_+ \in [0, 1/(1-\gamma)]$ by
\begin{align*}
    u_- &:= \inf \{u \geq 0 : \phi(u) \geq -B\} \\
    u_+ &:= \sup \{u \leq 1/(1-\gamma) : \phi(u) \leq B\}.
\end{align*}
Note that this infimum and supremum are both finite and furthermore are both contained in $[0, 1/(1-\gamma)]$, because the sets each contain $u_0$ (since $\phi$ is strictly increasing and $\phi(u_0) = 0$), which was checked to be within $[0, 1/(1-\gamma)]$.
Intuitively, $u_-$ and $ u_+$ are designed so that $\phi(u_-) = -B, \phi(u_+) = B$, but if this would cause them to leave the interval $[0, 1/(1-\gamma)]$, then they are clipped.
The following lemma captures some immediate properties of $u_-$ and $u_+$.
\begin{lemma}
    If not $\phi(u_-) = -B$, then $0 \geq \phi(u_-) \geq -B$ and $u_- = 0$. Also, if not $\phi(u_+) = B$, then $0 \leq \phi(u_+) \leq B$ and $u_+ = \frac{1}{1-\gamma}$.
\end{lemma}
\begin{proof}
    From Lemma \ref{lem:u_0_and_uhat_properties}, $u_0 \in [0, 1/(1-\gamma)]$ and $\phi(u_0) = 0$, so $u_0 \in \{u \geq 0 : \phi(u) \geq -B\}$, which implies that $u_- \leq u_0 \leq 1/(1-\gamma)$ and thus, since $\phi$ is strictly increasing, that $\phi(u_-) \leq  0$.
    Since the function $\phi$ is continuous, we clearly have that $\phi(u_-) \geq -B$, and furthermore since it is strictly increasing, it is invertible on its range. Therefore either $-B \in \phi([0, u_0])$ in which case $u_- = \phi^{-1}(-B)$, or otherwise all elements of $\phi([0, u_0])$ are $\geq -B$ and then $u_- = 0$.

    The proofs for $u_+$ are completely analogous.
\end{proof}
Finally, also notice that
\begin{equation}
    \phihat(u_-) = \frac{u_- - r(s,a)}{\gamma} -P_{sa} \Vhat^{(s, u_-)} + (P_{sa} - \Phat_{sa}) \Vhat^{(s, u_-)}  = \phi(u_-) - (\Phat_{sa} - P_{sa}) \Vhat^{(s, u_-)} \label{eq:phi_phihat_u_-}
\end{equation}
and likewise
\begin{equation}
    \phihat(u_+) = \phi(u_+) - (\Phat_{sa} - P_{sa}) \Vhat^{(s, u_+)}. \label{eq:phi_phihat_u_+}
\end{equation}
By the union bound and Bernstein's inequality (Lemma \ref{lem:bernstein}), with probability at least $1 - 2\delta'$ (conditional on $(\Phat_{s'a'})_{(s'a') \neq (s,a)}$),
\begin{align}
    (\Phat_{sa} - P_{sa}) \Vhat^{(s, u_-)}
    &\geq - \sqrt{\frac{2 \Var_{P_{sa}} [\Vhat^{(s, u_-)}] \log(1/\delta')}{n}} - \frac{\spannorm{\Vhat^{(s, u_-)}} \log(1/\delta')}{3n}, \label{eq:u_-_bernstein_1}\\
    (\Phat_{sa} - P_{sa}) \Vhat^{(s, u_+)}
    &\leq  \sqrt{\frac{2 \Var_{P_{sa}} [\Vhat^{(s, u_+)}] \log(1/\delta')}{n}} + \frac{\spannorm{\Vhat^{(s, u_+)}} \log(1/\delta')}{3n}. \label{eq:u_+_bernstein_1}
\end{align}
Using Lemma \ref{lem:residual_properties} and the triangle inequality, we can further bound
\begin{align}
    \sqrt{\Var_{P_{sa}} [\Vhat^{(s, u_-)}]} &\leq \sqrt{\Var_{P_{sa}} [\Vhat^{(s, u_0)}] } + \sqrt{\Var_{P_{sa}} [\Vhat^{(s, u_-)} - \Vhat^{(s, u_0)}] } \nonumber\\
    &\leq \sqrt{\Var_{P_{sa}} [\Vhat^{(s, u_0)}]} + \sqrt{\frac{\gamma^2}{4(1-\gamma)}} B, \label{eq:u_-_variance_bd}
\end{align}
and similarly
\begin{equation}
    \sqrt{\Var_{P_{sa}} [\Vhat^{(s, u_+)}]}  \leq \sqrt{\Var_{P_{sa}} [\Vhat^{(s, u_0)}]} + \sqrt{\frac{\gamma^2}{4(1-\gamma)}} B. \label{eq:u_+_variance_bd}
\end{equation}
Under the above events, combining~\eqref{eq:u_-_bernstein_1} with~\eqref{eq:u_-_variance_bd} we can bound
\begin{align}
    (\Phat_{sa} - P_{sa}) \Vhat^{(s, u_-)}
    &\geq - \sqrt{\frac{2 \Var_{P_{sa}} [\Vhat^{(s, u_0)}] \log(1/\delta')}{n}} - \sqrt{\frac{2 \gamma^2 \log(1/\delta')}{4n(1-\gamma)}}B- \frac{ \log(1/\delta')}{3n(1-\gamma)} \nonumber \\
    & \geq - \sqrt{\frac{2 \Var_{P_{sa}} [\Vhat^{(s, u_0)}] \log(1/\delta')}{n}} - \frac{ \log(1/\delta')}{3n(1-\gamma)} - \frac{B}{2}  \nonumber \\
    &= -\frac{B}{2} - \frac{B}{2} = -B \label{eq:u_-_bernstein_2}
\end{align}
where the first inequality uses the fact that $\spannorm{\Vhat^{(s, u_-)}} \leq \frac{1}{1-\gamma}$, and then the second inequality makes an assumption that $n \geq \frac{2\gamma^2\log(1/\delta')}{1-\gamma}$.
Again, under the above event, we also similarly have
\begin{equation}
    (\Phat_{sa} - P_{sa}) \Vhat^{(s, u_+)}
    \leq B \label{eq:u_+_bernstein_2}
\end{equation}
by combining~\eqref{eq:u_+_bernstein_1} with~\eqref{eq:u_+_variance_bd}.

Now we will show that under these events we have $u_- \leq \widehat{u} \leq u_+$.
First we show that $u_- \leq \widehat{u}$.
If $\phi(u_-) = -B$, then combining with~\eqref{eq:phi_phihat_u_-} and~\eqref{eq:u_-_bernstein_2}, we have that
\begin{equation*}
    \phihat(u_-) = \phi(u_-)- (\Phat_{sa} - P_{sa})\Vhat^{(s, u_-)} \leq -B + B = 0.
\end{equation*}
Since $\phihat(\widehat{u}) = 0$, we have that $\phihat(u_-) \leq \phihat(\widehat{u})$. Since $\phihat$ is strictly increasing, we have that $u_- \leq \widehat{u}$.
For the other case that we do not have $\phi(u_-) = -B$, then $u_- = 0$, and therefore $u_- \leq \widehat{u} \in [0, 1/(1-\gamma)]$.

Next we use analogous arguments to show that $\widehat{u} \leq u_+$.
If $\phi(u_+) = B$ then
using~\eqref{eq:phi_phihat_u_+} and~\eqref{eq:u_+_bernstein_2} yields
\begin{equation*}
    \phihat(u_+) = \phi(u_+) -  (\Phat_{sa} - P_{sa})\Vhat^{(s, u_+)} \geq B - B = 0.
\end{equation*}
If it is not the case that $\phi(u_+) = B$, then $u_+ = \frac{1}{1-\gamma}$ and so still $\widehat{u} \leq u_+$.

We have now shown that
\begin{equation*}
    u_- \leq \widehat{u} \leq u_+.
\end{equation*}
Since $\phi$ is also strictly increasing, this further implies that
\begin{equation*}
    -B \leq \phi(u_-) \leq \phi(\widehat{u}) \leq \phi(u_+) \leq B.
\end{equation*}
Analogously to $u_-$ and $u_+$ we have
\begin{equation}
    \phihat(\widehat{u}) = \phi(\widehat{u}) - (\Phat_{sa} - P_{sa})\Vhat^{(s, \widehat{u})}, \label{eq:phi_uhat_identity}
\end{equation}
so by rearranging and using the fact that $\phihat(\widehat{u}) = 0$,
\begin{equation*}
    (\Phat_{sa} - P_{sa})\Vhat^{(s, \widehat{u})} = \phi(\widehat{u}) - \phihat(\widehat{u}) = \phi(\widehat{u}) \in [-B, B],
\end{equation*}
that is,
\begin{equation*}
    \left|(\Phat_{sa} - P_{sa})\Vhat^{(s, \widehat{u})} \right| \leq B = 2 \left( \sqrt{\frac{2\Var_{P_{sa}}[\Vhat^{(s, u_0)}]\log(1/\delta')}{n}} + \frac{\log(1/\delta')}{3n(1-\gamma)} \right).
\end{equation*}
Finally, using the triangle inequality and Lemma \ref{lem:residual_properties} to bound
\begin{equation*}
    \sqrt{\Var_{P_{sa}}[\Vhat^{(s, u_0)}]} \leq \sqrt{\Var_{P_{sa}}[\Vhat^{(s, \widehat{u})}]} + \sqrt{\Var_{P_{sa}}[\Vhat^{(s, \widehat{u})} - \Vhat^{(s, u_0)}]} \leq \sqrt{\Var_{P_{sa}}[\Vhat^{(s, \widehat{u})}]} + \sqrt{\frac{\gamma^2}{4(1-\gamma)}}B, 
\end{equation*}
we can bound
\begin{align*}
    B &= 2 \left( \sqrt{\frac{2\Var_{P_{sa}}[\Vhat^{(s, u_0)}]\log(1/\delta')}{n}} + \frac{\log(1/\delta')}{3n(1-\gamma)} \right) \\
    & \leq 2 \sqrt{\frac{2\Var_{P_{sa}}[\Vhat^{(s, \widehat{u})}]\log(1/\delta')}{n}} + \frac{2\log(1/\delta')}{3n(1-\gamma)} + 2 \sqrt{\frac{2\gamma^2 \log(1/\delta')}{4(1-\gamma)n}}B \\
    & \leq 2 \sqrt{\frac{2\Var_{P_{sa}}[\Vhat^{(s, \widehat{u})}]\log(1/\delta')}{n}} + \frac{2\log(1/\delta')}{3n(1-\gamma)} + \frac{1}{2}B
\end{align*}
using a requirement that $n \geq \frac{16\gamma\log(1/\delta')}{1-\gamma}$ for the second inequality. Rearranging and also combining with the previous bound, this implies that
\begin{align*}
    \left|(\Phat_{sa} - P_{sa})\Vhat^{(s, \widehat{u})} \right| \leq B \leq 2 \left( 2 \sqrt{\frac{2\Var_{P_{sa}}[\Vhat^{(s, \widehat{u})}]\log(1/\delta')}{n}} + \frac{2\log(1/\delta')}{3n(1-\gamma)} \right).
\end{align*}
Since $\Vhat^{(s, \widehat{u})} = \VabsPhat$ by Lemma \ref{lem:u_0_and_uhat_properties}, taking $\delta' = \delta/2$, and also using the tower property to bound the unconditional probability, this completes the proof of Lemma \ref{lem:active_case_empirical_bernstein}.

\section{Proofs of Main Results}

In this section we complete the proofs of Theorems \ref{thm:main_theorem_with_horizon} and \ref{thm:main_average_reward_reduction}.

We note that the $\theta$ appearing in the below theorem is a random quantity since it depends on $\spannorm{\Vhat^\star}$. Our subsequent results built upon Theorem \ref{thm:main_intermediate_horizon_version} will bound $\spannorm{\Vhat^\star}$, either almost surely or with high probability.
\begin{theorem}
\label{thm:main_intermediate_horizon_version}
    Fix $\delta \in (0,1)$ and a deterministic discounted-optimal policy $\pistar \in \PiSD$. Then with probability at least $1 -\delta$, for all stationary randomized policies $\pihat\in \PiSR$ satisfying
    \begin{equation*}
        \infnorm{\Vhat^\star - \Vhat^{\pihat}} \leq \frac{1}{(1-\gamma)n},
    \end{equation*}
    we have
    \begin{equation*}
        V^{\pihat} \geq V^\star - \theta \one \qquad\text{and}\qquad \infnorm{\Vhat^\star - V^\star} \leq \theta
    \end{equation*}
    where $\theta \leq \frac{1}{1-\gamma}$ and
    \begin{equation*}
        \theta \leq \sqrt{\frac{288 \log(8SA/\delta)}{(1-\gamma)^2 n}} \left(\sqrt{\spannorm{V^{\pistar}}} + \sqrt{\spannorm{\Vhat^\star}}\right) + \frac{64 \log(8SA/\delta)}{(1-\gamma)^2 n}.
    \end{equation*}
\end{theorem}
\begin{proof}
Fix $\delta' \in (0,1)$ and suppose that $n \geq 6$. Fix a deterministic optimal discounted policy $\pistar$. Applying Lemma \ref{lem:fixed_pol_empirical_bernstein} with $V = V^{\pistar}$, we obtain that with probability at least $1 - \delta'$,
\begin{equation}
    \left|(\Phat_{\pistar} - P_{{\pistar}})V^{\pistar} \right| \leq \sqrt{\frac{3\Var_{\Phat_{{\pistar}}}[V^{\pistar}]\log(4S/\delta')}{n}} + \frac{3\spannorm{V^{\pistar}}\log(4S/\delta')}{n}. \label{eq:completing_proof_true_bernstein_1}
\end{equation}
Using the above bound~\eqref{eq:completing_proof_true_bernstein_1} and Lemma \ref{lem:error_to_var_param} with $V = V^{\pistar}$, $M = \spannorm{V^{\pistar}}$, $\varepsilon=0$, and $\alpha = \alpha_1: = 3 \log(4S/\delta')$, we have under the same event that
\begin{equation}
    \infnorm{\Vhat^{\pistar} - V^{\pistar}} \leq \frac{\gamma \sqrt{\alpha_1}}{\sqrt{n}\sqrt{1-\gamma}} \sqrt{\infnorm{ (I - \gamma \Phat_{\pistar})^{-1} \Var_{\Phat_{\pistar}}[V^{\pistar}]} } + \gamma \frac{\alpha_1 \spannorm{V^{\pistar}}}{n(1-\gamma)}.\label{eq:completing_proof_true_bernstein_2}
\end{equation}
Furthermore, under the same event, using Lemma \ref{lem:var_lemma} with the same choices $V = V^{\pistar}$, $M = \spannorm{V^{\pistar}}$, $\varepsilon=0$, and $\alpha = \alpha_1$, we obtain
\begin{equation*}
    \infnorm{(I - \gamma \Phat_{\pistar})^{-1} \Var_{\Phat_{\pistar}}[V^{\pistar}]} \leq 8\frac{\spannorm{V^{\pistar}}^2 }{\gamma^2} \frac{\alpha_1}{n(1-\gamma)} + \frac{4\spannorm{V^{\pistar}} }{\gamma^2(1-\gamma)}
\end{equation*}
where we used that $\spannorm{r_\pi} \leq 1$. Further supposing that $n \geq \frac{16 \log(4S/\delta')}{1-\gamma}$ and using that $\spannorm{V^{\pistar}} \leq \frac{1}{1-\gamma}$, we can simplify this bound to
\begin{equation*}
    \infnorm{(I - \gamma \Phat_{\pistar})^{-1} \Var_{\Phat_{\pistar}}[V^{\pistar}]} \leq 8\frac{\spannorm{V^{\pistar}} }{\gamma^2} \frac{3}{16(1-\gamma)} + \frac{4\spannorm{V^{\pistar}} }{\gamma^2(1-\gamma)} \leq \frac{6\spannorm{V^{\pistar}} }{\gamma^2(1-\gamma)}.
\end{equation*}
Combining with~\eqref{eq:completing_proof_true_bernstein_2}, we obtain
\begin{align}
    \infnorm{\Vhat^{\pistar} - V^{\pistar}} &\leq \frac{\gamma \sqrt{\alpha_1}}{\sqrt{n}\sqrt{1-\gamma}} \sqrt{ \frac{6\spannorm{V^{\pistar}} }{\gamma^2(1-\gamma)} } + \gamma \frac{\alpha_1 \spannorm{V^{\pistar}}}{n(1-\gamma)} \nonumber\\
    &\leq \frac{1}{1-\gamma} \left( \sqrt{\frac{6 \alpha_1 \spannorm{V^{\pistar}}}{n}}  + \frac{\alpha_1\spannorm{V^{\pistar}}}{n}\right). \label{eq:opt_pol_err_bd}
\end{align}

Next, using Lemma \ref{lem:absorbing_bernstein_ineq} with failure probability $\delta'/SA$ for all $s \in \S, a \in \A$ and taking a union bound, if $n \geq \frac{16 \log(4SA/\delta')}{1-\gamma}$, then with additional failure probability at most $\delta'$ we have
\begin{equation*}
    \left|(\Phat_{sa} - P_{sa})\Vhat^\star \right| \leq \sqrt{\frac{32\Var_{P_{sa}}[\Vhat^\star]\log(4SA/\delta')}{n}} + \frac{32\log(4SA/\delta')}{(1-\gamma)n}
\end{equation*}
for all $s \in \S, a \in \A$. Hence, for any stationary randomized policy $\pihat \in \PiSR$, we have for any $s \in \S$ under this event that
\begin{align*}
    \left|(\Phat_{\pihat} - P_{\pihat})\Vhat^\star \right|(s)
    &\leq \sum_{a \in \A}\pihat(a \mid s)\left|(\Phat_{sa} - P_{sa})\Vhat^\star \right|\\
    & \leq \sum_{a \in \A}\pihat(a \mid s)\sqrt{\frac{32\Var_{P_{sa}}[\Vhat^\star]\log(4SA/\delta')}{n}} + \frac{32\log(4SA/\delta')}{(1-\gamma)n} \\
    & \leq \sqrt{\frac{32 \sum_{a \in \A}\pihat(a \mid s) \Var_{P_{sa}}[\Vhat^\star]\log(4SA/\delta')}{n}} + \frac{32\log(4SA/\delta')}{(1-\gamma)n} \\
    & \leq \sqrt{\frac{32  \Var_{P_{\pihat}(s, \cdot)}[\Vhat^\star]\log(4SA/\delta')}{n}} + \frac{32\log(4SA/\delta')}{(1-\gamma)n}
\end{align*}
where the penultimate step is by Jensen's inequality and the final step is because
\begin{align*}
    \Var_{P_{\pihat}(s, \cdot)}[\Vhat^\star] &= \sum_{a \in \A} \pihat(a \mid s) \left( P_{sa}(\Vhat^\star)^{\circ 2} - (P_{sa}\Vhat^\star)^2 \right) \\
    & \qquad\qquad + \sum_{a' \in \A} \pihat(a' \mid s) (P_{sa'}\Vhat^\star)^2 - \left(\sum_{a'' \in \A} \pihat(a'' \mid s) P_{sa''}\Vhat^\star\right)^2 \\
    & \geq \sum_{a \in \A} \pihat(a \mid s) \left( P_{sa}(\Vhat^\star)^{\circ 2} - (P_{sa}\Vhat^\star)^2 \right) \\
    &= \sum_{a \in \A}  \pihat(a \mid s) \Var_{P_{sa}}[\Vhat^\star],
\end{align*}
where the inequality is again by Jensen's inequality.
Hence we have shown the elementwise bound
\begin{equation}
    \left|(\Phat_{\pihat} - P_{\pihat})\Vhat^\star \right| \leq \sqrt{\frac{32\Var_{P_{\pihat}}[\Vhat^\star]\log(4SA/\delta')}{n}} + \frac{32\log(4SA/\delta')}{(1-\gamma)n}. \label{eq:empirical_pihat_concentration_event_bd}
\end{equation}

Since $\spannorm{\Vhat^\star} \leq \frac{1}{1-\gamma}$ and we assumed that $\infnorm{\Vhat^\star - \Vhat^{\pihat}} \leq \frac{1}{(1-\gamma)n}$, applying Lemma \ref{lem:error_to_var_param} with $M = \frac{1}{1-\gamma}$, $\varepsilon = \frac{1}{(1-\gamma)n}$, and $\alpha = \alpha_2 := 32\log(4SA/\delta')$ (and the roles of $\Phat$ and $P$ reversed), we obtain that
\begin{equation}
    \infnorm{\Vhat^{\pihat} - V^{\pihat}} \leq \frac{\gamma \sqrt{\alpha_2}}{\sqrt{n} \sqrt{1-\gamma}} \sqrt{ \infnorm{(I - \gamma P_{\pihat})^{-1} \Var_{P_{\pihat}} [\Vhat^\star] }} + \gamma \frac{\frac{2}{1-\gamma} + \frac{\alpha_2}{1-\gamma} }{n(1-\gamma)}. \label{eq:empirical_pol_err_bd_1}
\end{equation}
Next we again bound the variance parameter appearing in this bound.
By the assumption that $\infnorm{\Vhat^{\pihat} - \Vhat^\star} \leq \frac{1}{(1-\gamma)n}$, we have
\begin{align*}
    \infnorm{\Vhat^\star - r_{\pihat} - \gamma \Phat_{\pihat} \Vhat^\star} &= \infnorm{\Vhat^{\pihat} - r_{\pihat} - \gamma \Phat_{\pihat} \Vhat^{\pihat} + (I - \gamma \Phat_{\pihat})(\Vhat^\star - \Vhat^{\pihat})} 
    = \infnorm{(I - \gamma \Phat_{\pihat})(\Vhat^\star - \Vhat^{\pihat})} \\
    &\leq (1+\gamma) \infnorm{\Vhat^\star - \Vhat^{\pihat}} \leq \frac{2}{(1-\gamma)n} \leq \frac{1}{8}
\end{align*}
where the final inequality uses that $n \geq \frac{16}{1-\gamma}$.
Using~\eqref{eq:empirical_pihat_concentration_event_bd} again and applying Lemma \ref{lem:var_lemma} (again with $\Phat$ and $P$ reversed) with $\alpha = \alpha_2$, $M = \frac{1}{1-\gamma}$, $\varepsilon = 1/8$, and noting that $\Vhat^\star \geq \Vhat^{\pihat}$ and $\spannorm{r_{\pihat}} \leq 1$, we have
\begin{align}
    \infnorm{(I - \gamma P_{\pihat})^{-1} \Var_{P_{\pihat}} [\Vhat^\star] } &\leq 8 \frac{\spannorm{\Vhat^\star} \alpha_2}{\gamma^2 (1-\gamma)^2 n} + \frac{4 \spannorm{\Vhat^\star} (1 + 1/8)}{\gamma^2 (1-\gamma)} \nonumber\\
    & \leq \frac{8 \spannorm{\Vhat^\star}}{2 \gamma^2 (1-\gamma)} + \frac{4 \spannorm{\Vhat^\star} (1 + 1/8)}{\gamma^2 (1-\gamma)} \nonumber \\
    & \leq \frac{9 \spannorm{\Vhat^\star}}{\gamma^2 (1-\gamma)} \nonumber
\end{align}
where the second inequality uses that $n \geq \frac{64 \log(4SA/\delta')}{1-\gamma}$. Combining with~\eqref{eq:empirical_pol_err_bd_1}, we obtain
\begin{align}
    \infnorm{\Vhat^{\pihat} - V^{\pihat}} &\leq \frac{\gamma \sqrt{\alpha_2}}{\sqrt{n} \sqrt{1-\gamma}} \sqrt{ \frac{9 \spannorm{\Vhat^\star}}{\gamma^2 (1-\gamma)}} + \gamma \frac{\frac{2}{1-\gamma} + \frac{\alpha_2}{1-\gamma} }{n(1-\gamma)} \nonumber\\
    & \leq \frac{1}{1-\gamma} \left( \sqrt{\frac{9 \alpha_2 \spannorm{\Vhat^\star}}{n}} + \frac{2 + \alpha_2}{(1-\gamma)n} \right). \label{eq:empirical_pol_err_bd_2}
\end{align}

Finally, combining~\eqref{eq:opt_pol_err_bd},~\eqref{eq:empirical_pol_err_bd_2}, the fact that $\infnorm{\Vhat^{\pihat} - \Vhat^\star} \leq \frac{1}{(1-\gamma)n}$, and Lemma \ref{lem:deterministic_error_decomp}, we have that with probability at least $1 - 2 \delta'$,
\begin{align*}
    V^{\pihat} \geq V^\star - \theta \one \qquad \text{and} \qquad \infnorm{\Vhat^\star - V^\star} \leq \theta,
\end{align*}
where
\begin{align*}
    \theta &= \frac{1}{1-\gamma} \left( \sqrt{\frac{6 \alpha_1 \spannorm{V^{\pistar}}}{n}}  + \frac{\alpha_1\spannorm{V^{\pistar}}}{n} + \sqrt{\frac{9 \alpha_2 \spannorm{\Vhat^\star}}{n}} + \frac{2 + \alpha_2}{(1-\gamma)n} \right) + \frac{1}{(1-\gamma)n} \\
    & \leq \frac{1}{1-\gamma} \left( \sqrt{\frac{6 \alpha_1 \spannorm{V^{\pistar}}}{n}}  + \sqrt{\frac{9 \alpha_2 \spannorm{\Vhat^\star}}{n}} + \frac{3 + \alpha_1 + \alpha_2}{(1-\gamma)n} \right) \\
    & \leq \sqrt{\frac{288 \log(4SA/\delta')}{(1-\gamma)^2 n}} \left(\sqrt{\spannorm{V^{\pistar}}} + \sqrt{\spannorm{\Vhat^\star}}\right) + \frac{64 \log(4SA/\delta')}{(1-\gamma)^2 n}. 
\end{align*}
We conclude the proof by taking $\delta' = \delta/2$, and also noting that all the assumptions made in the proof lower-bounding the size of $n$ are implied by $n \geq \frac{64 \log(4SA/\delta')}{1-\gamma}$, and in the case that this condition does not hold then the upper bound on $\theta$ above is larger than $\frac{1}{1-\gamma}$ and so it is still true (since the discounted value functions take values in $[0, 1/(1-\gamma)]$), and thus we can drop this explicit requirement.
\end{proof}

We also note that the calculations leading to~\eqref{eq:opt_pol_err_bd} in the above proof hold for any stationary deterministic policy, leading to the following lemma.
\begin{lemma}
    \label{lem:policy_eval_standalone_discounted}
    Let $\pi$ be a stationary deterministic policy, fix $\delta \in (0,1)$, and suppose $n \geq \frac{16\log(4S/\delta)}{1-\gamma}$. Then with probability at least $1-\delta$, letting $\alpha = 3 \log(4S/\delta)$, we have
    \begin{equation*}
        \infnorm{\Vhat^\pi - V^\pi} \leq \frac{1}{1-\gamma} \left( \sqrt{\frac{6 \alpha \spannorm{V^{\pi}}}{n}}  + \frac{\alpha \spannorm{V^{\pi}}}{n}\right).
    \end{equation*}
\end{lemma}

\begin{proof}[Proof of Theorem \ref{thm:main_theorem_with_horizon}]
    We will make use of Theorem \ref{thm:main_intermediate_horizon_version}. Using the bounds $\spannorm{V^{\pistar}}, \spannorm{\Vhat^\star} \leq \frac{1}{1-\gamma}$, we can bound the quantity $\theta$ appearing in Theorem \ref{thm:main_intermediate_horizon_version} by
    \begin{align*}
        \theta &\leq \sqrt{\frac{288 \log(8SA/\delta)}{(1-\gamma)^2 n}} \left(\sqrt{\spannorm{V^{\pistar}}} + \sqrt{\spannorm{\Vhat^\star}}\right) + \frac{64 \log(8SA/\delta)}{(1-\gamma)^2 n} \\
        & \leq \sqrt{\frac{1152 \log(8SA/\delta)}{(1-\gamma)^3 n}} + \frac{64 \log(8SA/\delta)}{(1-\gamma)^2 n}.
    \end{align*}
    Furthermore, if $n < \frac{1152 \log(8SA/\delta)}{(1-\gamma)}$ then the first term of the above bound will be greater than $ \frac{1}{1-\gamma}$, and the bound $\theta \leq \frac{1}{1-\gamma}$ is always trivially true. In the case that $n \geq \frac{1152 \log(8SA/\delta)}{(1-\gamma)}$, we can then bound the second term above (by upper-bounding $1/\sqrt{n}$) as
    \begin{equation*}
        \frac{64 \log(8SA/\delta)}{(1-\gamma)^2 n} \leq \frac{64 \log(8SA/\delta)}{(1-\gamma)^2 \sqrt{n}}\sqrt{\frac{1-\gamma}{1152 \log(8SA/\delta)}} \leq 2 \sqrt{\frac{\log(8SA/\delta)}{(1-\gamma)^3n}}. 
    \end{equation*}
    We can combine this with the above bound and simplify constants to obtain that
    \begin{equation*}
        \theta \leq \sqrt{\frac{1152 \log(8SA/\delta)}{(1-\gamma)^3 n}} + 2 \sqrt{\frac{\log(8SA/\delta)}{(1-\gamma)^3n}} \leq 36 \sqrt{\frac{\log(8SA/\delta)}{(1-\gamma)^3n}}.
    \end{equation*}

    Finally, by Hoeffding's inequality (Lemma \ref{lem:hoeffding}) and a union bound, with additional failure probability at most $\delta$ we have that
    \begin{equation*}
        \left|(\Phat - P)V^\star \right| \leq \frac{\sqrt{\spannorm{V^\star}^2 \log(2SA/\delta)}}{\sqrt{2n}} \one.
    \end{equation*}
    Combining with Lemma \ref{lem:Q_est_from_V_est} and the bound $\infnorm{\Vhat^\star - V^\star} \leq \theta$, we have
    \begin{align*}
        \infnorm{\Qhat^\star - Q^\star}
        &\leq 36 \sqrt{\frac{\log(8SA/\delta)}{(1-\gamma)^3n}} + \sqrt{\frac{\spannorm{V^\star} \log(2SA/\delta)}{(1-\gamma)^2 2n}} \\
        & \leq 37 \sqrt{\frac{\log(8SA/\delta)}{(1-\gamma)^3n}}
    \end{align*}
    since $\spannorm{V^\star} \leq \frac{1}{1-\gamma}$. We obtain the desired conclusion by adjusting $\delta$.
\end{proof}

\begin{theorem}
\label{thm:main_first_span_based_bd}
    Fix $\delta \in (0,1)$. Then with probability at least $1 -\delta$, for all stationary randomized policies $\pihat \in \PiSR$ satisfying
    \begin{equation*}
        \infnorm{\Vhat^\star - \Vhat^{\pihat}} \leq \frac{1}{(1-\gamma)n},
    \end{equation*}
    we have
    \begin{equation*}
        V^{\pihat} \geq V^\star - \eta \one \qquad\text{and}\qquad \infnorm{\Vhat^\star - V^\star} \leq \eta
    \end{equation*}
    where
    \begin{equation*}
        \eta \leq \sqrt{\frac{4608 \log(8SA/\delta)}{(1-\gamma)^2 n}} \sqrt{\spannorm{V^{\star}}}  + \frac{704 \log(8SA/\delta)}{(1-\gamma)^2 n}. 
    \end{equation*}
\end{theorem}
\begin{proof}
    Under the event described in Theorem \ref{thm:main_intermediate_horizon_version} (which holds with probability at least $1 - \delta$), we have $\infnorm{\Vhat^\star - V^\star} \leq \theta$ and
    \begin{equation}
        \theta \leq \sqrt{\frac{288 \log(8SA/\delta)}{(1-\gamma)^2 n}} \left(\sqrt{\spannorm{V^{\pistar}}} + \sqrt{\spannorm{\Vhat^\star}}\right) + \frac{64 \log(8SA/\delta)}{(1-\gamma)^2 n}. \label{eq:main_first_span_based_bd_1}
    \end{equation}
    We can then further calculate using triangle inequality for $\spannorm{\cdot}$ and $\spannorm{\cdot} \leq 2\infnorm{\cdot}$ that
    \begin{equation*}
        \spannorm{\Vhat^\star} \leq \spannorm{V^\star} + \spannorm{\Vhat^\star - V^\star} \leq \spannorm{V^\star} + 2 \infnorm{\Vhat^\star - V^\star} \leq \spannorm{V^\star} + 2\theta.
    \end{equation*}
    Combining this bound with~\eqref{eq:main_first_span_based_bd_1} and using that $\sqrt{a+b} \leq \sqrt{a} + \sqrt{b}$, we have (still under the same event)
    \begin{align*}
        \theta
        &\leq \sqrt{\frac{288 \log(8SA/\delta)}{(1-\gamma)^2 n}} \left(\sqrt{\spannorm{V^{\pistar}}} + \sqrt{\spannorm{V^\star} + 2\theta}\right) + \frac{64 \log(8SA/\delta)}{(1-\gamma)^2 n} \\
        &\leq \sqrt{\frac{\alpha}{(1-\gamma)^2 n}} \left(2\sqrt{\spannorm{V^{\pistar}}} + \sqrt{2\theta}\right) + \frac{\beta}{(1-\gamma)^2 n}
    \end{align*}
    where we abbreviated $\alpha = 288 \log(8SA/\delta), \beta = 64 \log(8SA/\delta)$.
    We can rearrange this to
    \begin{equation*}
        \theta - \sqrt{\frac{2\alpha}{(1-\gamma)^2 n}} \sqrt{\theta} - 2\sqrt{\frac{\alpha\spannorm{V^\star}}{(1-\gamma)^2 n}} - \frac{\beta}{(1-\gamma)^2 n} \leq 0.
    \end{equation*}
    We can use the quadratic equation to bound
    \begin{equation*}
        \sqrt{\theta} \leq \frac{1}{2} \left( \sqrt{\frac{2\alpha}{(1-\gamma)^2 n}} + \sqrt{\frac{2\alpha}{(1-\gamma)^2 n} +8\sqrt{\frac{\alpha\spannorm{V^\star}}{(1-\gamma)^2 n}} + \frac{4\beta}{(1-\gamma)^2 n} }\right).
    \end{equation*}
    Squaring both sides and using that $\left( \frac{\sqrt{a} + \sqrt{b}}{2}\right)^2 = \frac{a + b + 2\sqrt{ab}}{4} \leq \frac{a+b}{2}$ by AM-GM, we have
    \begin{align*}
        \theta
        & \leq \frac{1}{2}\left(\frac{2\alpha}{(1-\gamma)^2 n} + \frac{2\alpha}{(1-\gamma)^2 n} +8\sqrt{\frac{\alpha\spannorm{V^\star}}{(1-\gamma)^2 n}} + \frac{4\beta}{(1-\gamma)^2 n}\right)\\
        &= 4\sqrt{\frac{\alpha\spannorm{V^\star}}{(1-\gamma)^2 n}} + 2\frac{\alpha + \beta}{(1-\gamma)^2n}.
    \end{align*}
    This concludes the proof.
\end{proof}

\begin{proof}[Proof of Theorem \ref{thm:main_average_reward_reduction}]
    We apply Theorem \ref{thm:main_first_span_based_bd} with the choice of $\gbar$ from the theorem statement. We operate under the good event of Theorem \ref{thm:main_first_span_based_bd}, which holds with probability at least $1 -  \delta$, and we let $\eta'$ be the quantity which is referred to as $\eta$ within the statement of Theorem \ref{thm:main_first_span_based_bd}.
    By the second part of Lemma \ref{lem:average_reward_facts} \citep{zurek_spanagnostic_2025a}, and since $P$ is weakly communicating, we have
    \begin{equation*}
        \rho^{\pihat} \geq \rho^\star - (1-\gbar)\infnorm{V_{\gbar}^{\pihat} - V_{\gbar}^\star}\one - (1-\gbar) \spannorm{V_{\gbar}^\star}\one \geq \rho^\star - (1-\gbar)\eta'\one - (1-\gbar)2 \spannorm{h^\star}\one,
    \end{equation*}
    where we used that $\spannorm{V_{\gbar}^\star} \leq 2 \spannorm{h^\star}$ from the first part of Lemma \ref{lem:average_reward_facts} \citep{wei_modelfree_2020}. We also have
    \begin{equation*}
        \infnorm{(1-\gbar)\Vhat_{\gbar}^\star - \rho^\star} \leq (1-\gbar)\infnorm{\Vhat_{\gbar}^\star - V_{\gbar}^\star} + \infnorm{\rho^\star - (1-\gbar)V_{\gbar}^\star} \leq (1-\gbar)\eta' + 2(1-\gbar) \spannorm{h^\star}
    \end{equation*}
    where the bound $\infnorm{\rho^\star - (1-\gbar)V_{\gbar}^\star} \leq (1-\gbar)\spannorm{V_{\gbar}^\star} \leq (1-\gbar) 2 \spannorm{h^\star}$ combines parts 1 and 3 of Lemma \ref{lem:average_reward_facts}.

    Therefore we may take
    \begin{align*}
        \eta
        &= (1-\gbar)\eta' + 2(1-\gbar) \spannorm{h^\star} \\
        & \leq (1-\gbar) \left( \sqrt{\frac{4608 \log(8SA/\delta)}{(1-\gbar)^2 n}} \sqrt{\spannorm{V_{\gbar}^{\star}}}  + \frac{704 \log(8SA/\delta)}{(1-\gbar)^2 n} \right) + 2(1-\gbar) \spannorm{h^\star} \\
        &= \sqrt{\frac{4608 \log(8SA/\delta)}{ n}} \sqrt{\spannorm{V_{\gbar}^{\star}}} + 704 \sqrt{\frac{H\log(8SA/\delta)}{n}} + 2 \sqrt{\frac{\spannorm{h^\star}^2\log(8SA/\delta)}{nH}} \\
        & \leq 802 \sqrt{\frac{H\log(8SA/\delta)}{n}}
    \end{align*}
    where in the final inequality we used that $\spannorm{V_{\gbar}^\star} \leq 2 \spannorm{h^\star}$ and that $\spannorm{h^\star}\leq H$, and simplified.
\end{proof}

\section{Diameter-Based Concentration Refinement}

The main result of this section is the following Bernstein-like concentration inequality for the empirical optimal value function $\Vhat^\star$, which crucially replaces the dependence on $\frac{1}{1-\gamma}$ with the diameter $D$.
\begin{lemma}
    \label{lem:absorbing_bernstein_ineq_diameter}
    Suppose $P$ is communicating and $\diam(P) \leq D$. Fix $s\in \S, a\in \A$ and let $\delta \in (0,1)$. Suppose $n \geq 2664 D\log(12S^2/\delta)$. Then with probability at least $1 - \delta$,
    \begin{equation*}
        \left|(\Phat_{sa} - P_{sa})\Vhat^\star \right| \leq \sqrt{\frac{18\Var_{P_{sa}}[\Vhat^\star]\log(6/\delta)}{n}} + \frac{84D\log(6/\delta)}{n}.
    \end{equation*}
\end{lemma}
Lemma \ref{lem:absorbing_bernstein_ineq_diameter} is finally proved in Subsection \ref{sec:diameter_concentration_completing_the_proof}.

\subsection{Empirical Diameter Bound}

The main result of this subsection is Lemma \ref{lem:empirical_diameter_bound}, which bounds the diameter of the empirical MDP with high probability under the same sample size conditions as the target theorem. This is a refinement of \cite[Lemma 4]{zurek_plugin_2025} which achieves the same result but under a worse requirement on the sample size. The proof technique is also the same and controls the empirical diameter in terms of the empirical discounted value functions for the auxiliary tasks of reaching all states, but our improved discounted MDP results are used to analyze the convergence of the discounted value functions.

\begin{definition}
\label{def:g_goal_mdps}
    Fix $g \in \S$. Define the $g$-goal empirical MDP $\widehat{\M}_{\to g} = (\S, \A, \Phat^{(g)}, r^{\to g})$ where, for any $s' \in \S, a' \in \A$,
    \begin{equation*}
        \Phat^{(g)}_{s'a'} = \begin{cases}
            \Phat_{s'a'} & s' \neq g \\
            e_{g}^\top & s' =g
        \end{cases},
        \qquad
        r^{\to g}(s',a') = \begin{cases}
            0 & s' \neq g \\
            1 & s' =g
        \end{cases}.
    \end{equation*}
    We define the abbreviation $\Vhat^\star_{\to g,\gamma} := V^\star_{\widehat{\M}_{\to g},\gamma}$ for the $\gamma$-discounted optimal value function of the $g$-goal empirical MDP.
    We also define the $g$-goal true MDP $\M_{\to g} = (\S, \A, P^{(g)}, r^{\to g})$ where, for any $s' \in \S, a' \in \A$,
    \begin{equation*}
        P^{(g)}_{s'a'} = \begin{cases}
            P_{s'a'} & s' \neq g \\
            e_{g}^\top & s' =g
        \end{cases}.
    \end{equation*}
    We also let $\pistar_{\to g} \in \PiSD$ be a policy satisfying $\max_{s_0 \neq g} \E^{P_{\pistar_{\to g}}}_{s_0}[\tau_{g}] \leq \diam(P)$, which exists by Lemma \ref{lem:diameter_equivalence}. We also define the notation $V^\star_{\to g,\gamma} := V^\star_{{\M}_{\to g},\gamma}$ and $V^{\pi}_{\to g,\gamma} := V^{\pi}_{{\M}_{\to g},\gamma}$ for the $\gamma$-discounted optimal and fixed-policy-$\pi$ value functions of the $g$-goal true MDP, respectively.
\end{definition}
We remark that $\Phat^{(g)}$ was already defined as part of Definition \ref{def:absorbing_mdp}, but we repeat it above for clarity.

\begin{lemma}
    \label{lem:value_to_diameter_bound}
    Fix $\gamma \in (0,1)$ and suppose that there exists some $\beta \in (0,1)$ such that, for all $g \in \S$, we have that (elementwise)
    \begin{equation*}
        \Vhat^\star_{\to g, \gamma} \geq \frac{\beta}{1-\gamma}\one.
    \end{equation*}
    Then
    \begin{equation*}
        \diam (\Phat) \leq \frac{\gamma}{1-\gamma}\frac{1-\beta}{\beta}.
    \end{equation*}
\end{lemma}
\begin{proof}
    Fix $g \in \S$, and let ${\pi_g}$ achieve the optimal value so that $\Vhat^\star_{\to g, \gamma} = (I - \gamma \Phat^{(g)}_{\pi_g})^{-1} r^{\to g}_{\pi_g}$ (where $\Phat^{(g)}_{\pi_g}$ is the transition matrix of policy ${\pi_g}$ in the MDP with transition kernel $\Phat^{(g)}$). Assuming without loss of generality that $g$ corresponds to the last element of the list of states, we can write
    \begin{equation*}
        \Phat^{(g)}_{\pi_g} = \begin{bmatrix}
            Q & q \\
            \zero_{1 \times S-1} & 1
        \end{bmatrix}
    \end{equation*}
    where $Q \in \R^{S-1 \times S-1}$ is a substochastic matrix recording the probabilities of transitions within $\S \setminus \{g\}$, and $q \in \R^{S-1 \times 1}$ is a column vector of the probabilities of transitions into $g$. We also note that $\Vhat^\star_{\to g, \gamma}(g) = \frac{1}{1-\gamma}$ since $g$ is absorbing and has reward $1$. Letting $V \in \R^{S-1}$ denote the restriction of $\Vhat^\star_{\to g, \gamma}$ to $\S \setminus \{g\}$, the restriction of the Bellman equation for $\Vhat^\star_{\to g, \gamma}$ yields for any $s \in \S \setminus \{g\}$ that
    \begin{equation*}
        V(s) = \Vhat^\star_{\to g, \gamma}(s) = r^{\to g}_{\pi_g}(s) + \gamma e_s^\top \Phat^{(g)}_{{\pi_g}} \Vhat^\star_{\to g, \gamma} = 0+ \gamma e_s^\top Q V + \gamma q(s) \Vhat^\star_{\to g, \gamma}(g) = \frac{\gamma}{1-\gamma}q(s) + \gamma e_s^\top Q V.
    \end{equation*}
    In vectorized notation this is $V = \frac{\gamma}{1-\gamma}q + \gamma Q V$, which is equivalent to $V = \frac{\gamma}{1-\gamma}(I - \gamma Q)^{-1} q$.
    Also from the fact that $\Phat^{(g)}_{\pi_g} \one = \one$, we have that $q = \one - Q \one$, and by rearranging we can also obtain that $(I - Q)^{-1}q = \one$. (We note that the fact that $I - Q$ is invertible follows from the fact that $\Vhat^\star_{\to g, \gamma} > \zero$ combined with standard arguments for finite Markov chains.)

    Now we can complete the core argument. Using $w := (I - \gamma Q)^{-1} \one$ as a Lyapunov function, we have
    \begin{align*}
        Qw - w &= Q (I - \gamma Q)^{-1} \one - (I - \gamma Q)^{-1} \one = (I - \gamma Q)^{-1}Q \one - (I - \gamma Q)^{-1} \one \\
        &= (I - \gamma Q)^{-1} (-q) = -\frac{1-\gamma}{\gamma}V
    \end{align*}
    where we used that $Q$ commutes with $(I - \gamma Q)^{-1}$ (which can be easily seen from the Neumann series), the fact that $q =  \one - Q \one$, and the identity $V = \frac{\gamma}{1-\gamma}(I - \gamma Q)^{-1} q$.
    By the assumption that $V \geq \frac{\beta}{1-\gamma}\one$ we then have
    \begin{equation*}
        Qw - w \leq -\frac{1-\gamma}{\gamma}\frac{\beta}{1-\gamma}\one = \frac{-\beta}{\gamma}\one.
    \end{equation*}
    Rearranging, using the nonnegativity of $(I - Q)^{-1}$, and the fact that $\E^{\Phat^{(g)}_{\pi_g}}_{s_0}[\hitt_g] = e_{s_0}^\top (I - Q)^{-1} \one$ for any $s_0 \in \S \setminus \{g\}$, we then have for any $s_0 \in \S \setminus \{g\}$ that
    \begin{equation*}
        \E^{\Phat^{(g)}_{\pi_g}}_{s_0}[\hitt_g] = e_{s_0}^\top (I - Q)^{-1} \one \leq e_{s_0}^\top (I - Q)^{-1} \frac{\gamma}{\beta}(I - Q)w = \frac{\gamma}{\beta}w(s_0).
    \end{equation*}
    Now it remains to bound $w$. We calculate for any $s_0 \in \S \setminus \{g\}$ that
    \begin{equation*}
        w(s_0) = e_{s_0}^\top(I - \gamma Q)^{-1} \one = e_{s_0}^\top(I - \gamma \Phat^{(g)}_{\pi_g})^{-1} (\one - e_g) = \frac{1}{1-\gamma} - \Vhat^\star_{\to g, \gamma}(s_0) \leq \frac{1-\beta}{1-\gamma}.
    \end{equation*}
    Combining the last two displays we have that
    \begin{equation*}
        \E^{\Phat_{\pi_g}}_{s_0}[\hitt_g] = \E^{\Phat^{(g)}_{\pi_g}}_{s_0}[\hitt_g] \leq \frac{\gamma}{\beta}w(s_0) \leq \frac{\gamma}{\beta}\frac{1-\beta}{1-\gamma},
    \end{equation*}
    where we also used the fact that $\E^{\Phat_{\pi_g}}_{s_0}[\hitt_g] = \E^{\Phat^{(g)}_{\pi_g}}_{s_0}[\hitt_g]$ (since the distribution of $\hitt_g$ is the same in $\Phat_{\pi_g}$ and $\Phat^{(g)}_{\pi_g}$).
    This holds uniformly over $s_0 \in \S \setminus \{g\}$ and also $g \in \S$, so we conclude that
    \begin{equation*}
        \diam (\Phat) = \max_{s_0, g \in \S} \inf_{\pi \in \PiSR} \E_{s_0}^{\Phat_\pi}[\hitt_g] \leq \max_{s_0, g \in \S} \E_{s_0}^{\Phat_{\pi_g}}[\hitt_g] \leq \frac{\gamma}{1-\gamma}\frac{1-\beta}{\beta}.
    \end{equation*}
\end{proof}

\begin{lemma}
\label{lem:empirical_diameter_bound}
    Fix $\delta \in (0,1)$ and suppose $P$ is communicating and has diameter $\diam(P) \leq D$. Then if $n \geq (444 \cdot 6) D \log(4S^2/\delta)$, with probability at least $1-\delta$ we have
    \begin{equation*}
        \diam(\Phat) \leq 2D.
    \end{equation*}
\end{lemma}
\begin{proof}
    First set $\gbar$ so that $\frac{1}{1-\gbar} = 6D$. In the rest of this proof $\gbar$ will be the only discount factor used, so we drop it from the value functions used here for notational simplicity.
    
    Now fix $s \in \S$.
    To apply Lemma \ref{lem:value_to_diameter_bound} we only need to lower-bound $\Vhat^\star_{\to s}$, which can be done by lower-bounding $\Vhat^{\pistar_{\to s}}_{\to s}$. To do so we will both upper-bound $\infnorm{\Vhat^{\pistar_{\to s}}_{\to s} - V^{\pistar_{\to s}}_{\to s}}$ and lower-bound $V^{\pistar_{\to s}}_{\to s}$.
    First, combining Lemma \ref{lem:val_fn_lower_bound_diameter} (which is taken from \citealt{zurek_plugin_2025}) with the bound on the time for $\pistar_{\to s}$ to hit $s$ from any $s_0$, we have
    \begin{equation}
        V^{\pistar_{\to s}}_{\to s}(s_0) \geq \frac{1}{1-\gbar} - \E_{s_0}^{P_{\pistar_{\to s}}} [\hitt_{s}] \geq \frac{1}{1-\gbar} - D = 5D. \label{eq:empirical_diameter_bound_1}
    \end{equation}
    
    Next we bound $\infnorm{\Vhat^{\pistar_{\to s}}_{\to s} - V^{\pistar_{\to s}}_{\to s}}$.
    We apply Lemma \ref{lem:policy_eval_standalone_discounted} with failure probability $\delta/S$ to obtain that, assuming $n \geq \frac{16 \log(4S^2/\delta)}{1-\gbar}$, we have with probability at least $1 - \delta/S$ that
    \begin{equation*}
        \infnorm{\Vhat^{\pistar_{\to s}}_{\to s} - V^{\pistar_{\to s}}_{\to s}} \leq \frac{1}{1-\gbar} \left( \sqrt{\frac{6\alpha \spannorm{V^{\pistar_{\to s}}_{\to s}}}{n}} + \frac{\alpha \spannorm{V^{\pistar_{\to s}}_{\to s}}}{n} \right)
    \end{equation*}
    where $\alpha = 3 \log(4S^2/\delta)$. Combining the trivial bound $V^{\pistar_{\to s}}_{\to s} \leq \frac{1}{1-\gbar}\one$ with the fact from Lemma \ref{lem:val_fn_lower_bound_diameter} that $V^{\pistar_{\to s}}_{\to s} \geq \left(\frac{1}{1-\gbar} - D\right) \one$, we have that $\spannorm{V^{\pistar_{\to s}}_{\to s}} \leq D$. Combining this with the above display, and further assuming that $n \geq \frac{444}{1-\gbar} \log(4S^2/\delta)$,
    we have
    \begin{align*}
        \infnorm{\Vhat^{\pistar_{\to s}}_{\to s} - V^{\pistar_{\to s}}_{\to s}} \leq 6D \left( \sqrt{\frac{6 D\alpha}{n}} + \frac{\alpha D}{n}\right) \leq D \left(\sqrt{\frac{6^2 \cdot 3 }{444}} + \frac{3 }{444} \right) < \frac{D}{2}.
    \end{align*}
    Finally, combining this with~\eqref{eq:empirical_diameter_bound_1}, we conclude that
    \begin{equation*}
        \Vhat^\star_{\to s} \geq \Vhat^{\pistar_{\to s}}_{\to s} \geq V^{\pistar_{\to s}}_{\to s} - \infnorm{\Vhat^{\pistar_{\to s}}_{\to s} - V^{\pistar_{\to s}}_{\to s}}\one \geq 5D\one - D/2\one = 4.5D\one.
    \end{equation*}

    Now, unfixing $s$ and applying this argument for all $s \in \S$ and taking a union bound, with probability at least $1 -\delta$ the above holds for all $s \in \S$ simultaneously, that is, with probability at least $1 - \delta$ we have for all $s \in \S$ that
    \begin{equation*}
        \Vhat^\star_{\to s} \geq 4.5D\one = \frac{1}{1-\gbar} \frac{3}{4} \one.
    \end{equation*}
    Therefore, applying Lemma \ref{lem:value_to_diameter_bound} with $\beta = 3/4$, we have
    \begin{equation*}
        \diam(\Phat) \leq \frac{\gbar}{1-\gbar} \frac{1-\beta}{\beta} \leq \frac{1}{1-\gbar} \frac{1-\beta}{\beta} = 6D \frac{1/4}{3/4} = 2D
    \end{equation*}
    as desired.
\end{proof}

\begin{lemma}
    \label{lem:diameter_active_case_diameter_bound_concentration_reduction}
    Fix $\delta \in (0,1)$ and $n > 0$, and suppose $P$ is communicating and $\diam(P) \leq D$. Also fix some $\alpha, \beta > 0.$ Suppose that
    \begin{equation}
        \P \left( \diam(\Phat) \leq 2D, \left\lvert(\Phat_{sa} - P_{sa})\Vhat^\star \right\rvert  > \sqrt{\frac{\alpha \Var_{P_{sa}} [\Vhat^\star]}{n}} + \frac{\beta D}{n} \right) \leq 2\delta.
    \end{equation}
    Further suppose that $n \geq (444 \cdot 6) D \log(4S^2/\delta)$. Then
    \begin{equation*}
        \P \left(\left\lvert(\Phat_{sa} - P_{sa})\Vhat^\star \right\rvert  > \sqrt{\frac{\alpha \Var_{P_{sa}} [\Vhat^\star]}{n}} + \frac{\beta D}{n} \right) \leq 3\delta.
    \end{equation*}
\end{lemma}
\begin{proof}
    First, for any two events $E, F$, we have that
    \begin{equation*}
        \P(E) = \P(E \text{ and }F) + \P(E \text{ and not }F) \leq \P(E \text{ and }F) + \P(\text{not }F).
    \end{equation*}
    Applying this with
    \begin{align*}
        E &= \left\{ \left\lvert(\Phat_{sa} - P_{sa})\Vhat^\star \right\rvert  > \sqrt{\frac{\alpha \Var_{P_{sa}} [\Vhat^\star]}{n}} + \frac{\beta D}{n} \right\} ,\\
        F &= \left\{\diam(\Phat) \leq 2D\right\},
    \end{align*}
    we obtain that
    \begin{multline*}
        \P \left( \left\lvert(\Phat_{sa} - P_{sa})\Vhat^\star \right\rvert  > \sqrt{\frac{\alpha \Var_{P_{sa}} [\Vhat^\star]}{n}} + \frac{\beta D}{n} \right) \\
        \leq \P \left( \diam(\Phat) \leq 2D, \left\lvert(\Phat_{sa} - P_{sa})\Vhat^\star \right\rvert  > \sqrt{\frac{\alpha \Var_{P_{sa}} [\Vhat^\star]}{n}} + \frac{\beta D}{n} \right) + \P\left( \diam(\Phat) > 2D \right).
    \end{multline*}
    The first of these probabilities is bounded by $2\delta$ by assumption and the second is bounded by $\delta$ by Lemma \ref{lem:empirical_diameter_bound}.
\end{proof}

\subsection{Active/Inactive Split and Inactive Case}
In this subsection we show that, similarly to the discounted case, the desired lemma can be obtained by separately analyzing the active and inactive MDPs $\VabsPhat$ and $\Vabsinact$, and then we also handle the inactive case. 
We let $\Fsa$ denote the $\sigma$-algebra generated by the random variables $(\Phat_{s'a'})_{s'a' \neq sa}$.
Also below $\inf_\pi$ denotes the infimum over stationary randomized policies $\inf_{ \pi \in \PiSR}$.

\begin{lemma}
\label{lem:diameter_case_active_inactive_split}
    Fix $\delta \in (0,1)$ and $n > 0$. Also fix some $\alpha, \beta > 0$.
    Suppose that
    \begin{equation}
        \P \left( \spannorm{\Vabsinact} \leq 2D, \left\lvert(\Phat_{sa} - P_{sa})\Vabsinact \right\rvert  > \sqrt{\frac{\alpha \Var_{P_{sa}} [\Vabsinact]}{n}} + \frac{\beta D}{n} \right) \leq \delta \label{eq:diameter_case_active_inactive_split_inactive_case}
    \end{equation}
    and
    \begin{equation}
        \P \left( \spannorm{\VabsPhat} \leq 2D, \inf_\pi \max_{s_0 \in \S} \E_{s_0}^{\Phat_{\pi}}[\hitt_{s}] \leq 2D, \left\lvert(\Phat_{sa} - P_{sa})\VabsPhat \right\rvert  > \sqrt{\frac{\alpha \Var_{P_{sa}} [\VabsPhat]}{n}} + \frac{\beta D}{n} \right) \leq \delta. \label{eq:diameter_case_active_inactive_split_active_case}
    \end{equation}
    Then
    \begin{equation*}
        \P \left( \diam(\Phat) \leq 2D, \left\lvert(\Phat_{sa} - P_{sa})\Vhat^\star \right\rvert  > \sqrt{\frac{\alpha \Var_{P_{sa}} [\Vhat^\star]}{n}} + \frac{\beta D}{n} \right) \leq 2\delta.
    \end{equation*}
\end{lemma}
\begin{proof}
    Within the below calculations we define $\text{$(s,a)$ active}$ to mean that $a$ is the unique element of $\argmax_{a' \in \A} r(s,a') + \gamma \Phat_{sa'} \Vhat^\star$, and we define $\text{$(s,a)$ inactive}$ to be the negation of this event. Also to avoid repetition we define, for any $V \in \R^S$, the event
    \begin{equation*}
        \Delta(V) := \left\{ \left\lvert(\Phat_{sa} - P_{sa})V \right\rvert  > \sqrt{\frac{\alpha \Var_{P_{sa}} [V]}{n}} + \frac{\beta D}{n} \right\}.
    \end{equation*}
    Then
    \begin{align*}
        \P \left( \diam(\Phat) \leq 2D, \Delta(\Vhat^\star) \right)
        &= \P \left( \diam(\Phat) \leq 2D, \Delta(\Vhat^\star), \text{$(s,a)$ active} \right) \\
        & \qquad + \P \left( \diam(\Phat) \leq 2D, \Delta(\Vhat^\star), \text{$(s,a)$ inactive} \right) \\
        &\leq \P \left( \inf_\pi \max_{s_0 \in \S}  \E_{s_0}^{\Phat_{\pi}}[\hitt_{s}] \leq 2D, \spannorm{\Vhat^\star} \leq 2D, \Delta(\Vhat^\star), \text{$(s,a)$ active} \right) \\
        & \qquad + \P \left( \spannorm{\Vhat^\star} \leq 2D, \Delta(\Vhat^\star), \text{$(s,a)$ inactive} \right) \\
        &= \P \left(  \inf_\pi \max_{s_0 \in \S} \E_{s_0}^{\Phat_{\pi}}[\hitt_{s}] \leq 2D, \spannorm{\VabsPhat} \leq 2D, \Delta(\VabsPhat), \text{$(s,a)$ active} \right) \\
        & \qquad + \P \left( \spannorm{\Vabsinact} \leq 2D, \Delta(\Vabsinact), \text{$(s,a)$ inactive} \right) \\
        &\leq \P \left( \inf_\pi \max_{s_0 \in \S}\E_{s_0}^{\Phat_{\pi}}[\hitt_{s}] \leq 2D, \spannorm{\VabsPhat} \leq 2D, \Delta(\VabsPhat) \right) \\
        & \qquad + \P \left( \spannorm{\Vabsinact} \leq 2D, \Delta(\Vabsinact)\right) \\
        & \leq 2\delta.
    \end{align*}
    The first inequality step used that $\diam(\Phat) \leq 2D$ implies that both $\inf_\pi \max_{s_0 \in \S}  \E_{s_0}^{\Phat_{\pi}}[\hitt_{s}] \leq 2D$ and $ \spannorm{\Vhat^\star} \leq 2D$.
    The next equality step uses Lemma \ref{lem:active_inactive_split}, which implies that $\Vhat^\star = \VabsPhat$ when $(s,a)$ is active and $\Vhat^\star = \Vabsinact$ when $(s,a)$ is inactive. The final inequalities then use the simple fact that dropping an event from a conjunction of events cannot decrease the probability, and then the assumptions.
\end{proof}

\begin{lemma}
\label{lem:diameter_inactive_case_main_result}
    Fix $n > 0, \delta \in (0,1)$. Let $\alpha = 2 \log (2/\delta)$.
    Then
    \begin{equation*}
        \P \left( \spannorm{\Vabsinact} \leq 2D, \left\lvert(\Phat_{sa} - P_{sa})\Vabsinact \right\rvert  > \sqrt{\frac{\alpha \Var_{P_{sa}} [\Vabsinact]}{n}} + \frac{\alpha D}{n} \right) \leq \delta.
    \end{equation*}
\end{lemma}
\begin{proof}
    $\Vabsinact$ is $\Fsa$-measurable, so by Bernstein's inequality (Lemma \ref{lem:bernstein}) we have
    \begin{equation*}
        \P \left( \left\lvert(\Phat_{sa} - P_{sa})\Vabsinact \right\rvert  > \sqrt{\frac{2 \Var_{P_{sa}} [\Vabsinact] \log(2/\delta)}{n}} + \frac{ \spannorm{\Vabsinact}\log(2/\delta)}{3n} \middle\vert \Fsa \right) \leq \delta.
    \end{equation*}
    By the $\Fsa$-measurability of $\Vabsinact$ we then have, abbreviating $\zeta = \frac{\log(2/\delta)}{n}$ in the next calculation, that
    \begin{align*}
        &\P \left( \spannorm{\Vabsinact} \!\leq 2D,\left\lvert(\Phat_{sa} - P_{sa})\Vabsinact \right\rvert  > \sqrt{2 \zeta \Var_{P_{sa}} [\Vabsinact]} + 2 D \zeta \middle\vert \Fsa \right) \\
        & \leq \P \left( \spannorm{\Vabsinact} \!\leq 2D,\left\lvert(\Phat_{sa} - P_{sa})\Vabsinact \right\rvert  > \sqrt{2 \zeta \Var_{P_{sa}} [\Vabsinact]} + \frac{\spannorm{\Vabsinact} \zeta}{3} \middle\vert \Fsa \right) \\
        &= \ind\left(\spannorm{\Vabsinact} \!\leq 2D\right)\P \left( \left\lvert(\Phat_{sa} - P_{sa})\Vabsinact \right\rvert  > \sqrt{2 \zeta \Var_{P_{sa}} [\Vabsinact]} + \frac{\spannorm{\Vabsinact} \zeta}{3} \middle\vert \Fsa \right) \\
        &\leq  \delta \ind\left(\spannorm{\Vabsinact} \!\leq 2D\right) \leq \delta.
    \end{align*}
    Then recalling that $\alpha = 2\log(2/\delta)$ and taking the expectation, by the tower property we have that
    \begin{equation*}
        \P \left( \spannorm{\Vabsinact} \leq 2D,\left\lvert(\Phat_{sa} - P_{sa})\Vabsinact \right\rvert  > \sqrt{\frac{\alpha \Var_{P_{sa}} [\Vabsinact]}{n}} + \frac{\alpha D}{n} \right) \leq \delta.
    \end{equation*}
\end{proof}

\subsection{Active Case}
The goal of this subsection is to show~\eqref{eq:diameter_case_active_inactive_split_active_case}, which we eventually accomplish in Lemma \ref{lem:diameter_active_case_main_result} below. First we develop several useful intermediate steps.
Throughout this subsection we fix some natural number $n$, $\delta' \in (0,1)$, and we let $\alpha = 18\log(1/\delta'), \beta = 84 \log(1/\delta')$.

\begin{lemma}
    \label{lem:variance_residual_bound_span_version}
    Supposing that $u,v \in [0, 1/(1-\gamma)]$ satisfy $u \leq v$, then
    \begin{equation*}
        \Var_{P_{sa}}[\Vhat^{(s,v)} - \Vhat^{(s,u)}] \leq |\phi(v) - \phi(u)|\left(\spannorm{\Vhat^{(s,v)}} + \spannorm{\Vhat^{(s,u)}}\right).
    \end{equation*}
\end{lemma}
\begin{proof}
    Using the Bhatia-Davis inequality (Lemma \ref{lem:bhatia_davis}) with the fact from~\eqref{eq:absorbing_MDP_diff_sandwich_bounds} that $\Vhat^{(s,v)} - \Vhat^{(s,u)} \leq (v-u)\one$, we have
    \begin{align}
        &\Var_{P_{sa}}[\Vhat^{(s,v)} - \Vhat^{(s,u)}]\nonumber \\
        &\leq \left( (v-u) - P_{sa}(\Vhat^{(s,v)} - \Vhat^{(s,u)}) \right)\left( P_{sa}(\Vhat^{(s,v)} - \Vhat^{(s,u)}) - \min_{s' \in \S} (\Vhat^{(s,v)} - \Vhat^{(s,u)})(s') \right). \label{eq:variance_residual_bound_span_version_1}
    \end{align}
    (Note that relative to the step~\eqref{eq:residual_properties_2}, we bound the second term more sharply instead of lower-bounding the minimum entry of $\Vhat^{(s,v)} - \Vhat^{(s,u)}$ by $0$.)
    For the first term of~\eqref{eq:variance_residual_bound_span_version_1}, we have
    \begin{equation*}
        (v-u) - P_{sa}(\Vhat^{(s,v)} - \Vhat^{(s,u)}) \leq \frac{v-u}{\gamma} - P_{sa}(\Vhat^{(s,v)} - \Vhat^{(s,u)}) = \phi(v) - \phi(u).
    \end{equation*}
    For the second term of~\eqref{eq:variance_residual_bound_span_version_1}, we can bound
    \begin{align*}
        P_{sa}(\Vhat^{(s,v)} - \Vhat^{(s,u)}) - \min_{s' \in \S} (\Vhat^{(s,v)} - \Vhat^{(s,u)})(s') &\leq \spannorm{\Vhat^{(s,v)} - \Vhat^{(s,u)}} \\
        & \leq \spannorm{\Vhat^{(s,v)}} + \spannorm{\Vhat^{(s,u)}}.
    \end{align*}
    Combining these bounds yields the desired statement.
\end{proof}

\begin{lemma}
    \label{lem:bounded_span_interval_monotone}
    Suppose that $u \in [0,1/(1-\gamma)] $ satisfies
    \begin{equation*}
        \Vhat^{(s, u)} \leq (u + 2D)\one \qquad \text{and}\qquad \min_{\pi\in \PiSD} \max_{s_0 \in \S} \E_{s_0}^{\Phat_{\pi}}[\hitt_{s}] \leq 2D.
    \end{equation*}
    Then for all $v \in [0,1/(1-\gamma)] $ such that $v \geq u$, we have $\spannorm{\Vhat^{(s,v)}} \leq 4D$.
\end{lemma}
\begin{proof}
    First we show that for all $v \geq u$ we have $\Vhat^{(s, v)} \leq (v + 2D)\one$.
    We calculate that
    \begin{equation*}
        \Vhat^{(s, v)} - v \one =  \Vhat^{(s, u)} - u \one + \left(\Vhat^{(s, v)} - \Vhat^{(s, u)} - (v-u) \one \right) \leq \Vhat^{(s, u)} - u \one \leq 2D\one
    \end{equation*}
    where the first inequality is from~\eqref{eq:absorbing_MDP_diff_sandwich_bounds} and the second inequality is by the assumption on $u$.

    Next, by Lemma \ref{lem:optimal_value_fn_difference_diameter} we have for any $v \in [0, 1/(1-\gamma)]$ and any $s' \in \S$ that
    \begin{equation*}
        \Vhat^{(s,v)}(s') \geq \Vhat^{(s,v)}(s) - \min_{\pi \in \PiSD} \E_{s'}^{\Phat^{(s)}_\pi}[\tau_s] = \Vhat^{(s,v)}(s) - \min_{\pi \in \PiSD} \E_{s'}^{\Phat_\pi}[\tau_s] \geq v - 2D
    \end{equation*}
    since $\Vhat^{(s,v)}(s) = v$ from Lemma \ref{lem:absorbing_mdp_properties_1} and by the assumption that $\min_{\pi\in \PiSD} \max_{s_0 \in \S} \E_{s_0}^{\Phat_{\pi}}[\hitt_{s}] \leq 2D$.
    
    Therefore we have shown that $(v - 2D)\one \leq \Vhat^{(s,v)} \leq (v + 2D)\one$, and thus that $\spannorm{\Vhat^{(s,v)}} \leq 4D$.
\end{proof}

\paragraph{A sufficient inequality}
First we reduce showing a bound of the form~\eqref{eq:diameter_case_active_inactive_split_active_case} to a more directly obtainable conditional probability bound.
For notational convenience define the event
\begin{equation*}
    \Ehit := \left \{ \inf_\pi \max_{s_0 \in \S} \E_{s_0}^{\Phat_{\pi}}[\hitt_{s}] \leq 2D \right\}.
\end{equation*}
Then
\begin{align}
    &\P \left( \spannorm{\VabsPhat} \leq 2D, \Ehit, \left\lvert(\Phat_{sa} - P_{sa})\VabsPhat \right\rvert  > \sqrt{\frac{\alpha \Var_{P_{sa}} [\VabsPhat]}{n}} + \frac{\beta D}{n} \right) \nonumber\\
    &= \E \P \left( \spannorm{\VabsPhat} \leq 2D, \Ehit, \left\lvert(\Phat_{sa} - P_{sa})\VabsPhat \right\rvert  > \sqrt{\frac{\alpha \Var_{P_{sa}} [\VabsPhat]}{n}} + \frac{\beta D}{n}\middle \vert \Fsa \right) \nonumber\\
    &= \E\left[\ind(\Ehit)\P \left( \spannorm{\VabsPhat} \leq 2D, \left\lvert(\Phat_{sa} - P_{sa})\VabsPhat \right\rvert  > \sqrt{\frac{\alpha \Var_{P_{sa}} [\VabsPhat]}{n}} + \frac{\beta D}{n}\middle \vert \Fsa \right) \right] \label{eq:active_case_conditional_prob_bound_reduction}
\end{align}
using the tower property and the $\Fsa$-measurability of $\Ehit$ (since furthermore $\ind(\Ehit)$ can be written as a function of $(\Phat_{s'a'})_{s' \neq s}$).
Thus it suffices to bound
\begin{equation}
    \P \left( \spannorm{\VabsPhat} \leq 2D, \left\lvert(\Phat_{sa} - P_{sa})\VabsPhat \right\rvert  > \sqrt{\frac{\alpha \Var_{P_{sa}} [\VabsPhat]}{n}} + \frac{\beta D}{n}\middle \vert \Fsa \right) \label{eq:active_case_sufficient_bound_main}
\end{equation}
on the $\Fsa$-measurable event $\Ehit$. From here on we will bound the above conditional probability~\eqref{eq:active_case_sufficient_bound_main} under several different cases of the realizations of $(\Phat_{s'a'})_{s'a' \neq sa}$.

\paragraph{Case 0: All $u \in [0,1/(1-\gamma)]$ have $ \spannorm{\Vhat^{(s,u)}} > 2D$.}
First, for the easy case that there does not exist any $u \in [0,1/(1-\gamma)]$ such that $\spannorm{\Vhat^{(s,u)}} \leq 2D$ (which is a $\Fsa$-measurable event as this is a function of $(\Phat_{s'a'})_{s' \neq s}$), we have that
\begin{equation*}
    \P \left( \spannorm{\VabsPhat} \leq 2D, \left\lvert(\Phat_{sa} - P_{sa})\VabsPhat \right\rvert  > \sqrt{\frac{\alpha \Var_{P_{sa}} [\VabsPhat]}{n}} + \frac{\beta D}{n}\middle \vert \Fsa \right) = 0
\end{equation*}
because, as shown in Lemma \ref{lem:u_0_and_uhat_properties}, $\VabsPhat = \Vhat^{(s, \widehat{u})}$, so $\spannorm{\VabsPhat} \leq 2D$ has conditional probability $0$.

\paragraph{Additional setup for cases 1-3}
Now we handle three remaining more nontrivial cases of the realization of $(\Phat_{s'a'})_{s'a' \neq sa}$. For all of these we will now utilize the following definitions:
\begin{align*}
    \vmins &:= \inf\{v \in [0, 1/(1-\gamma)] : \Vhat^{(s, v)} \leq (v + 2D)\one\}, \\
    v_0 &:= \max\{u_0, \vmins\},\\
    W &:= 2 \left( \sqrt{\frac{2\Var_{P_{sa}}[\Vhat^{(s, v_0)}]\log(1/\delta')}{n}} + \frac{10D\log(1/\delta')}{n} \right).
\end{align*}
Note that since we have already eliminated the case that there does not exist any $u \in [0,1/(1-\gamma)]$ such that $\spannorm{\Vhat^{(s,u)}} \leq 2D$, we are guaranteed that $\vmins < \infty$ and thus that $\vmins \in [0,1/(1-\gamma)]$. Since $u_0 \in [0, 1/(1-\gamma)]$ (from Lemma \ref{lem:u_0_and_uhat_properties}), this further implies that $v_0 \in [0, 1/(1-\gamma)]$.

\begin{lemma}
\label{lem:diameter_active_case_span_bounds}
    We have $\Vhat^{(s, \vmins)} \leq (\vmins + 2D)\one$.
    Furthermore, on the event $\Ehit$, $\spannorm{\Vhat^{(s,v)}}\leq 4D$ for all $v \geq \vmins$. In particular $\spannorm{\Vhat^{(s,v_0)}}, \spannorm{\Vhat^{(s,\vmins)}}\leq 4D$.
\end{lemma}
\begin{proof}
    The fact that $\Vhat^{(s, \vmins)} \leq (\vmins + 2D)\one$ follows immediately from the (Lipschitz) continuity of $v \mapsto \Vhat^{(s, v)}$.
    Then the fact that $\spannorm{\Vhat^{(s,v)}}\leq 4D$ for all $v \geq \vmins$ on the event $\Ehit$ follows immediately from Lemma \ref{lem:bounded_span_interval_monotone}. The final statement then follows from the fact that $v_0 \geq \vmins$ by definition of $v_0$ and obviously $\vmins \geq \vmins$.
\end{proof}

\paragraph{Case 1: $\phi(\vmins) > W$.}
We must have either $\vmins > \uhat$ or $\vmins \leq \uhat$. If $\vmins > \uhat$, then by definition of $\vmins$ there exists some $s' \in \S$ such that 
\begin{equation*}
    \Vhat^{(s, \uhat)}(s') > \uhat + 2D.
\end{equation*}
But since $\Vhat^{(s, \uhat)}(s) = \uhat$, this implies that $\spannorm{\Vhat^{(s, \uhat)}} > 2D$. Since $\Vhat^{(s, \uhat)} = \VabsPhat$ by Lemma \ref{lem:u_0_and_uhat_properties}, we then have that $\spannorm{\VabsPhat} > 2D$ in the case that $\vmins > \uhat$.

For the case that $\vmins \leq \uhat$, since $\phi(\vmins) > W$, $\phi$ is strictly increasing (Lemma \ref{lem:residual_properties}), and $\phi(\uhat) = (\Phat_{sa} - P_{sa})\Vhat^{(s, \uhat)}$ (from~\eqref{eq:phi_uhat_identity} and since $\phihat(\uhat)=0$), we have that
\begin{equation*}
    W < \phi(\vmins) \leq \phi(\uhat) = (\Phat_{sa} - P_{sa})\Vhat^{(s, \uhat)}.
\end{equation*}
Also since $\phihat$ is strictly increasing and $\vmins \leq \uhat$, we have $\phihat(\vmins) \leq \phihat(\uhat) = 0$. Therefore we have that $\phi(\vmins) - \phihat(\vmins) > W$, and we recall that by an identical calculation to~\eqref{eq:phi_phihat_u_-} we have
\begin{equation*}
    \phi(\vmins) - \phihat(\vmins) = (\Phat_{sa} - P_{sa})\Vhat^{(s, \vmins)}.
\end{equation*}
Now we can directly apply Bernstein's inequality (conditionally) to obtain that
\begin{align*}
    \P \left( (\Phat_{sa} - P_{sa})\Vhat^{(s, \vmins)} > \sqrt{\frac{2\Var_{P_{sa}}[\Vhat^{(s, \vmins)}] \log(1/\delta')}{n}} + \frac{\spannorm{\Vhat^{(s, \vmins)}} \log(1/\delta')}{3n} \middle\vert \Fsa \right) \leq \delta'.
\end{align*}
Furthermore we have that $\spannorm{\Vhat^{(s, \vmins)}} \leq 4D$ from Lemma \ref{lem:diameter_active_case_span_bounds}, and also since $\phi(\vmins) > W > 0 = \phi(u_0)$, we have $\vmins > u_0$, so $v_0 = \vmins$ in this case. Then
\begin{multline*}
    \sqrt{\frac{2\Var_{P_{sa}}[\Vhat^{(s, \vmins)}] \log(1/\delta')}{n}} + \frac{\spannorm{\Vhat^{(s, \vmins)}} \log(1/\delta')}{3n} \\
    \leq \sqrt{\frac{2\Var_{P_{sa}}[\Vhat^{(s, v_0)}] \log(1/\delta')}{n}} + \frac{4D \log(1/\delta')}{3n} \leq W,
\end{multline*}
so
\begin{equation*}
    \P \left( (\Phat_{sa} - P_{sa})\Vhat^{(s, \vmins)} > W \middle\vert \Fsa \right) \leq \delta'.
\end{equation*}

Therefore, in this case that $\phi(\vmins) > W$, we have shown that
\begin{equation*}
    \P\left( \spannorm{\VabsPhat} \leq 2D \middle \vert \Fsa \right) \leq \P\left( \vmins \leq \uhat \middle \vert \Fsa \right) \leq \P\left( (\Phat_{sa} - P_{sa})\Vhat^{(s, \vmins)} > W \middle \vert \Fsa \right) \leq \delta'.
\end{equation*}
This further implies that~\eqref{eq:active_case_sufficient_bound_main} is bounded by
\begin{multline*}
    \P \left( \spannorm{\VabsPhat} \leq 2D, \left\lvert(\Phat_{sa} - P_{sa})\VabsPhat \right\rvert  > \sqrt{\frac{\alpha \Var_{P_{sa}} [\VabsPhat]}{n}} + \frac{\beta D}{n}\middle \vert \Fsa \right) \\
     \leq \P\left( \spannorm{\VabsPhat} \leq 2D \middle \vert \Fsa \right) \leq \delta'
\end{multline*}
as desired.

\paragraph{A useful lemma}
Before considering the next cases, we present a calculation that will be used for both cases 2 and 3 as a standalone lemma. While the conclusion~\eqref{eq:active_case_cases2_3_phi_uhat_ineq_conclusion} of the following lemma is not immediately transparent, under each of the following two cases on the value of $\phi(\vmins)$ we will be able to easily relate it to a lower bound on $|\phi(\uhat)|$ in terms of $W$.
\begin{lemma}
\label{lem:active_case_cases2_3_phi_uhat_ineq}
    Suppose that $\Ehit$ holds. If
    \begin{equation*}
        \spannorm{\VabsPhat} \leq 2D \qquad \text{and} \qquad \left\lvert(\Phat_{sa} - P_{sa})\VabsPhat \right\rvert  > \sqrt{\frac{\alpha \Var_{P_{sa}} [\VabsPhat]}{n}} + \frac{\beta D}{n},
    \end{equation*}
    then
    \begin{equation}
        |\phi(\uhat)| > \frac{3}{2} W - \frac{|\phi(v_0) - \phi(\uhat)|}{2} .\label{eq:active_case_cases2_3_phi_uhat_ineq_conclusion}
    \end{equation}
\end{lemma}
\begin{proof}
    By Lemma \ref{lem:variance_residual_bound_span_version}, we have
\begin{align*}
    \sqrt{\Var_{P_{sa}}[\Vhat^{(s, v_0)} - \VabsPhat]} & \leq \sqrt{|\phi(v_0) - \phi(\uhat)|\left(\spannorm{\VabsPhat} + \spannorm{ \Vhat^{(s, v_0)} }\right)} \\
    & \leq \sqrt{|\phi(v_0) - \phi(\uhat)|6D}
\end{align*}
since $\spannorm{\VabsPhat} \leq 2D$ by assumption and $\spannorm{\Vhat^{(s, v_0)}} \leq 4D$ by Lemma \ref{lem:diameter_active_case_span_bounds} (since $\Ehit$ holds).
Therefore
\begin{align*}
    \sqrt{\frac{2\Var_{P_{sa}}[\Vhat^{(s, v_0)}]\log(1/\delta')}{n}}  
    &  \leq \sqrt{\frac{2\log(1/\delta')}{n}}\left(\sqrt{\Var_{P_{sa}}[\VabsPhat]} + \sqrt{\Var_{P_{sa}}[\Vhat^{(s, v_0)} - \VabsPhat]} \right) \\
    &  \leq \sqrt{\frac{2\Var_{P_{sa}}[\VabsPhat]\log(1/\delta')}{n}} + \sqrt{\frac{12|\phi(v_0) - \phi(\uhat)|D\log(1/\delta')}{n}}  \\
    &  \leq \sqrt{\frac{2\Var_{P_{sa}}[\VabsPhat]\log(1/\delta')}{n}} + \frac{|\phi(v_0) - \phi(\uhat)|}{6} + \frac{18 D \log(1/\delta')}{n}
\end{align*}
where we used AM-GM in the final inequality.
By rearranging, we have that
\begin{multline*}
    \sqrt{\frac{2\Var_{P_{sa}}[\VabsPhat]\log(1/\delta')}{n}} + \frac{28 D \log(1/\delta')}{n} \\
    \geq \sqrt{\frac{2\Var_{P_{sa}}[\Vhat^{(s, v_0)}]\log(1/\delta')}{n}} - \frac{|\phi(v_0) - \phi(\uhat)|}{6} + \frac{10D \log(1/\delta')}{n}
\end{multline*}
and thus
\begin{multline*}
    3 \left(\sqrt{\frac{2\Var_{P_{sa}}[\VabsPhat]\log(1/\delta')}{n}} + \frac{28 D \log(1/\delta')}{n}\right) \\
    \begin{aligned}
        &\geq 3 \left( \sqrt{\frac{2\Var_{P_{sa}}[\Vhat^{(s, v_0)}]\log(1/\delta')}{n}} - \frac{|\phi(v_0) - \phi(\uhat)|}{6} + \frac{10D \log(1/\delta')}{n} \right) \\
        & = \frac{3}{2}W - \frac{|\phi(v_0) - \phi(\uhat)|}{2}.
    \end{aligned}
\end{multline*}

Since $\phi(\uhat) = (\Phat_{sa} - P_{sa})\Vhat^{(s, \uhat)}$ and $\Vhat^{(s, \uhat)} = \VabsPhat$, combining the previous display with the assumptions of this lemma and recalling that $\alpha = 18\log(1/\delta'), \beta = 84 \log(1/\delta')$ (and that $18 = 3^2 \cdot 2, 84 = 3 \cdot 28$) yield that
\begin{align*}
    |\phi(\uhat)| &= \left| (\Phat_{sa} - P_{sa})\Vhat^{(s, \uhat)} \right| > 3 \left(\sqrt{\frac{2\Var_{P_{sa}}[\VabsPhat]\log(1/\delta')}{n}} + \frac{28 D \log(1/\delta')}{n}\right) \\
    &\geq \frac{3}{2}W - \frac{|\phi(v_0) - \phi(\uhat)|}{2}.
\end{align*}
\end{proof}

\paragraph{Case 2: $-W \leq \phi(\vmins) \leq W$.}
In this case we additionally define
\begin{equation}
    v_+ := \begin{cases}
        \phi^{-1}(W) & \text{if $W \in \phi([0, 1/(1-\gamma)])$} \\
        \frac{1}{1-\gamma} & \text{otherwise} \label{eq:v_plus_defn}
    \end{cases}.
\end{equation}
Identically to the previous case, if $\vmins > \uhat$ then by definition of $\vmins$ we would have that $\spannorm{\VabsPhat} > 2D$. 
Hence by contraposition $\spannorm{\VabsPhat} \leq 2D \implies \vmins \leq \uhat$.
Now we argue that
\begin{equation}
    |\phi(v_0) - \phi(\uhat)| \leq |\phi(\uhat)|. \label{eq:active_case_case_2_eq1}
\end{equation}
If $v_0 = u_0$, then $\phi(v_0) = \phi(u_0) = 0$ and so~\eqref{eq:active_case_case_2_eq1} holds with equality. If $v_0 = \vmins$, then $\vmins \geq u_0$, and combining with the inequality $\vmins \leq \uhat$ we have that $u_0 \leq \vmins \leq \uhat$, and so by monotonicity of $\phi$
\begin{equation*}
    0 = \phi(u_0) \leq \phi(\vmins) = \phi(v_0) \leq \phi(\uhat)
\end{equation*}
and hence
\begin{equation*}
    |\phi(v_0) - \phi(\uhat)| = \phi(\uhat) - \phi(v_0) \leq \phi(\uhat) \leq |\phi(\uhat)|.
\end{equation*}
Thus we have established~\eqref{eq:active_case_case_2_eq1}.
Now if the conclusion~\eqref{eq:active_case_cases2_3_phi_uhat_ineq_conclusion} of Lemma \ref{lem:active_case_cases2_3_phi_uhat_ineq} holds, then combining it with~\eqref{eq:active_case_case_2_eq1} would yield that
\begin{equation*}
    |\phi(\uhat)| > \frac{3}{2}W - \frac{|\phi(\uhat)|}{2}
\end{equation*}
and thus that
\begin{equation}
    |\phi(\uhat)| > W. \label{eq:active_case_case_2_eq2}
\end{equation}
Using the fact that in this case $\phi(\vmins) \geq -W$, the inequality $\vmins \leq \uhat$ also implies that $\phi(\uhat) \geq \phi(\vmins) \geq -W$. Because of this, we can further deduce from~\eqref{eq:active_case_case_2_eq2} that $\phi(\uhat) > W$.
Thus all of these deductions allow us to bound the probability~\eqref{eq:active_case_sufficient_bound_main} as
\begin{equation}
    \P \left( \spannorm{\VabsPhat} \leq 2D, \left\lvert(\Phat_{sa} - P_{sa})\VabsPhat \right\rvert  > \sqrt{\frac{\alpha \Var_{P_{sa}} [\VabsPhat]}{n}} + \frac{\beta D}{n}\middle \vert \Fsa \right) \leq \P \left( \phi(\uhat) > W \middle \vert \Fsa \right). \label{eq:active_case_case_2_eq3}
\end{equation}

If we have $v_+ = \frac{1}{1-\gamma}$ then by construction of $v_+$ we must have $\phi(v_+) \leq W$, but then since $v_+ = \frac{1}{1-\gamma}$ by the strict increase of $\phi$ we have that $\phi(\uhat) \leq \phi(\frac{1}{1-\gamma}) \leq W$ (since $\uhat \leq \frac{1}{1-\gamma}$, as shown in Lemma \ref{lem:u_0_and_uhat_properties}). Hence if $v_+ = \frac{1}{1-\gamma}$ then the probability on the RHS of~\eqref{eq:active_case_case_2_eq3} is equal to $0$.

Otherwise, if $v_+ \neq \frac{1}{1-\gamma}$ then $v_+$ satisfies $\phi(v_+) = W$. Then if $\phi(\uhat) > W$, the strict increase of $\phi$ implies that $\uhat > v_+$, which then implies that $0 = \phihat(\uhat) > \phihat(v_+)$ (since $\phihat$ is strictly increasing and $\phihat(\uhat) = 0$), and so this finally implies that $\phi(v_+) - \phihat(v_+) > W$. Therefore we can further bound the probability on the RHS of~\eqref{eq:active_case_case_2_eq3} as
\begin{equation}
    \P \left( \phi(\uhat) > W \middle \vert \Fsa \right) \leq \P \left( \phi(v_+) - \phihat(v_+) > W  \middle \vert \Fsa \right) = \P \left( ( \Phat_{sa} - P_{sa}) \Vhat^{(s, v_+)} > W  \middle \vert \Fsa \right)\label{eq:active_case_case_2_eq4}
\end{equation}
(using that $\phi(v_+) - \phihat(v_+) = ( \Phat_{sa} - P_{sa}) \Vhat^{(s, v_+)}$ by an identical calculation to~\eqref{eq:phi_phihat_u_-}).
Now we finally argue that this can be directly bounded by $\delta'$ using Bernstein's inequality conditionally.

First, by Bernstein's inequality we have that
\begin{equation}
    \P \left( ( \Phat_{sa} - P_{sa}) \Vhat^{(s, v_+)} > \sqrt{\frac{2 \Var_{P_{sa}}[\Vhat^{(s, v_+)}]\log(1/\delta')}{n}} + \frac{\spannorm{\Vhat^{(s, v_+)}} \log(1/\delta')}{3n}  \middle \vert \Fsa \right) \leq \delta'. \label{eq:active_case_case_2_eq5}
\end{equation}
Since $v_+ \geq \vmins$, by Lemma \ref{lem:diameter_active_case_span_bounds} we have that $\spannorm{\Vhat^{(s, v_+)}} \leq 4D$. Also by Lemma \ref{lem:variance_residual_bound_span_version}, the fact that $\spannorm{\Vhat^{(s, v_0)}} \leq 4D$ (also from Lemma \ref{lem:diameter_active_case_span_bounds}), and the fact that $|\phi(v_+) -\phi(v_0)| = \phi(v_+) - \phi(v_0) \leq W$ (since $v_+ \geq v_0$, $\phi(v_+) \leq W$, and $\phi(v_0) \geq \phi(u_0) \geq 0$), we have that
\begin{align*}
    \Var_{P_{sa}}[\Vhat^{(s,v_+)} - \Vhat^{(s,v_0)}] &\leq |\phi(v_+) - \phi(v_0)|\left(\spannorm{\Vhat^{(s,v_+)}} + \spannorm{\Vhat^{(s,v_0)}}\right) \leq 8DW.
\end{align*}
Consequently
\begin{align}
    &\sqrt{\frac{2 \Var_{P_{sa}}[\Vhat^{(s, v_+)}]\log(1/\delta')}{n}} + \frac{\spannorm{\Vhat^{(s, v_+)}} \log(1/\delta')}{3n} \nonumber\\
    & \leq \sqrt{\frac{2 \Var_{P_{sa}}[\Vhat^{(s, v_0)}]\log(1/\delta')}{n}} + \sqrt{\frac{2 \Var_{P_{sa}}[\Vhat^{(s, v_+)} - \Vhat^{(s, v_0)}]\log(1/\delta')}{n}} + \frac{4D \log(1/\delta')}{3n} \nonumber\\
    & \leq \sqrt{\frac{2 \Var_{P_{sa}}[\Vhat^{(s, v_0)}]\log(1/\delta')}{n}} + \sqrt{\frac{16DW\log(1/\delta')}{n}} + \frac{4D \log(1/\delta')}{3n} \nonumber\\
    & \leq \sqrt{\frac{2 \Var_{P_{sa}}[\Vhat^{(s, v_0)}]\log(1/\delta')}{n}} + \frac{W}{2} + \frac{8D \log(1/\delta')}{n} + \frac{4D \log(1/\delta')}{3n} \nonumber\\
    & \leq W, \label{eq:active_case_case_2_variance_bound_calculation}
\end{align}
using AM-GM for the penultimate inequality.
Combining with~\eqref{eq:active_case_case_2_eq5} we have
\begin{equation*}
    \P \left( ( \Phat_{sa} - P_{sa}) \Vhat^{(s, v_+)} > W  \middle \vert \Fsa \right) \leq \delta'
\end{equation*}
as desired. Recalling~\eqref{eq:active_case_case_2_eq3} and~\eqref{eq:active_case_case_2_eq4}, we have thus shown that 
\begin{equation*}
    \P \left( \spannorm{\VabsPhat} \leq 2D, \left\lvert(\Phat_{sa} - P_{sa})\VabsPhat \right\rvert  > \sqrt{\frac{\alpha \Var_{P_{sa}} [\VabsPhat]}{n}} + \frac{\beta D}{n}\middle \vert \Fsa \right) \leq \delta'.
\end{equation*}

\paragraph{Case 3: $\phi(\vmins) < -W$.}
We make the same choice~\eqref{eq:v_plus_defn} of $v_+$ as in the previous case.
Since $\phi(\vmins) < -W < 0 = \phi(u_0)$, we have $\vmins < u_0$, and so also $v_0 = \max\{\vmins, u_0\} = u_0$.
Additionally, since $\vmins \in [0, 1/(1-\gamma)]$, $\phi$ is strictly increasing, $\phi(\vmins) < -W$, and $\phi(u_0)= 0$ and $u_0 \in [0, 1/(1-\gamma)]$, we know that $\phi^{-1}(-W) \in [\vmins, u_0] \subseteq [0, 1/(1-\gamma)]$, and so we can additionally define $v_- := \phi^{-1}(-W)$.

Given these facts, we have
\begin{equation*}
    |\phi(v_0) - \phi(\uhat)| = |\phi(u_0) - \phi(\uhat)| = |\phi(\uhat)|,
\end{equation*}
and so by Lemma \ref{lem:active_case_cases2_3_phi_uhat_ineq} and then using this fact to simplify its conclusion, we have that
\begin{multline}
    \P \left( \spannorm{\VabsPhat} \leq 2D, \left\lvert(\Phat_{sa} - P_{sa})\VabsPhat \right\rvert  > \sqrt{\frac{\alpha \Var_{P_{sa}} [\VabsPhat]}{n}} + \frac{\beta D}{n}\middle \vert \Fsa \right) \\
    \leq \P \left( |\phi(\uhat)| > \frac{3}{2}W - \frac{|\phi(\uhat)|}{2} \middle \vert \Fsa \right) = \P \left( |\phi(\uhat)| > W \middle \vert \Fsa \right). \label{eq:active_case_case_3_eq-1}
\end{multline}
Now clearly if $|\phi(\uhat)| > W$, either $\phi(\uhat) > W$ or $\phi(\uhat) < -W$.
Actually all of our calculations from the previous case still apply to yield that
\begin{equation}
    \P \left( \phi(\uhat) > W \middle \vert \Fsa \right) \leq \delta'. \label{eq:active_case_case_3_eq-0.5}
\end{equation}
So by the union bound it remains to also bound the conditional probability of $\phi(\uhat) <  -W$.

If $\phi(\uhat) < -W$, then since $\phi(v_-) = -W$, the strict increase of $\phi$ implies that $\uhat < v_-$. Then we have $0 = \phihat(\uhat) < \phihat(v_-) $ by the strict increase of $\phihat$, and so we have that $\phihat(v_-) - \phi(v_-) > 0 - (-W) = W$.
This implication implies
\begin{equation}
    \P \left( \phi(\uhat) <  -W \middle \vert \Fsa \right) \leq \P \left( \phihat(v_-) - \phi(v_-) > W \middle \vert \Fsa \right). \label{eq:active_case_case_3_eq0}
\end{equation}
Now we can bound the RHS analogously to the previous case. We have that $\phihat(v_-) - \phi(v_-) = - (\Phat_{sa} - P_{sa})\Vhat^{(s, v_-)}$, and by Bernstein's inequality
\begin{equation}
    \P \left( ( \Phat_{sa} - P_{sa}) \Vhat^{(s, v_-)} < -\sqrt{\frac{2 \Var_{P_{sa}}[\Vhat^{(s, v_-)}]\log(1/\delta')}{n}} - \frac{\spannorm{\Vhat^{(s, v_-)}} \log(1/\delta')}{3n}  \middle \vert \Fsa \right) \leq \delta'. \label{eq:active_case_case_3_eq1}
\end{equation}
Since we have already argued that $\vmins \leq v_-$, we have by Lemma \ref{lem:diameter_active_case_span_bounds} that $\spannorm{\Vhat^{(s, v_-)}} \leq 4D$, and this lemma also yields that $\spannorm{\Vhat^{(s, v_0)}} \leq 4D$. Then by Lemma \ref{lem:variance_residual_bound_span_version} and the fact that $|\phi(v_0) - \phi(v_-)| = |\phi(u_0) - \phi(v_-)| = W$, we have that
\begin{equation*}
    \Var_{P_{sa}}[\Vhat^{(s,v_0)} - \Vhat^{(s,v_-)}] \leq |\phi(v_0) - \phi(v_-)|\left(\spannorm{\Vhat^{(s,v_0)}} + \spannorm{\Vhat^{(s,v_-)}}\right) \leq 8DW.
\end{equation*}
Therefore analogously to the calculation~\eqref{eq:active_case_case_2_variance_bound_calculation} from the previous case we have
\begin{equation*}
    \sqrt{\frac{2 \Var_{P_{sa}}[\Vhat^{(s, v_-)}]\log(1/\delta')}{n}} + \frac{\spannorm{\Vhat^{(s, v_-)}} \log(1/\delta')}{3n} \leq W
\end{equation*}
and so, using~\eqref{eq:active_case_case_3_eq1}, we have
\begin{align*}
    &\P\left(\phi(v_-) - \phihat(v_-) < -W \middle \vert \Fsa \right) \\
    &\leq \P \left( ( \Phat_{sa} - P_{sa}) \Vhat^{(s, v_-)} < -\sqrt{\frac{2 \Var_{P_{sa}}[\Vhat^{(s, v_-)}]\log(1/\delta')}{n}} - \frac{\spannorm{\Vhat^{(s, v_-)}} \log(1/\delta')}{3n}  \middle \vert \Fsa \right) \leq \delta'.
\end{align*}
Combining with~\eqref{eq:active_case_case_3_eq0} we thus have shown that
\begin{equation*}
    \P \left( \phi(\uhat) <  -W \middle \vert \Fsa \right) \leq \delta',
\end{equation*}
and so by~\eqref{eq:active_case_case_3_eq-0.5}, the union bound, and recalling~\eqref{eq:active_case_case_3_eq-1}, we have shown in this case that
\begin{multline*}
    \P \left( \spannorm{\VabsPhat} \leq 2D, \left\lvert(\Phat_{sa} - P_{sa})\VabsPhat \right\rvert  > \sqrt{\frac{\alpha \Var_{P_{sa}} [\VabsPhat]}{n}} + \frac{\beta D}{n}\middle \vert \Fsa \right) \\
    \leq \P \left( |\phi(\uhat)| > W \middle \vert \Fsa \right) \leq \P \left( \phi(\uhat) <  -W \middle \vert \Fsa \right) + \P \left( \phi(\uhat) > W \middle \vert \Fsa \right) \leq 2\delta'
\end{multline*}
as desired.

Having covered all cases of the realization of $(\Phat_{s'a'})_{s'a' \neq sa}$ (assuming $\Ehit$ holds) and bounded~\eqref{eq:active_case_sufficient_bound_main} as
\begin{equation*}
    \P \left( \spannorm{\VabsPhat} \leq 2D, \left\lvert(\Phat_{sa} - P_{sa})\VabsPhat \right\rvert  > \sqrt{\frac{\alpha \Var_{P_{sa}} [\VabsPhat]}{n}} + \frac{\beta D}{n}\middle \vert \Fsa \right) \leq 2\delta',
\end{equation*}
by the equality~\eqref{eq:active_case_conditional_prob_bound_reduction} we have thus shown that
\begin{equation*}
    \P \left( \spannorm{\VabsPhat} \leq 2D, \Ehit, \left\lvert(\Phat_{sa} - P_{sa})\VabsPhat \right\rvert  > \sqrt{\frac{\alpha \Var_{P_{sa}} [\VabsPhat]}{n}} + \frac{\beta D}{n} \right) \leq 2\delta'.
\end{equation*}
Taking $\delta' = \delta/2$ yields the below conclusion.

\begin{lemma}
    \label{lem:diameter_active_case_main_result}
    Fix $n > 0, \delta \in (0,1)$ and let $\alpha = 18 \log(2/\delta), \beta = 84\log(2/\delta)$. Then
    \begin{equation*}
        \P \left( \spannorm{\VabsPhat} \leq 2D, \Ehit, \left\lvert(\Phat_{sa} - P_{sa})\VabsPhat \right\rvert  > \sqrt{\frac{\alpha \Var_{P_{sa}} [\VabsPhat]}{n}} + \frac{\beta D}{n} \right) \leq \delta.
    \end{equation*}
\end{lemma}

\subsection{Proof of Lemma \ref{lem:absorbing_bernstein_ineq_diameter}}
\label{sec:diameter_concentration_completing_the_proof}

\begin{proof}[Proof of Lemma \ref{lem:absorbing_bernstein_ineq_diameter}]
    For arbitrary $\delta \in (0,1), n > 0$, Lemmas \ref{lem:diameter_inactive_case_main_result} and \ref{lem:diameter_active_case_main_result} yield the conditions of Lemma \ref{lem:diameter_case_active_inactive_split} with $\alpha = 18\log(2/\delta), \beta = 84\log(2/\delta)$ (since $2 \log(2/\delta) \leq 18 \log(2/\delta)$). Under the further assumption that $n \geq 2664 D \log(4S^2/\delta) = (444 \cdot 6) D \log(4S^2/\delta)$, Lemma \ref{lem:diameter_active_case_diameter_bound_concentration_reduction} then yields that
    \begin{equation*}
            \P \left(\left\lvert(\Phat_{sa} - P_{sa})\Vhat^\star \right\rvert  > \sqrt{\frac{\alpha \Var_{P_{sa}} [\Vhat^\star]}{n}} + \frac{\beta D}{n} \right) \leq 3\delta.
        \end{equation*}
    Adjusting the failure probability then completes the proof.
\end{proof}

\subsection{Proofs of Main Results for Average-Reward Plug-In Approach}
\begin{theorem}
    \label{thm:high_discount_plug_in}
    Suppose $P$ is communicating and $\diam(P) \leq D$.
    Fix $\delta \in (0,1)$.
    Then with probability at least $1-\delta$,
    for all deterministic policies $\pihat\in \PiSD$ satisfying
    \begin{equation*}
        \infnorm{\Vhat^\star - \Vhat^{\pihat}} \leq \frac{1}{n},
    \end{equation*}
    we have
    \begin{equation*}
        V^{\pihat} \geq V^\star - \theta \one \qquad\text{and}\qquad \infnorm{\Vhat^\star - V^\star} \leq \theta
    \end{equation*}
    where
    \begin{equation*}
        \theta = \frac{52}{1-\gamma} \sqrt{\frac{D\log(24 S^3 A/\delta)}{n}}. 
    \end{equation*}
\end{theorem}
\begin{proof}
    First, we assume that $n \geq 2664 D \log(24S^3A/\delta)$. Then applying Lemma \ref{lem:absorbing_bernstein_ineq_diameter} with failure probability $\delta/(2SA)$ for all $s \in \S, a \in \A$, and taking a union bound, we have that with probability at least $1-\delta/2$,
    \begin{equation*}
        \left|(\Phat - P)\Vhat^\star \right| \leq \sqrt{\frac{18\Var_{P}[\Vhat^\star]\log(12SA/\delta)}{n}} + \frac{84D\log(12SA/\delta)}{n}\one.
    \end{equation*}
    We also note that the underlying good event constructed within the proof of Lemma \ref{lem:absorbing_bernstein_ineq_diameter} also establishes that $\Phat$ is communicating and $\diam(\Phat) \leq 2D$. This furthermore implies by Lemma \ref{lem:optimal_value_fn_difference_diameter} that $\spannorm{\Vhat^\star} \leq \diam(\Phat) \leq 2D$. Specializing the above display to the rows of the form $(s, \pihat(s))$, we obtain that
    \begin{equation}
        \left|(\Phat_{\pihat} - P_{\pihat})\Vhat^\star \right| \leq \sqrt{\frac{18\Var_{P_{\pihat}}[\Vhat^\star]\log(12SA/\delta)}{n}} + \frac{84D\log(12SA/\delta)}{n}\one. \label{eq:high_discount_plug_in_empirical_pol_bd}
    \end{equation}
    Since $\infnorm{\Vhat^\star - \Vhat^{\pihat}} \leq \frac{1}{n}$ by assumption, applying Lemma \ref{lem:error_to_var_param} with $V = \Vhat^\star$, $\alpha = 18\log(12SA/\delta)$, $M  = 5 D$ (note $5 = 90/18 > 84/18 $ and $5D > 2D \geq \spannorm{\Vhat^\star}$), we obtain
    \begin{equation}
        \infnorm{\Vhat^{\pihat} - V^{\pihat}} \leq \frac{\gamma \sqrt{\alpha}}{\sqrt{n}\sqrt{1-\gamma}} \sqrt{\infnorm{(I - \gamma P_{\pihat})^{-1} \Var_{P_{\pihat}}[\Vhat^\star]}} + \gamma \frac{2 + \alpha M}{n (1-\gamma)}.\label{eq:high_discount_plug_in_empirical_val_err_1}
    \end{equation}
    We can then use Lemma \ref{lem:var_lemma} to bound the variance parameter, using the same choices of $V, \alpha, M$ and noting that $\Vhat^\star \geq \Vhat^{\pihat}$ and
    \begin{align*}
        \infnorm{\Vhat^\star - r_{\pihat} - \gamma \Phat_{\pihat}\Vhat^\star} &\leq \infnorm{(I - \gamma \Phat_{\pihat})(\Vhat^\star -\Vhat^{\pihat})} + \infnorm{\Vhat^{\pihat} - r_{\pihat} - \gamma \Phat_{\pihat} \Vhat^{\pihat}} \\
        & \leq \infinfnorm{I - \gamma \Phat_{\pihat}}\infnorm{\Vhat^\star -\Vhat^{\pihat}} \leq (1 + \gamma)\frac{1}{n} \leq \frac{2}{n},
    \end{align*}
    to obtain that
    \begin{align*}
        \infnorm{(I - \gamma P_{\pihat})^{-1} \Var_{P_{\pihat}}[\Vhat^\star]} &\leq 8 \frac{\spannorm{\Vhat^\star}M}{\gamma^2} \frac{\alpha}{n(1-\gamma)} + \frac{4 \spannorm{\Vhat^\star} (1 + \frac{2}{n})}{\gamma^2 (1-\gamma)} \\
        & \leq \frac{\spannorm{\Vhat^\star}}{\gamma^2 (1-\gamma)} \left(\frac{8 \cdot 5D \alpha}{148 D \alpha} + 4 \cdot (1+\frac{2}{2664}) \right) \leq 5\frac{\spannorm{\Vhat^\star}}{\gamma^2 (1-\gamma)} 
    \end{align*}
    where we simplified using the assumption that $n \geq 2664 D \log(24S^3A/\delta) \geq 148 D \alpha$. Using this in~\eqref{eq:high_discount_plug_in_empirical_val_err_1} yields that
    \begin{equation}
        \infnorm{\Vhat^{\pihat} - V^{\pihat}} \leq \frac{1}{1-\gamma}\sqrt{\frac{5\alpha \spannorm{\Vhat^\star}}{n}} + \frac{1}{1-\gamma} \frac{2 + 5\alpha D}{n}. \label{eq:high_discount_plug_in_empirical_val_err_2}
    \end{equation}

    Now we repeat these steps to bound $\infnorm{\Vhat^{\pistar} - V^{\pistar}}$ for a deterministic $\gamma$-discounted optimal policy $\pistar$.
    Using Lemma \ref{lem:fixed_pol_empirical_bernstein} with failure probability $\delta/2$ (note the previous assumption on $n$ implies $n \geq 6$ because $D \geq 1$), we have that
    \begin{equation}
        \left|(\Phat_{\pistar} - P_{\pistar})V^\star \right| \leq \sqrt{\frac{3\Var_{\Phat_{\pistar}}[V^\star]\log(8S/\delta)}{n}} + \frac{3\spannorm{V^\star}\log(8S/\delta)}{n}.\label{eq:high_discount_plug_in_true_pol_bd}
    \end{equation}
    Since the already-defined $\alpha$ has $\alpha \geq 3\log(8S/\delta)$, applying Lemmas \ref{lem:error_to_var_param} and \ref{lem:var_lemma} with this $\alpha, V = V^{\pistar}, \varepsilon = 0$, and $M = \spannorm{V^\star}$, we obtain that
    \begin{align*}
        \infnorm{\Vhat^{\pistar} - V^{\pistar}} & \leq \gamma \sqrt{\frac{\alpha}{(1-\gamma)n}} \sqrt{\infnorm{(I - \gamma \Phat_{\pistar})^{-1} \Var_{\Phat_{\pistar}} [V^\star]}} + \frac{\gamma \alpha \spannorm{V^\star}}{n(1-\gamma)}, \\
        \infnorm{(I - \gamma \Phat_{\pistar})^{-1} \Var_{\Phat_{\pistar}} [V^\star]} & \leq 8 \frac{\spannorm{V^\star}^2 \alpha }{\gamma^2 n(1-\gamma)} + \frac{4 \spannorm{V^\star} \cdot 1}{\gamma^2 (1-\gamma)}  \leq \frac{\spannorm{V^\star}}{\gamma^2 (1-\gamma)} \left(\frac{8}{148} + 4 \right) \\
        &\leq 5 \frac{\spannorm{V^\star}}{\gamma^2 (1-\gamma)},
    \end{align*}
    respectively, where we simplified using the fact that $\spannorm{V^\star }\leq D$ and as already noted $n \geq 148 D \alpha$. Combining these yields the bound
    \begin{equation*}
        \infnorm{\Vhat^{\pistar} - V^{\pistar}} \leq \frac{1}{1-\gamma}\sqrt{\frac{5\alpha \spannorm{V^\star}}{n}} + \frac{\alpha \spannorm{V^\star}}{(1-\gamma)n}.
    \end{equation*}

    Using this display as well as~\eqref{eq:high_discount_plug_in_empirical_val_err_2} in Lemma \ref{lem:deterministic_error_decomp}, we have that $V^{\pihat} \geq V^\star - \theta \one$ and $\infnorm{\Vhat^\star - V^\star} \leq \theta$ where
    \begin{align*}
        \theta &= \frac{1}{n} + \frac{1}{1-\gamma} \left(\sqrt{\frac{5\alpha \spannorm{\Vhat^\star}}{n}} + \frac{2 + 5\alpha D}{n} + \sqrt{\frac{5\alpha \spannorm{V^\star}}{n}} + \frac{\alpha \spannorm{V^\star}}{n}\right) \\
        & \leq \frac{1}{1-\gamma} \left( \sqrt{\frac{5 \alpha}{n}} \left(\sqrt{\spannorm{\Vhat^\star}} + \sqrt{\spannorm{V^\star}} \right) +\frac{3 + 6 \alpha D}{n} \right) \\
        & \leq \frac{1}{1-\gamma} \left( \sqrt{\frac{5 \alpha}{n}} \left(\sqrt{2D} + \sqrt{D} \right) +\sqrt{\frac{3 + 6 \alpha D}{n}} \right) \\
        & \leq \frac{52}{1-\gamma} \sqrt{\frac{D\log(24 S^3 A/\delta)}{n}}.
    \end{align*}
    Finally we note that when the assumption $n \geq 2664 D \log(24S^3A/\delta)$ does not hold, the final expression above is $> \frac{1}{1-\gamma}$ and so the desired bound holds trivially.
\end{proof}

Let $\distn := \{d \in \Delta(\S) : \forall s' \in \S, d(s') = m/n \text{ for some integer $m$ with $0 \leq m \leq n$}\}$ be the set of all distributions over $\S$ such that all probabilities are a multiple of $1/n$.
Let $\emps := \{\Bar{P} \in \R^{SA \times S} : \Bar{P}_{sa} \in \distn \text{ for all $s\in \S, a \in \A$}\}$ be the set of all MDP transition kernels with all rows in $\distn$. Note that by construction of the empirical MDP we have $\Phat \in \emps$ almost surely.
Finally let $\emdps$ be the set of all possible associated MDPs $\emdps = \{(\S, \A, \Bar{P}, r) : \Bar{P} \in \emps\}$.

\begin{lemma}
    \label{lem:bias_optimal_planning_discount_properties}
    Suppose $P$ is weakly communicating. Then there exists $\gbar \in (0,1)$ such that the following is true:
\begin{enumerate}
    \item We have $\infnorm{(1-\gbar) V_{\gbar}^\star - \rho^\star} \leq (1-\gbar)\spannorm{V_{\gbar}^\star} \leq \frac{1}{n}$.
    \item For any MDP $\widetilde{\M}_n \in \emdps$, if $\pihat \in \PiSD$ is bias-optimal in $\widetilde{\M}_n$, then $\infnorm{V^{\star}_{\widetilde{\M}_n, \gbar} - V^{\pihat}_{\widetilde{\M}_n, \gbar}} \leq \frac{1}{n}$.
    \item For any MDP $(\S, \A, \Bar{P}, r) = \widetilde{\M}_n \in \emdps$ such that $\Bar{P}$ is weakly communicating, we have that $\infnorm{(1-\gbar) V_{\widetilde{\M}_n,\gbar}^\star - {\rho}_{\widetilde{\M}_n}^\star} \leq \frac{1}{n}.$
\end{enumerate}
\end{lemma}
\begin{proof}
For the first conclusion of the lemma, as long as $\gbar$ is large enough that $\frac{1}{1-\gbar} \geq 2n \spannorm{h^\star}$, by the third and first parts of Lemma \ref{lem:average_reward_facts} we have
\begin{equation*}
    \infnorm{(1-\gbar) V_{\gbar}^\star - \rho^\star} \leq (1-\gbar)\spannorm{V_{\gbar}^\star} \leq 2 (1-\gbar) \spannorm{h^\star} \leq \frac{1}{n}.
\end{equation*}

We can take a similar approach for the third conclusion of the lemma. By the same steps in the MDP $\widetilde{\M}_n$ we have
\begin{equation*}
    \infnorm{(1-\gbar) V_{\widetilde{\M}_n,\gbar}^\star - {\rho}_{\widetilde{\M}_n}^\star} \leq (1-\gbar)\spannorm{V_{\gbar, \widetilde{\M}_n}^\star} \leq 2 (1-\gbar) \spannorm{h_{\widetilde{\M}_n}^\star} \leq 2(1-\gbar) \max_{\widetilde{\M} \in \emdps} \spannorm{h_{\widetilde{\M}}^\star},
\end{equation*}
which is finite since $\emdps$ is finite. Therefore it suffices to furthermore require that
\[\frac{1}{1-\gbar} \geq 2n \max_{\widetilde{\M} \in \emdps} \spannorm{h_{\widetilde{\M}}^\star}.\]

Finally we handle the second conclusion of the lemma. By definition for a bias-optimal policy $\pi$ in an MDP $\widetilde{\M}$ we have $\infnorm{V^\star_{\widetilde{\M}, \gamma} - V^\pi_{\widetilde{\M}, \gamma}} \to 0$ as $\gamma \uparrow 1$, so there exists a fixed $\gamma(\widetilde{M}, \pi) \in (0,1)$ such that $\gamma \geq \gamma(\widetilde{M}, \pi)$ implies that $\infnorm{V^\star_{\widetilde{\M}, \gamma} - V^\pi_{\widetilde{\M}, \gamma}} \leq \frac{1}{n}$.
Since the number of stationary deterministic policies is finite and the set $\emdps$ is finite, we have that $\max_{\widetilde{\M} \in \emdps} \max_{\pi \in \PiSD : \pi \text{ bias-optimal in $\widetilde{\M}$}} \gamma(\widetilde{M}, \pi) < 1$. Hence it suffices to finally also require that $\gbar $ is at least this quantity.
\end{proof}

\begin{proof}[Proof of Theorem \ref{thm:main_average_reward_plug_in}]
    We set $\gbar$ to satisfy the conditions in Lemma \ref{lem:bias_optimal_planning_discount_properties}. Then applying Theorem \ref{thm:high_discount_plug_in} with discount factor $\gbar$, with probability at least $1-\delta$ we have for all policies $\pihat$ satisfying $\infnorm{\Vhat^{\star}_{\gbar} - \Vhat^{\pihat}_{\gbar}} \leq \frac{1}{n}$ that
    \begin{equation*}
            V_{\gbar}^{\pihat} \geq V_{\gbar}^\star - \theta \one \qquad\text{and}\qquad \infnorm{\Vhat_{\gbar}^\star - V_{\gbar}^\star} \leq \theta
    \end{equation*}
    where
    \begin{equation*}
        \theta = \frac{52}{1-\gbar} \sqrt{\frac{D\log(24 S^3 A/\delta)}{n}}. 
    \end{equation*}
    By Lemma \ref{lem:bias_optimal_planning_discount_properties}, all deterministic stationary bias-optimal policies for $\Phat$ satisfy $\infnorm{\Vhat^{\star}_{\gbar} - \Vhat^{\pihat}_{\gbar}} \leq \frac{1}{n}$. From here on we assume $\pihat$ is bias-optimal.
    
    By Lemma \ref{lem:average_reward_facts} and the first part of Lemma \ref{lem:bias_optimal_planning_discount_properties}, we have
    \begin{align*}
        \rho^{\pihat} &\geq \rho^\star - (1-\gbar) \infnorm{V_{\gbar}^\star - V_{\gbar}^{\pihat}} - (1-\gbar) \spannorm{V_{\gbar}^\star} \\
        & \geq \rho^\star - 52 \sqrt{\frac{D \log(24S^3 A/\delta)}{n}} - \frac{1}{n} \\
        & \geq \rho^\star - 53 \sqrt{\frac{D \log(24S^3 A/\delta)}{n}}.
    \end{align*}
    
    From the proof of Theorem \ref{thm:high_discount_plug_in}, when we are in the non-vacuous regime that $52 \sqrt{\frac{D \log(24S^3 A/\delta)}{n}} < 1$, under the same event we will further have that $\Phat$ is communicating. Then we can apply the final part of Lemma \ref{lem:bias_optimal_planning_discount_properties} to obtain that $\infnorm{(1-\gbar) \Vhat_{\gbar}^\star - \widehat{\rho}^\star} \leq \frac{1}{n}$.
    Therefore
    \begin{align*}
        \infnorm{\widehat{\rho}^\star - \rho^\star} & \leq \infnorm{(1-\gbar) V_{\gbar}^\star - \rho^\star}+(1-\gbar)\infnorm{\Vhat_{\gbar}^\star - V_{\gbar}^\star} +\infnorm{(1-\gbar) \Vhat_{\gbar}^\star - \widehat{\rho}^\star} \\
        & \leq \frac{1}{n} + (1-\gbar)\frac{52}{1-\gbar} \sqrt{\frac{D\log(24 S^3 A/\delta)}{n}} + \frac{1}{n} \\
        & \leq 54 \sqrt{\frac{D\log(24 S^3 A/\delta)}{n}}
    \end{align*}
    where the first term is bounded from Lemma \ref{lem:bias_optimal_planning_discount_properties}. Also whenever we are in the vacuous regime where $52 \sqrt{\frac{D \log(24S^3 A/\delta)}{n}} \geq 1$ then the above is larger than $1$ and so is still true.
\end{proof}

\subsection{Extension to Weakly Communicating MDPs}
\begin{theorem}
\label{thm:average_reward_plug_in_weakly_communicating}
    Suppose $P$ is weakly communicating and, letting $\M_c = (\S_c, \A, P_c, r_c)$ be the communicating sub-MDP, suppose that $|\S_c| \geq 2$ and $\diam(P_c) \leq D$. Fix $\delta \in (0,1)$.
    Then with probability at least $1 - \delta$, for all deterministic policies $\pihat$ that are bias-optimal for $\Phat$,
    we have
    \begin{equation*}
        \rho^{\pihat} \geq \rho^\star - \eta \one \qquad \text{and}\qquad \max_{s \in \S_c}\left|\widehat{\rho}^{\star}(s) - \rho^\star(s)\right| \leq \eta,
    \end{equation*}
    where
    \begin{equation*}
        \eta = 54 \sqrt{\frac{D\log(24 S^3 A/\delta)}{n}}.
    \end{equation*}
\end{theorem}
Compared to Theorem \ref{thm:main_average_reward_plug_in}, the above theorem obtains the same suboptimality guarantee for the output policy $\pihat$, but $\widehat{\rho}^\star$ is only guaranteed to approximate $\rho^\star$ on the states of the communicating sub-MDP $\S_c$. It is easy to see that this is unimprovable without a quantitative dependence on the values of $P$ on states in $\S \setminus \S_c$: one can take a communicating MDP and append a state $s_\bot$ where all actions give reward $r_\bot$ and have probability $p_\bot$ of reaching the communicating MDP, and otherwise transition back to $s_\bot$; if $p_\bot \leq \frac{1}{2An}$ then with at least constant probability no transitions will be observed from $s_\bot$ to the communicating MDP and then $\widehat{\rho}^\star(s_\bot) = r_\bot$ (which can be chosen to be far from $\rho^\star$).

\begin{proof}[Proof of Theorem \ref{thm:average_reward_plug_in_weakly_communicating}]
    By applying Theorem \ref{thm:main_average_reward_plug_in} to the communicating sub-MDP, we immediately have that with probability at least $1-\delta$,
    \begin{equation*}
        \rho^{\pihat}(s) \geq \rho^\star(s) - \eta ~\forall s \in \S_c \qquad \text{and}\qquad \max_{s \in \S_c}\left|\widehat{\rho}^{\star}(s) - \rho^\star(s)\right| \leq \eta.
    \end{equation*}
    Additionally, since all $s \in \S \setminus \S_c$ are transient under all policies, including $\pihat$, since $P_{\pihat}^\infty \rho^{\pihat} = P_{\pihat}^\infty P_{\pihat}^\infty r_{\pihat} = P_{\pihat}^\infty r_{\pihat} = \rho^{\pihat}$, we have $\rho^{\pihat}(s) = \sum_{s' \in \S_c} (P_{\pihat}^\infty)_{ss'}\rho^{\pihat}(s') \geq \min_{s' \in \S_c} \rho^{\pihat}(s') \geq \rho^\star(s) - \eta$,
    using the fact that $\rho^\star$ is state-independent.
\end{proof}
Also we note that the assumption that the communicating sub-MDP has at least $2$ states is made to avoid an edge case because technically Theorem \ref{thm:main_average_reward_plug_in} assumes at least two states and we have not explicitly defined a convention for the diameter of an MDP with one state. The conclusions of the theorem will actually be trivially true in the case that the sub-MDP has only one state, for the same reasons as discussed in Section \ref{sec:setup}.

\section{Conclusion}
In this work we resolved the minimax sample complexity of discounted and average-reward reinforcement learning with an oblivious generative model up to only absolute constants. While this is one of the best-studied benchmark problems in the theory of reinforcement learning, it is undoubtedly a highly idealized setting. Still, algorithms and analytical techniques originating from the study of this problem have led to a wide range of improvements in the theory of RL, and we hope that this work might have the same effect.

\section*{Acknowledgments}
M.\ Zurek and Y.\ Chen acknowledge support by National Science Foundation grants CCF-2233152 and DMS-2023239, by a Cisco Systems Fellowship, and by a Vilas Associates Award.

\section*{AI Use Statement}
The main ideas and all writing were contributed by the authors. AI was then used to check the arguments. 

\bibliographystyle{plainnat}
\bibliography{references_2}

\appendix

\section{Auxiliary Lemmas}

The following lemmas are the standard Hoeffding and Bernstein inequalities for bounded random variables \citep{boucheron_concentration_2013}.
\begin{lemma}
    \label{lem:hoeffding}
    Let $A, B \in \R$. Suppose that $Z_1, \dots, Z_n$ are i.i.d.\ random variables taking values in $[A,B]$ and let $\delta \in (0,1)$. Then with probability at least $1-\delta$,
    \begin{equation*}
        \frac{1}{n} \sum_{i=1}^n (Z_i - \E Z_1) \leq \sqrt{\frac{(B-A)^2 \log(1/\delta)}{2n}}.
    \end{equation*}
\end{lemma}
\begin{lemma}
\label{lem:bernstein}
    Let $A, B \in \R$. Suppose that $Z_1, \dots, Z_n$ are i.i.d.\ random variables taking values in $[A,B]$ and let $\delta \in (0,1)$. Then with probability at least $1-\delta$,
    \begin{equation*}
        \frac{1}{n}\sum_{i=1}^n (Z_i - \E Z_1) \leq \sqrt{\frac{2\Var [Z_1] \log(1/\delta)}{n}} + \frac{(B-A)\log(1/\delta)}{3n}.
    \end{equation*}
\end{lemma}

\begin{lemma}
\label{lem:empirical_bernstein}
    Suppose that $Z_1, \dots, Z_n$ are i.i.d. random variables taking values in $[A,B]$ and let
    \begin{equation*}
        \widehat{\Sigma}_n = \frac{1}{n^2} \sum_{1 \leq i < j \leq n} (Z_i - Z_j)^2 = \frac{1}{n} \sum_{i=1}^n \left( Z_i - \left(\frac{1}{n}\sum_{j=1}^n Z_j\right)\right)^2
    \end{equation*}
    be the (uncorrected/biased) empirical variance.
    Let $\delta \in (0,1)$ and suppose $n \geq 6$. Then with probability at least $1-\delta$,
    \begin{equation*}
        \frac{1}{n}\sum_{i=1}^n (Z_i - \E Z_1) \leq \sqrt{\frac{3\widehat{\Sigma}_n \log(2/\delta)}{n}} + \frac{3(B-A)\log(2/\delta)}{n}.
    \end{equation*}
\end{lemma}
\begin{proof}
    This is a minor modification of \cite[Theorem 4]{maurer_empirical_2009}, which, after transforming the random variables $(Z_i)_{i}$ to $(\frac{B- Z_i}{B-A})_i$, in order to meet the conditions of their theorem, then yields that for any $n \geq 2$ and any $\delta \in (0,1)$, with probability at least $1-\delta$,
    \begin{equation*}
        \frac{1}{n}\sum_{i=1}^n (Z_i - \E Z_1) \leq \sqrt{\frac{2\widehat{\Sigma}_n \log(2/\delta)}{n-1}} + \frac{7(B-A)\log(2/\delta)}{3(n-1)}.
    \end{equation*}
    (Also note separately that this holds trivially for the $B=A$ case.)
    Whenever furthermore $n \geq 6$, it can then easily be verified that
    \begin{equation*}
        \frac{2}{n-1} < \frac{7}{3(n-1)} \leq \frac{3}{n},
    \end{equation*}
    which can be combined to yield the desired conclusion.
\end{proof}

The following lemma is the well-known Bhatia-Davis inequality \citep{bhatia_better_2000} applied to next-state-value random variables.
\begin{lemma}
Let $p \in \Delta(\S)$ and $V \in \R^S$. Then
\label{lem:bhatia_davis}
    \begin{equation*}
        \Var_p[V] \leq (\max_{s \in \S}V(s) - pV) (pV - \min_{s \in \S}V(s)).
    \end{equation*}
\end{lemma}

The following lemma collects useful facts from prior work for the average-reward results. The bound $\spannorm{V_\gamma^\star} \leq 2 \spannorm{h^\star}$ was shown by \cite{wei_modelfree_2020} and the next two bounds appear in \cite{zurek_spanagnostic_2025a}. Similar arguments also appear in \cite{wang_sampleoptimal_2022}.
\begin{lemma}
    \label{lem:average_reward_facts}
    Let $\M = (\S, \A, P, r)$ be an MDP. All value and bias functions below are with respect to $\M$. Then the following hold:
    \begin{enumerate}
        \item If $P$ is weakly communicating, for any $\gamma \in [0,1)$, we have $\spannorm{V_\gamma^\star }\leq 2\spannorm{h^\star}$.
        \item For any $\gamma \in [0,1)$ and any stationary policy $\pi$, we have
        \begin{equation*}
            \rho^\pi \geq \rho^\star - (1-\gamma)\infnorm{V_\gamma^{\star} - V_\gamma^\pi} - (1-\gamma)\spannorm{V_\gamma^\star}.
        \end{equation*}
        \item For any $\gamma \in [0,1)$,
        \begin{equation*}
            \infnorm{(1-\gamma)V_\gamma^\star - \rho^\star} \leq (1-\gamma)\spannorm{V_\gamma^\star}.
        \end{equation*}
    \end{enumerate}
\end{lemma}

The following equivalence is a standard fact which follows immediately from the fact that, for any fixed target state $s'$, the associated stochastic-shortest path problem of minimizing the time to reach $s'$ admits a stationary deterministic policy which is optimal from all starting states simultaneously \cite[Proposition 2]{bertsekas_analysis_1991}.
\begin{lemma}
\label{lem:diameter_equivalence}
    Suppose that $\diam(P) < \infty$. Then
    \begin{equation*}
        \diam(P) := \max_{s,s' \in \S : s \neq s'} \min_{\pi \in \PiSD} \E^{P_\pi}_{s}[\tau_{s'}] = \max_{s' \in \S} \min_{\pi \in \PiSD} \max_{s \neq s'}\E^{P_\pi}_{s}[\tau_{s'}].
    \end{equation*}
\end{lemma}

The following lemma relates the value function of the $g$-goal true MDP as defined in Definition \ref{def:g_goal_mdps} to the diameter. It appears as the first calculation within \cite[Appendix D]{zurek_plugin_2025}.
\begin{lemma}
    \label{lem:val_fn_lower_bound_diameter}
    Fix $s_0, g \in \S$ and $\gamma \in (0,1)$. Then for any stationary policy $\pi$,
    \begin{equation*}
        V^\pi_{\to g,\gamma} (s_0) \geq \frac{1}{1-\gamma} - \E^{P_\pi}_{s_0}[\hitt_g].
    \end{equation*}
\end{lemma}
The next lemma is a generalization that holds for arbitrary value functions (still assuming $\zero \leq r \leq \one$), and it also can be seen as a discounted adaptation of \cite[Exercise 38.3]{lattimore_bandit_2020}.
\begin{lemma}
\label{lem:optimal_value_fn_difference_diameter}
    Fix $s_0, s \in \S$ and $\gamma \in (0,1)$. Then for any stationary policy $\pi$ such that $\E^{P_\pi}_{s_0}[\tau_s] < \infty$,
    \begin{equation*}
        V^\star_\gamma(s_0) \geq V_\gamma^\star(s) - \E^{P_\pi}_{s_0}[\tau_s].
    \end{equation*}
    Consequently, if $P$ is communicating, then $\spannorm{V_\gamma^\star} \leq \diam(P)$.
\end{lemma}
\begin{proof}
    Let $\tilde{\pi}$ be a history-dependent policy which follows $\pi$ until reaching $s$, and then follows an optimal policy $\pistar$ afterwards. It is a well-known fact that the discounted value of such a policy is at most $V_\gamma^\star$ \citep{puterman_markov_1994}. Then (treating $\gamma^\infty = 0$)
    \begin{align*}
        V_\gamma^\star(s_0) &\geq V_\gamma^{\tilde{\pi}}(s_0) = \E^{\tilde{\pi}}_{s_0}\sum_{t=0}^\infty \gamma^t r(S_t,A_t) = \E_{s_0}^{P_{\pi}}\left[ \sum_{t=0}^{\tau_s-1}\gamma^t r(S_t, A_t) + \gamma^{\tau_s}V_\gamma^\star(s) \right] \geq \E_{s_0}^{P_{\pi}}\left[ \gamma^{\tau_s}V_\gamma^\star(s) \right]
    \end{align*}
    where the second equality uses the strong Markov property and the fact that $\tilde{\pi}$ follows $\pi$ before reaching $s$. To lower bound the final term, we notice that for any $t \geq 0$ we have from the formula for a geometric series that
    \begin{equation*}
        \frac{1-\gamma^t}{1-\gamma} = \sum_{k=0}^{t-1} \gamma^k \leq t \implies \gamma^t \geq 1 - (1-\gamma)t.
    \end{equation*}
    Plugging this bound into the previous inequality, we obtain
    \begin{equation*}
        V_\gamma^\star(s_0)\geq V_\gamma^\star(s) \E_{s_0}^{P_{\pi}}\left[ \gamma^{\tau_s} \right] \geq V_\gamma^\star(s) \E_{s_0}^{P_{\pi}}[1 - (1-\gamma)\tau_s] \geq V_\gamma^\star(s) - \E_{s_0}^{P_{\pi}}[\tau_s]
    \end{equation*}
    where the final inequality uses that $(1-\gamma)V_\gamma^\star(s)\leq 1$.

    Next, if $P$ is communicating, then for the $s,s' \in \S$ such that $\spannorm{V^\star_\gamma} = V^\star_\gamma(s) - V^\star_\gamma(s')$, there exists some policy $\pi$ such that $\E_{s'}^{P_\pi}[\tau_s] \leq \diam(P)$. Then by the first statement, we have that
    \begin{equation*}
        \spannorm{V^\star_\gamma} = V^\star_\gamma(s) - V^\star_\gamma(s') \leq \E_{s'}^{P_\pi}[\tau_s] \leq \diam(P).
    \end{equation*}
\end{proof}
We remark that the bound on $\spannorm{V^\star_\gamma}$ has appeared previously \citep[Lemma E.2]{wei_online_2017}, using a similar argument, but we also need the first statement within our proofs.

\section{Average-Reward Lower Bound Proof}
\label{sec:average_reward_lower_bound_proof}

In this section we complete the proof of Theorem \ref{thm:avg_rew_lower_bound}.

\subsection{Notation}
Here we define some additional notation used only within this proof.
For any two discrete distributions with PMFs $\mu, \nu$, we define their KL divergence $\kl{\mu}{\nu} = \sum_x \mu(x) \log \frac{\mu(x)}{\nu(x)}$.
For any $p \in [0,1]$ we let $\Ber(p)$ denote the law of a Bernoulli random variable with parameter $p$, and additionally define
$\klb(p,q) := \kl{\Ber(p)}{\Ber(q)}$ and $h(p) = -p \log p - (1-p) \log(1-p)$.
Also, for any unichain policy $\pi \in \PiSR$, we define its state-action occupancy $y_{\M}^\pi(s,a) := \pi(a \mid s)\mu_{\M}^\pi(s)$.

\subsection{Construction}
\label{sec:lb_construction}
Fix even positive natural numbers $A \geq 4, S \geq 2$.
Fix an integer $ D \geq \max\{1, \log_{A/2}(S/2)\}$ and $\varepsilon \in (0,1/4]$.
Define $\pz = \frac{1}{2D}$ and $\ps = \frac{1}{(2-\varepsilon)D}$.

\paragraph{Two-state gadget MDP}
First we describe a two-state gadget MDP. Let $\bs, \ts$ be two states. For $i \in \{0, 1, \dots, A/2\}$, we define the transition kernel $Q^{(i)}$ over these two states and action space $\A_{/2} := \{1, \dots, A/2\}$ by
\begin{gather*}
    Q^{(i)}(\bs \mid \bs, a) := \begin{cases}
        1-\ps & a = i \\
        1 - \pz & a \neq i
    \end{cases} , \quad Q^{(i)}(\ts \mid \bs, a) := \begin{cases}
        \ps & a = i \\
        \pz & a \neq i
    \end{cases}, \\
    Q^{(i)}(\ts \mid \ts, a) := 1- \pz,  \quad Q^{(i)}(\bs \mid \ts, a) := \pz,
\end{gather*}
and we also define the action-independent reward function $\tr : \{\bs, \ts\} \times \A_{/2} \to [0,1]$ by
\begin{equation*}
    \tr(s,a) := \begin{cases}
        1 & s = \ts \\
        0 & s = \bs
    \end{cases}.
\end{equation*}
In words, all actions for either state of $\ts, \bs$ can be understood as having a high probability of remaining in the current state and a small probability of transitioning to the other state; furthermore, the probability of moving to the other state is $\pz$ for all states and all actions, except for action $i$ in state $\bs$, where the probability is the slightly larger value $\ps$. Because state $\ts$ has reward $1$ for all actions while $\bs$ has reward $0$ for all actions, choosing action $i$ in state $\bs$ is optimal because it leads to $\ts$ more quickly (and the action chosen in $\ts$ does not matter). However, this is only true when $i \geq 1$; if $i = 0$, then all state-actions have the same probability $\pz$ of transitioning to the other state, and all policies have the same gain.
For any $i \in \{0, \dots, A/2\}$, we denote the gadget MDP with parameter $i$ by $\M_2^{(i)} := (\{\bs, \ts\}, \A_{/2}, Q^{(i)}, \tr) $.


First we collect some simple properties of these gadget MDPs.
\begin{lemma}
\label{lem:lb_gadget_kl_ub}
    For any $i, j \in \A_{/2}$ such that $j \neq i$, we have
    \begin{equation*}
        \kl{Q^{(i)}(\cdot \mid \bs, i)}{Q^{(i)}(\cdot \mid \bs, j)}, \kl{Q^{(i)}(\cdot \mid \bs, j)}{Q^{(i)}(\cdot \mid \bs, i)} \leq \frac{\varepsilon^2}{3D}.
    \end{equation*}
\end{lemma}
\begin{proof}
    First note that we have
    \begin{equation*}
        \kl{Q^{(i)}(\cdot \mid \bs, i)}{Q^{(i)}(\cdot \mid \bs, j)} = \klb(\ps, \pz) \quad \text{and} \quad \kl{Q^{(i)}(\cdot \mid \bs, j)}{Q^{(i)}(\cdot \mid \bs, i)} = \klb(\pz, \ps).
    \end{equation*}
    By Lemma \ref{lem:bernoulli_kl_ub} we can bound
    \begin{align*}
        \klb(\ps, \pz) &\leq \frac{(\ps - \pz)^2}{\pz(1-\pz)} =\frac{\left( \frac{1}{2D} - \frac{1}{(2-\varepsilon)D} \right)^2}{\frac{1}{2D}\left(1 -\frac{1}{2D} \right)} = \frac{\varepsilon^2 D^2 (2D)^2}{(2D-1)(2D)^2 (2-\varepsilon)^2 D^2} \\
        &= \frac{\varepsilon^2}{(2D-1)(2-\varepsilon)^2} \leq \frac{\varepsilon^2}{3D}
    \end{align*}
    where in the final inequality we used that $D \geq 1$ so $2D-1 \geq D$, and that $(2-\varepsilon)^2 \geq 1.75^2 > 3$ since $\varepsilon \leq \frac{1}{4}$.
    Also starting with Lemma \ref{lem:bernoulli_kl_ub}, we have
    \begin{align*}
        \klb(\pz, \ps) &\leq \frac{(\ps - \pz)^2}{\ps(1-\ps)} = \frac{\left( \frac{1}{2D} - \frac{1}{(2-\varepsilon)D} \right)^2}{\frac{1}{(2-\varepsilon)D}\left(1 -\frac{1}{(2-\varepsilon)D} \right)} = \frac{\varepsilon^2 }{2^2 (2-\varepsilon)^2 D^2} \frac{(2-\varepsilon)^2D^2}{(2-\varepsilon )D - 1} \\
        &=  \frac{\varepsilon^2}{4 ((2-\varepsilon )D - 1)} \leq \frac{\varepsilon^2}{3D}
    \end{align*}
    where in the final inequality we use that $(2-\varepsilon )D - 1 \geq \frac{7}{4}D-1 \geq \frac{3}{4}D$ since $D \geq 1$ and $\varepsilon\leq \frac{1}{4}$.
\end{proof}

\begin{lemma}
    \label{lem:lb_gadget_opt_gain}
    For any $i \in \{1, \dots, A/2\}$, the optimal gain $\rho^\star_{\M_2^{(i)}}$ is equal to $\frac{2}{4-\varepsilon}\one$.
\end{lemma}
\begin{proof}
    By the many identical state-action pairs within the gadget MDP $\M_2^{(i)}$, there are only two stationary deterministic policies $\pi, \pi'$ which produce distinct distributions over state trajectories, where $\pi(\bs) = i$ and $\pi'(\bs) \neq i$. Also since the reward is action-independent, in order to compute each of their gains it will suffice to compute the stationary distributions of each of these policies, and then since there exists some stationary deterministic policy with optimal gain, the optimal gain will be the maximum of the gains of these two policies. Note that both policies are unichain since from either state under either policy, there is probability at least $\pz$ of transitioning to $\ts$. It is easy to check by solving the equations $\mu^\pi Q^{(i)}_\pi = \mu^\pi, \mu^{\pi'} Q^{(i)}_{\pi'} = \mu^{\pi'}$ (and normalizing) that
    \begin{align*}
        \begin{bmatrix}
             \mu^{\pi}(\bs) & \mu^{\pi}(\ts)
        \end{bmatrix} = \begin{bmatrix}
             \frac{2-\varepsilon}{4-\varepsilon} & \frac{2}{4-\varepsilon}
        \end{bmatrix}, \qquad \begin{bmatrix}
             \mu^{\pi'}(\bs) & \mu^{\pi'}(\ts)
        \end{bmatrix} = \begin{bmatrix}
             \frac{1}{2} & \frac{1}{2}
        \end{bmatrix}.
    \end{align*}
    Hence $\rho^\pi = \frac{2}{4-\varepsilon} \one > \frac{1}{2}\one = \rho^{\pi'}$, so the optimal gain is $\frac{2}{4-\varepsilon} \one$.
\end{proof}

\begin{lemma}
\label{lem:lb_gadget_diam_bound}
    For any $i \in \{0, \dots, A/2\}$, we have that $\diam(\M_2^{(i)}) \leq 2D$.
\end{lemma}
\begin{proof}
    It can actually be easily seen that under any policy the expected amount of time to hit the other state is bounded by $2D$. For concreteness let $\pi$ be some policy such that $\pi(\bs) \neq i$, so that $Q^{(i)}(\ts \mid \bs, \pi(\bs)) = \pz$. Then we have
    \begin{align*}
        &\E^{Q^{(i)}_\pi}_{\bs} [\tau_{\ts}] = (1-\pz)\left(1 + \E^{Q^{(i)}_\pi}_{\bs} [\tau_{\ts}]\right) + \pz 1 = \frac{1}{2D} \left((2D - 1)\left(1 + \E^{Q^{(i)}_\pi}_{\bs} [\tau_{\ts}]\right) + 1 \right) \\
        \implies &\E^{Q^{(i)}_\pi}_{\bs} [\tau_{\ts}] = 2D.
    \end{align*}
    Also an identical calculation with $\ts$ and $\bs$ swapped shows that $\E^{Q^{(i)}_\pi}_{\ts} [\tau_{\bs}] = 2D$ as well.
\end{proof}

\paragraph{Full MDPs}
Now we explain how these gadgets are combined into the hard MDPs used in the lower bound.

Define the state space $\S := \widetilde{\S} \cup \bar{\S}$ where $\widetilde{\S} := \{\ts_1, \dots, \ts_{S/2}\}, \bar{\S} := \{\bs_1, \dots, \bs_{S/2}\}$. Let $\A = \{1, \dots, A\}$ be the action space.
Next, for any fixed $i \in \{0, 1, \dots, S/2\}$ and $a \in \A_{/2}$, we describe the transition kernel $P^{(i,a)}$, which we define for each different state-action pair separately.
First, for all $j \in \{1, \dots, S/2\}$ and all $a' \in \A$, we let
\begin{equation*}
    P^{(i,a)} (\ts_j \mid \ts_j, a') := 1-\pz \quad \text{and}\quad P^{(i,a)} (\bs_j \mid \ts_j, a') := \pz.
\end{equation*}
Next, for all $j \in \{1, \dots, S/2\}$ and all $a' \in \A_{/2}$, we let
\begin{equation*}
    P^{(i,a)} (\bs_j \mid \bs_j, a') := \begin{cases}
        1-\pz & (i,a) \neq (j,a') \\
        1 - \ps & \text{otherwise}
    \end{cases} \quad \text{and}\quad P^{(i,a)} (\ts_j \mid \bs_j, a') := \begin{cases}
        \pz & (i,a) \neq (j,a') \\
        \ps & \text{otherwise}
    \end{cases}.
\end{equation*}
Finally, for all $j \in \{1, \dots, S/2\}$ and all $a' \in \A \setminus \A_{/2}$, we add the following deterministic transitions: letting $B := A/2$, action $a'$ in state $\bs_j$ leads to state $\bs_{B \cdot j + (a'-B-1) \bmod S/2}$ with probability $1$ (where we define $\bs_0 $ to be the same state as $\bs_{S/2}$). To understand this choice, we first note that $B+1$ is subtracted from $a'$ so that the choice of $a' \in \A \setminus \A_{/2}$ becomes equivalent to a choice of a value in $\{0, 1, \dots, A - B - 1\} = \{0, 1, \dots, B-1\}$ (since $B = A/2$), which is more convenient for the following arithmetic. 
It can be seen by a simple inductive argument that, for any $j$, after starting in $\bs_j$, after taking the actions $a'_1, \dots, a_D' \in \A \setminus \A_{/2}$, letting $b'_1 = a'_1 - B - 1, \dots, b_D' = a_D' - B - 1$ (which are all elements of $\{0, 1, \dots, B-1\}$), we will at time $D$ reach state
\[
    \bs_{(B^Dj + B^{D-1}b'_1 + B^{D-2}b'_2 + \dots + B^0 b_D') \bmod S/2}.
\]
Since we have assumed that $D \geq \log_{A/2} (S/2)$, or equivalently that $\frac{S}{2} \leq \left( \frac{A}{2}\right)^D$, we have that any $j' \in \{0, \dots, S/2-1\}$ can be written as $j' = B^Dj + B^{D-1}b'_1 + B^{D-2}b'_2 + \dots + B^0 b_D' \bmod S/2$, so there exists a path of length $D$ between any pair of states in $\bar{\S}$.

The reward function $r : \S \times \A \to [0,1]$ will again be action-independent and also $1$ on all states of the form $\ts_j$ and $0$ on all states of the form $\bs_j$, that is,
\begin{equation*}
    r(s, a') := \begin{cases}
        1 & s \in \widetilde{\S} \\
        0 & s \in \bar{\S}
    \end{cases}.
\end{equation*}
We denote this MDP by $\M^{(i,a)} := (\S, \A, P^{(i,a)}, r)$.
Finally we also define $\M^{(0)} := \M^{(0,1)}$ (and note that the MDPs $\M^{(0,1)}, \dots, \M^{(0, A/2)}$ are in fact all the same).

\begin{lemma}
\label{lem:lb_full_mdps_properties}
    For any $i \in \{1, \dots, S/2\}, a \in \A_{/2}$, we have $\diam(P^{(i,a)})\leq 5D$, and the optimal gain in the MDP $\M^{(i,a)}$ is at least $\frac{2}{4-\varepsilon}\one$.
\end{lemma}
\begin{proof}
    It suffices to show that $\diam(P^{(i,a)})\leq 5D$, since this implies $\M^{(i,a)}$ is communicating and thus that there exists a policy that can reach $\bs_{i}$, that can then furthermore deterministically choose action $a$ in this state to yield a unichain policy with gain $\frac{2}{4-\varepsilon}\one$ by Lemma \ref{lem:lb_gadget_opt_gain}.

    Fixing any two distinct states $s_0, s$, we will now exhibit a policy $\pi$ such that $\E^{P_\pi^{(i, a)}}_{s_0} [\tau_s] \leq 5D$. If $s_0, s$ are of the form $s_0 = \ts_j$, $s = \ts_{j'}$, then by (the proof of) Lemma \ref{lem:lb_gadget_diam_bound}, there are choices of $\pi(\ts_j)$ and $\pi(\bs_{j'})$ such that $\E_{s_0}^{P^{(i,a)}_\pi}[\tau_{\bs_j}] \leq 2D$ and $\E_{\bs_{j'}}^{P^{(i,a)}_\pi}[\tau_{\ts_{j'}}] \leq 2D$ (in fact any choice of action in $\A_{/2}$ suffices). Also, as argued above there exists a path of length at most $D$ from $\bs_j$ to $\bs_{j'}$, so we can choose $\pi$ on $\bar{\S}$ so that $\E_{\bs_{j}}^{P^{(i,a)}_\pi}[\tau_{\bs_{j'}}] \leq D$. (The path constructed above may have cycles, which is not directly compatible with a stationary policy, but the cycles can be removed without increasing the hitting time.) Then by the strong Markov property we will immediately have that $\E^{P_\pi^{(i, a)}}_{s_0} [\tau_s] \leq 5D$. Furthermore this same policy can easily be seen to have $\E_{\bs_{j}}^{P^{(i,a)}_\pi}[\tau_{\ts_{j'}}] \leq 3D, \E_{\ts_{j}}^{P^{(i,a)}_\pi}[\tau_{\bs_{j'}}] \leq 3D$ as well, so the same expected hitting time bound holds even if one or both of $s_0, s$ are members of $\bar{\S}$ rather than $\widetilde{\S}$.
\end{proof}

\subsection{Decoder}
\label{sec:lb_decoder}

Now we define the decoder $\Phi : \PiSR \to \{1, \dots, S/2\} \times \A_{/2}$.
For any policy $\pi \in \PiSR$, if $\pi$ is unichain in $P^{(0)}$, and furthermore if there exists a unique $i \in \{1, \dots, S/2\}, a \in \A_{/2}$ such that $y^{\pi}_{\M^{(0)}}(\bs_i,a) > \frac{1}{3}$, then let $\Phi(\pi) = (i,a)$.
If these conditions are not satisfied, then $\Phi$ can output $(1,1)$ (the value for this case does not matter and can be chosen arbitrarily).

\begin{lemma}
\label{lem:occupancy_lb_near_opt_pol}
    Let $\varepsilon' \in (0,\varepsilon/8]$, let $i^\star \in \{1, \dots, S/2\}, a^\star \in \A_{/2}$, and let $\pi \in \PiSR$ be a stationary randomized policy. If $\pi$ is $\varepsilon'$-gain-optimal in $\M^{(i^\star a^\star)}$, meaning $\infnorm{\rho^{\star}_{\M^{(i^\star a^\star)}} - \rho^{\pi}_{\M^{(i^\star a^\star)}} } \leq \varepsilon'$, then $\pi$ must be unichain in the MDP $\M^{(i^\star a^\star)}$, and furthermore the state-action occupancy $y^\pi_{\M^{(i^\star a^\star)}}$ satisfies
    \begin{equation*}
        y^\pi_{\M^{(i^\star a^\star)}}(\bs^\star, a^\star) \geq  \frac{2 - \varepsilon}{4 - \varepsilon} - 2(2-\varepsilon)\frac{\varepsilon'}{\varepsilon}
    \end{equation*}
    where $\bs^\star := \bs_{i^\star}$.
\end{lemma}
\begin{proof}
    First we let $y \in \R^{\S \times \A}$ be any invariant distribution for the chain over state-action pairs induced by policy $\pi$, meaning that, letting $\mu(s) := \sum_{a \in \A}y(s,a)$ denote the corresponding distribution over states, we have
    \begin{equation}
        \mu(s) = \sum_{s_0 \in \S} \sum_{a \in \A} y(s_0, a) P^{(i^\star a^\star)}(s \mid s_0, a), \qquad y(s,a) = \mu(s) \pi(a \mid s). \label{eq:decoder_invariant_eqn}
    \end{equation}
    Specializing this first equation to $\ts^\star := \ts_{i^\star}$, we have
    \begin{align*}
        \mu(\ts^\star) &= \sum_{a \in \A} y(\bs^\star, a) P^{(i^\star a^\star)}(\ts^\star \mid \bs^\star, a) + \sum_{a \in \A} y(\ts^\star, a) P^{(i^\star a^\star)}(\ts^\star \mid \ts^\star, a) \\
        &= \pz \sum_{a \in \A_{/2} : a \neq a^\star} y(\bs^\star, a) + \ps y(\bs^\star, a^\star) + (1-\pz)\mu(\ts^\star) \\
        &=\pz \sum_{a \in \A_{/2}} y(\bs^\star, a) + \left(\ps - \pz \right) y(\bs^\star, a^\star) + (1-\pz)\mu(\ts^\star)  
    \end{align*}
    since $\bs^\star$ and $\ts^\star$ are the only states that transition into $\ts^\star$. After rearranging we obtain
    \begin{equation}
        \mu(\ts^\star) = \sum_{a \in \A_{/2}} y(\bs^\star, a) + \frac{\ps - \pz}{\pz} y(\bs^\star, a^\star). \label{eq:lb_mu_ts_star_lb_by_y_bs_star}
    \end{equation}
    Also repeating the same calculation for $i \neq i^\star$ yields
    \begin{equation*}
        \mu(\ts_i) = \sum_{a \in \A_{/2}} y(\bs_i, a),
    \end{equation*}
    so after summing over $i$ we obtain
    \begin{equation}
        \sum_i \mu(\ts_i) = \frac{\ps - \pz}{\pz} y(\bs^\star, a^\star) + \sum_i \sum_{a \in \A_{/2}} y(\bs_i, a). \label{eq:ts_states_key_equality} 
    \end{equation}
    
    Since the total mass is equal to $1$, we also have
    \begin{equation*}
        \sum_i \mu(\ts_i) + \sum_i \sum_{a \in \A_{/2}} y(\bs_i, a) \leq 1.
    \end{equation*}
    Rearranging and combining with~\eqref{eq:ts_states_key_equality}, we obtain that
    \begin{align*}
        \sum_i \mu(\ts_i) \leq \frac{\ps - \pz}{\pz} y(\bs^\star, a^\star) + 1 - \sum_i \mu(\ts_i),
    \end{align*}
    which implies
    \begin{align}
        \sum_i \mu(\ts_i) &\leq \frac{1}{2} + \frac{1}{2}\frac{\ps - \pz}{\pz} y(\bs^\star, a^\star) = \frac{1}{2} + \frac{\varepsilon}{4-2\varepsilon}y(\bs^\star, a^\star). \label{eq:sum_mu_key_upper_bd}
    \end{align}

    Now we use $\varepsilon'$-optimality to lower-bound $\sum_i \mu(\ts_i)$.
    Since $\mu$ is invariant for $P_\pi^{(i^\star a^\star)}$ and thus also $\mu (P_\pi^{(i^\star a^\star)})^{\infty} = \mu$, we have that $\mu r_\pi = \mu (P_\pi^{(i^\star a^\star)})^{\infty} r_\pi = \mu \rho_{\M^{(i^\star a^\star)}}^{\pi}$. Since the only nonzero entries of $r_\pi$ are those corresponding to states of the form $\ts_i$, we furthermore have that
    \begin{equation*}
        \sum_i \mu(\ts_i) = \mu r_\pi = \mu \rho_{\M^{(i^\star a^\star)}}^{\pi}.
    \end{equation*}
    Since $\pi$ is $\varepsilon'$-optimal, since by Lemma \ref{lem:lb_full_mdps_properties} the optimal gain is at least $\frac{2}{4-\varepsilon}$, if we sample a starting state from distribution $\mu$ we would have that
    \begin{equation*}
        \sum_s \mu(s)  \rho^{\pi}_{\M^{(i^\star a^\star)}}(s) \geq \sum_s \mu(s)  (\rho^{\star}_{\M^{(i^\star a^\star)}}(s) - \varepsilon') \geq \frac{2}{4-\varepsilon} - \varepsilon'.
    \end{equation*}
    Combining with the previous display then yields that
    \begin{equation*}
        \sum_i \mu(\ts_i) \geq \frac{2}{4-\varepsilon} - \varepsilon'.
    \end{equation*}

    Combining this lower bound with the upper bound~\eqref{eq:sum_mu_key_upper_bd}, we obtain
    \begin{align*}
        \frac{2}{4-\varepsilon} - \varepsilon' \leq \sum_i \mu(\ts_i) \leq \frac{1}{2} + \frac{\varepsilon}{4-2\varepsilon}y(\bs^\star, a^\star)
    \end{align*}
    which after rearranging implies that
    \begin{equation*}
        y(\bs^\star, a^\star) \geq \frac{2 - \varepsilon}{4 - \varepsilon} - 2(2-\varepsilon)\frac{\varepsilon'}{\varepsilon}.
    \end{equation*}

    Finally, it is possible to check that as long as $\varepsilon' \leq \varepsilon/8$ this quantity is strictly positive, which implies that $(\bs^\star, a^\star)$ is a member of all recurrent classes, and thus that $\pi$ is unichain in $\M^{(i^\star a^\star)}$.
\end{proof}

Now we consider the behavior of policy $\pi$ in the MDP $\M^{(0)}$, since it is used in the decoder.
\begin{lemma}
\label{lem:dist_transfer_to_null_mdp}
    For any $i^\star \in \{1, \dots, S/2\}, a^\star \in \A_{/2}$, and any stationary randomized policy $\pi \in \PiSR$, if $\pi$ is unichain in $\M^{(i^\star a^\star)}$ then it is also unichain in $\M^{(0)}$ and we have 
    \begin{equation*}
        y^\pi_{\M^{(0)}}(\bs^\star, a^\star) \geq y^\pi_{\M^{(i^\star a^\star)}}(\bs^\star, a^\star).
    \end{equation*}
\end{lemma}
\begin{proof}
    First, since $\M^{(0)}$ and $\M^{(i^\star a^\star)}$ have the same transition supports for all state-action pairs, if $\pi$ is unichain in $\M^{(i^\star a^\star)}$ then it will also be unichain in $\M^{(0)}$.
    
    Next, for notational convenience abbreviate $y := y^\pi_{\M^{(i^\star a^\star)}} $ and $y' :=  y^\pi_{\M^{(0)}}$, and also $\mu(s) := \sum_{a \in \A} y(s,a)$ for all $s \in \S$.
    
    Note that the desired conclusion holds immediately if $y(\bs^\star, a^\star) = 0$, so from here on we may assume that it is strictly positive. The arguments from Lemma \ref{lem:occupancy_lb_near_opt_pol}, specifically~\eqref{eq:lb_mu_ts_star_lb_by_y_bs_star}, then imply that $\mu(\ts^\star) > 0$, which is needed for the next definition.
    We claim that it is sufficient to show that the unnormalized distribution
    \begin{equation*}
        \tilde{y}(s,a) := \begin{cases}
            y(s,a) & s \neq \ts^\star \\
            \frac{\pz \mu(\ts^\star) - (\ps - \pz) y(\bs^\star, a^\star)}{\pz \mu(\ts^\star)}y(\ts^\star, a) & s = \ts^\star
        \end{cases}
    \end{equation*}
    is invariant for $\M^{(0)}$ under $\pi$, because then by unichainedness and after normalizing, since
    \begin{align*}
        \sum_{s,a}\tilde{y}(s,a) &= \sum_{s\neq \ts^\star,a}y(s,a) + \sum_a \frac{\pz \mu(\ts^\star) - (\ps - \pz) y(\bs^\star, a^\star)}{\pz \mu(\ts^\star)}y(\ts^\star, a) \\
        &= \sum_{s\neq \ts^\star,a}y(s,a) + \frac{\pz \mu(\ts^\star) - (\ps - \pz) y(\bs^\star, a^\star)}{\pz \mu(\ts^\star)} \mu(\ts^\star) \\
        &= \sum_{s\neq \ts^\star,a}y(s,a) + \mu(\ts^\star) - \frac{\ps - \pz}{\pz} y(\bs^\star, a^\star) \leq 1,
    \end{align*}
    we will have (specializing to the coordinate $(\bs^\star, a^\star)$) that
    \begin{equation*}
        y'(\bs^\star, a^\star)=  \frac{ \tilde{y}(\bs^\star, a^\star) }{\sum_{s,a}\tilde{y}(s,a)} = \frac{ y(\bs^\star, a^\star) }{\sum_{s,a}\tilde{y}(s,a)} \geq y(\bs^\star, a^\star).
    \end{equation*}
    We also note that $\tilde{y}$ is indeed nonnegative because
    \begin{equation*}
        \frac{\pz \mu(\ts^\star) - (\ps - \pz)y(\bs^\star, a^\star)}{\pz \mu(\ts^\star)} = 1 - \frac{\ps - \pz}{\pz} \frac{y(\bs^\star, a^\star)}{\mu(\ts^\star)} \geq 1-1=0
    \end{equation*}
    where the final inequality is because $\mu(\ts^\star) \geq \frac{\ps - \pz}{\pz}y(\bs^\star, a^\star)$ as a consequence of~\eqref{eq:lb_mu_ts_star_lb_by_y_bs_star}.

    Now we check the invariance of $\tilde{y}$. We have that
    \begin{align*}
        \sum_{a \in \A} \tilde{y}(\bs^\star, a) &= \sum_{a \in \A} y(\bs^\star, a) \\
        &\stackrel{(i)}{=} \sum_{s_0 \in \S} \sum_{a \in \A} y(s_0, a) P^{(i^\star, a^\star)}(\bs^\star \mid s_0, a) \\
        &\stackrel{(ii)}{=} y(\bs^\star, a^\star) (\pz - \ps) + \sum_{s_0 \in \S} \sum_{a \in \A} y(s_0, a) P^{(0)}(\bs^\star \mid s_0, a) \\
        &= y(\bs^\star, a^\star) (\pz - \ps) + \sum_{a \in \A} y(\ts^\star, a) \pz + \sum_{s_0 \in \S : s_0 \neq \ts^\star} \sum_{a \in \A} y(s_0, a) P^{(0)}(\bs^\star \mid s_0, a) \\
        & \stackrel{(iii)}{=} \sum_{a \in \A} \tilde{y}(\ts^\star, a) \pz + \sum_{s_0 \in \S : s_0 \neq \ts^\star} \sum_{a \in \A} y(s_0, a) P^{(0)}(\bs^\star \mid s_0, a) \\
        &= \sum_{s_0 \in \S} \sum_{a \in \A} \tilde{y}(s_0, a) P^{(0)}(\bs^\star \mid s_0, a)
    \end{align*}
    where $(i)$ is by stationarity of $y$, $(ii)$ is because $P^{(i^\star, a^\star)}(\bs^\star \mid \bs^\star, a^\star) = 1-\ps = 1 - \pz + (\pz - \ps) = P^{(0)}(\bs^\star \mid \bs^\star, a^\star) + (\pz - \ps)$, and $(iii)$ is because summing the definition of $\tilde{y}(\ts^\star, \cdot)$ and multiplying by $\pz$ yields that
    \begin{equation}
        \pz \sum_a \tilde{y}(\ts^\star,a) = \pz \sum_a \frac{\pz \mu(\ts^\star) - (\ps - \pz) y(\bs^\star, a^\star)}{\pz \mu(\ts^\star)}y(\ts^\star, a) = \pz \mu(\ts^\star) - (\ps - \pz) y(\bs^\star, a^\star). \label{eq:decoder_ref_mdp_monotonicity_stat_1}
    \end{equation}
    
    Using this above calculation~\eqref{eq:decoder_ref_mdp_monotonicity_stat_1} again, we also have that
    \begin{align*}
        \sum_{a \in \A} \tilde{y}(\ts^\star, a) &= \mu(\ts^\star) - \frac{\ps - \pz}{\pz} y(\bs^\star, a^\star) \\
        &\stackrel{(i)}{=} \sum_{a \in \A} y(\ts^\star, a) P^{(i^\star, a^\star)}(\ts^\star \mid \ts^\star, a) + \sum_{a \in \A} y(\bs^\star, a) P^{(i^\star, a^\star)}(\ts^\star \mid \bs^\star, a) - \frac{\ps - \pz}{\pz} y(\bs^\star, a^\star) \\
        &\stackrel{(ii)}{=} (1- \pz)\sum_{a \in \A} y(\ts^\star, a)  + \sum_{a \in \A} y(\bs^\star, a) P^{(0)}(\ts^\star \mid \bs^\star, a) + (\ps - \pz)y(\bs^\star, a^\star)\\
        & \qquad - \frac{\ps - \pz}{\pz} y(\bs^\star, a^\star) \\
        &= (1-\pz) \frac{\pz\mu(\ts^\star) - (\ps - \pz) y(\bs^\star, a^\star) }{\pz} + \sum_{a \in \A} y(\bs^\star, a) P^{(0)}(\ts^\star \mid \bs^\star, a) \\
        & \stackrel{(iii)}{=} (1-\pz) \sum_{a \in \A} \tilde{y}(\ts^\star, a) + \sum_{a \in \A} y(\bs^\star, a) P^{(0)}(\ts^\star \mid \bs^\star, a) \\
        &= \sum_{a \in \A} \tilde{y}(\ts^\star, a)P^{(0)}(\ts^\star \mid \ts^\star, a) + \sum_{a \in \A} y(\bs^\star, a) P^{(0)}(\ts^\star \mid \bs^\star, a)
    \end{align*}
    where $(i)$ is by stationarity of $y$, $(ii)$ uses the definitions of $P^{(0)}$ and $P^{(i^\star, a^\star)}$, and $(iii)$ uses the calculation~\eqref{eq:decoder_ref_mdp_monotonicity_stat_1} again.

    For all other states $s \not \in \{\ts^\star, \bs^\star\}$, we have $\tilde{y}(s, \cdot) = y(s, \cdot)$ and $P^{(i^\star, a^\star)}(\cdot \mid s, \cdot) = P^{(0)}(\cdot \mid s, \cdot)$. Also it is immediate from the definition of $\tilde{y}$ that for all $s \in \S, a \in \A$,
    \begin{equation*}
        \tilde{y}(s,a) = \left(\sum_{a' \in \A} \tilde{y}(s,a')\right) \pi(a \mid s)
    \end{equation*}
    because the analogous condition holds for $y$ and $\tilde{y}$ rescales all actions within each state by the same amount. Therefore we have confirmed that $\tilde{y}$ is an invariant distribution as desired. 
\end{proof}

Finally we can confirm the correctness of the defined decoder $\Phi$.
\begin{lemma}
\label{lem:decoder_correctness}
    Suppose $\varepsilon \leq 1/4$. Let $i^\star \in \{1, \dots, S/2\}, a^\star \in \A_{/2}$, and let $\pi \in \PiSR$ be a stationary randomized policy. If $\pi$ is $\varepsilon/32$-gain-optimal in $\M^{(i^\star a^\star)}$, meaning $\infnorm{\rho^{\star}_{\M^{(i^\star a^\star)}} - \rho^{\pi}_{\M^{(i^\star a^\star)}} } \leq \varepsilon/32$, then $\Phi(\pi) = (i^\star, a^\star)$.
\end{lemma}
\begin{proof}
    First, supposing that $\pi$ is unichain, we argue that if there is any $i \in \{1, \dots, S/2\}, a \in \A_{/2}$ such that $y^{\pi}_{\M^{(0)}}(\bs_i,a) > \frac{1}{3}$, then it must be unique. By arguments similar to the previous lemmas it is easy to see that we must have $\sum_{a'} y^\pi(\ts_i, a') \geq y^{\pi}_{\M^{(0)}}(\bs_i,a)$, so this already accounts for strictly more than $2/3$ of the probability mass, and thus no remaining $(\bs_j, a'' )\neq (\bs_i,a) $ can have $y^{\pi}_{\M^{(0)}}(\bs_j,a'') > \frac{1}{3}$.

    Now we argue that whenever $\pi$ is $\varepsilon/32$-optimal in $\M^{(i^\star a^\star)}$, $\pi$ will be unichain in $\M^{(0)}$ and we will have $y^{\pi}_{\M^{(0)}}(\bs_{i^\star},a^\star) > 1/3$.
    This essentially just amounts to simply applying Lemmas \ref{lem:occupancy_lb_near_opt_pol} and \ref{lem:dist_transfer_to_null_mdp} and choosing $\varepsilon'$. Setting $\varepsilon' = \varepsilon/32$, we have from Lemma \ref{lem:occupancy_lb_near_opt_pol} that $\pi$ is unichain in $\M^{(i^\star, a^\star)}$ and
    \begin{equation*}
        y^\pi_{\M^{(i^\star a^\star)}}(\bs^\star, a^\star) \geq  \frac{2 - \varepsilon}{4 - \varepsilon} - (2-\varepsilon)\frac{1}{16} \geq \frac{2-\varepsilon}{4-\varepsilon} -\frac{1}{8} \geq \frac{7}{15} -\frac{1}{8} > .34 > 1/3
    \end{equation*}
    where we used that $\frac{2-\varepsilon}{4-\varepsilon}$ is minimized over $(0, 1/4]$ at $1/4$, where it has value $7/15$. Then by Lemma \ref{lem:dist_transfer_to_null_mdp} we have
    \begin{equation*}
        y^\pi_{\M^{(0)}}(\bs^\star, a^\star) \geq y^\pi_{\M^{(i^\star a^\star)}}(\bs^\star, a^\star) > 1/3
    \end{equation*}
    as desired. Thus by definition of the decoder $\Phi$, we will have that $\Phi(\pi) = (i^\star, a^\star)$.
\end{proof}

\subsection{Proof of Theorem \ref{thm:avg_rew_lower_bound}}
\label{sec:completing_proof_of_avg_rew_lower_bound}
For notational simplicity, for this final part of the argument we let $m = SA/4$ and re-index the constructed MDP transition kernels $(P^{(i, a)})_{i \in \{1, \dots, S/2\}, a \in \A_{/2}}$ as $(P^{(j)})_{j=1,\dots, m}$. For each $j \in \{1, \dots, m\}$ we let $\P_j$ denote the probability measure over the observed samples where, for each $s \in \S, a \in \A$, the $(S^{(i)}_{sa})_{1 \leq i \leq n(sa)}$ are sampled independently from $P^{(j)}(\cdot \mid s, a)$. Also for each $j \in \{1, \dots, m\}$ we define the event $E_j := \{\Phi(\Alg) = j\}$.

Since the output of $\Alg$ is assumed to be $\varepsilon/32$-optimal with probability at least $1-\delta$ when used on any of the MDPs $(P^{(j)})_{j=1,\dots, m}$, we have for all $j \in \{1, \dots, m\}$ that
\begin{equation}
    \P_j(E_j) \geq 1-\delta \label{eq:lb_main_pf_1}
\end{equation}
and consequently also by disjointness of the events $(E_k)_k$, for all $k \neq j$,
\begin{equation}
    \sum_k \P_j(E_k) \leq 1 \implies \sum_{k \neq j} \P_j(E_k) \leq 1 - \P_j (E_j) \leq \delta. \label{eq:lb_main_pf_2}
\end{equation}
Therefore
\begin{align*}
    \kl{\P_k}{\P_j} &\geq \kl{\Ber(\P_k(E_k))}{\Ber(\P_j(E_k))} = \klb(\P_k(E_k),\P_j(E_k)) \\
    & \geq \klb(1-\delta, \P_j(E_k))
\end{align*}
where the first inequality is by the data processing inequality, and the second inequality uses~\eqref{eq:lb_main_pf_1}. 
Now for any fixed $j$, summing over the $k \neq j$ we have that
\begin{align*}
    \sum_{k \neq j}\kl{\P_k}{\P_j} & \geq \sum_{k \neq j} \klb(1-\delta, \P_j(E_k)) \\
    & \geq (m-1) \klb\left(1-\delta, \frac{1}{m-1}\sum_{k \neq j}\P_j(E_k)\right) \\
    & \geq (m-1) \klb\left(1-\delta, \frac{\delta}{m-1}\right) \\
    & \geq (m-1) \left( (1-\delta) \log (\frac{m-1}{\delta}) - h(\delta)\right) \\
    & \geq (m-1) \left( \frac{3}{4} \log (\frac{m-1}{\delta}) - h(1/4)\right) \\
    & \geq \frac{m-1}{3}\log\left( \frac{m-1}{\delta} \right) \geq \frac{m-1}{3}\log\left( \frac{m}{2\delta} \right)
\end{align*}
by using convexity and Jensen's inequality, then~\eqref{eq:lb_main_pf_2}, then Lemma \ref{lem:bernoulli_kl_lb_2}, then that $\delta \leq \frac{1}{4}$, and finally that $h(1/4) < (\frac{3}{4} - \frac{1}{3})\log(4)$. Now summing over $j$, we have that
\begin{equation}
    \sum_{j}\sum_{k \neq j}\kl{\P_k}{\P_j} \geq \frac{m(m-1)}{3}\log\left( \frac{m}{2\delta} \right). \label{eq:lb_main_pf_3}
\end{equation}

Next, for any $k\neq j$, because $\P_k$ and $\P_j$ differ only in the distribution of the samples drawn from the $k$th and $j$th state-action rows out of the list $\left((\bs_i, a)\right)_{i \in \{1, \dots, S/2\}, a \in \A_{/2}}$, letting $n_k$ and $n_j$ be the corresponding numbers of samples from these state-action pairs, we have that
\begin{align*}
    \kl{\P_k}{\P_j} &= n_j \kl{P^{(k)}_{j}}{P^{(j)}_j} + n_k \kl{P^{(k)}_{k}}{P^{(j)}_k} \\
    &= n_j \klb(\pz, \ps) + n_k \klb(\ps, \pz) \\
    & \leq (n_j + n_k) \frac{\varepsilon^2}{3D}
\end{align*}
using Lemma \ref{lem:lb_gadget_kl_ub} for the final inequality.
This implies
\begin{equation*}
    \sum_{j}\sum_{k \neq j}\kl{\P_k}{\P_j} \leq \frac{\varepsilon^2}{3D}\sum_{j}\sum_{k \neq j}(n_j + n_k) = \frac{\varepsilon^2}{3D}2(m-1) \sum_{j=1}^m n_j.
\end{equation*}

Combining with~\eqref{eq:lb_main_pf_3}, we have
\begin{equation*}
    \frac{\varepsilon^2}{3D}2(m-1) \sum_{j=1}^m n_j \geq \sum_{j}\sum_{k \neq j}\kl{\P_k}{\P_j} \geq \frac{m(m-1)}{3}\log\left( \frac{m}{2\delta} \right)
\end{equation*}
so after rearranging,
\begin{equation*}
    \sum_{j=1}^m n_j \geq m \frac{1}{2} \frac{D}{\varepsilon^2}\log\left( \frac{m}{2\delta} \right) = SA \frac{1}{8} \frac{D}{\varepsilon^2}\log\left( \frac{SA}{8\delta} \right).
\end{equation*}
This completes the proof of Theorem \ref{thm:avg_rew_lower_bound}.

\subsection{Auxiliary Lemmas}

The following lemma is standard and follows immediately from the fact that the KL divergence is upper-bounded by the $\chi^2$ divergence.
\begin{lemma}
    \label{lem:bernoulli_kl_ub}
    For $p, q \in (0,1)$, we have
    \begin{equation*}
        \klb(p,q) \leq \frac{(p-q)^2}{q(1-q)}.
    \end{equation*}
\end{lemma}

We also give a simple lower bound.
\begin{lemma}
    \label{lem:bernoulli_kl_lb_2}
    For $p, q \in (0,1)$ and letting $h(p) = -p \log p - (1-p) \log(1-p)$, we have
    \begin{equation*}
        \klb(p,q) \geq p \log(1/q) - h(p).
    \end{equation*}
\end{lemma}
\begin{proof}
We calculate
    \begin{align*}
        \klb(p,q) &= p \log p + p \log (1/q) + (1-p) \log(1-p) + (1-p) \log(1/(1-q)) \\
        &= p \log (1/q) - h(p) + (1-p) \log(1/(1-q)) \\
        & \geq p \log (1/q) - h(p).
    \end{align*}
\end{proof}

\section{Discounted Lower Bounds}
\label{sec:discounted_lower_bounds}

In this section we discuss the gap in the lower bound of \citet[Theorem 3]{azar_minimax_2013} and prove our Theorem \ref{thm:lower_bound_discounted}.

\citet[Theorem 3]{azar_minimax_2013} states an $\Omega(SA (1-\gamma)^{-3} \varepsilon^{-2} \log(SA/\delta))$ sample complexity lower bound for estimating the $Q^\star$ function to uniform accuracy $\varepsilon$. However there are some gaps in the proof, including within \citet[Lemma 18]{azar_minimax_2013}. First, they claim that it suffices to consider estimators $\Qhat^\star$ of $Q^\star$ for which $\Qhat^\star(s,a)$ only depends on certain transition samples associated with $(s,a)$. This does not generally follow even when transitions sampled from different $(s,a)$ are independent. In particular for their instances as stated, a certain parameter which determines $Q^\star$ is shared between state-action pairs, making samples from different state-action pairs informative about other $Q^\star$ values.
They also assert for their instances that certain $Q^\star$ estimation error events are independent across state-action rows after conditioning on the number of samples from each row. Even for the aforementioned restricted form of state-action $Q^\star$ estimators, conditioning on the numbers of samples can introduce dependencies under general adaptive sampling.

For these reasons, we give the self-contained lower bound Theorem \ref{thm:lower_bound_discounted}, which applies to policy optimization under nonadaptive sampling.

Now we prove Theorem \ref{thm:lower_bound_discounted}.
\begin{proof}[Proof of Theorem \ref{thm:lower_bound_discounted}]
    Let $\A = \{1, \dots, A\}$.
    Let $s_\bot$ be a reward-zero absorbing state and we refer to the remaining $S-1$ states $s_1, \dots, s_{S-1}$ as reward-one states, as all of their actions have reward $1$, that is,
    \begin{equation*}
        r(s,a) = \begin{cases}
            1 & s \neq s_\bot \\
            0 & s = s_\bot
        \end{cases}.
    \end{equation*}
    We let $\S_1 = \{s_1, \dots, s_{S-1}\}$ denote the set of reward-one states.
    Let $\varepsilon' := (1-\gamma)^2 \varepsilon$.
    For $s, s' \neq s_\bot$ and $a \neq 1$, we define
    \begin{equation*}
        P^{(s,a)}_{s'a'} = \begin{cases}
            (\gamma + \frac{\varepsilon'}{2}) e_{s'}^\top  + (1 - \gamma - \frac{\varepsilon'}{2}) e_{s_{\bot}}^\top & a' = 1 \\
            \gamma e_{s'}^\top + (1-\gamma) e_{s_{\bot}}^\top & a' \neq 1, (s',a') \neq (s,a)\\
            (\gamma + \varepsilon') e_{s'}^\top  + (1 - \gamma - \varepsilon') e_{s_{\bot}}^\top & a' \neq 1, (s',a') = (s,a) \\
        \end{cases}.
    \end{equation*}
    We let $\M^{(s,a)}$ denote the MDP $(\S, \A, P^{(s,a)}, r)$.
    Now we describe this definition in words. The pair $(s,a)$ is a planted state-action. The next-state transition distributions for all reward-one states (not equal to $s_\bot$) have a similar form for all actions, where there is a certain probability of a self-transition and the remaining probability is a transition to the absorbing reward-zero state $s_\bot$; the actions only differ in their probability of the self-transition.
    All reward-one states have a ``baseline'' action $1$ which self-transitions with probability $\gamma+\frac{\varepsilon'}{2}$, all other non-planted actions self-transition with probability $\gamma$, and if that state happens to contain the planted action, then that action has self-transition probability $\gamma + \varepsilon'$.
    We also note that this is a valid definition because
    \begin{equation*}
        \gamma + \varepsilon' = \gamma + (1-\gamma)^2 \varepsilon \leq \gamma + (1-\gamma)/2 < \gamma + (1-\gamma) =1.
    \end{equation*}

    Now we compute the value functions of all policies in all states. Fix a state $s \neq s_\bot$, a stationary randomized policy $\pi$, and also $s^\star \neq s_\bot$, $a^\star \neq 1$. From the Bellman equation and the fact that  $s_\bot$ has value zero for any policy we have $V^\pi_{\M^{(s^\star, a^\star)}}(s) = 1 + \gamma (P^{(s^\star, a^\star)}_\pi)_{ss} V^\pi_{\M^{(s^\star, a^\star)}}(s)$. Abbreviating $v := V^\pi_{\M^{(s^\star, a^\star)}}(s), q = (P^{(s^\star, a^\star)}_\pi)_{ss}$, we have
    \begin{equation*}
        v = 1 + \gamma q v = \frac{1}{1-\gamma q}.
    \end{equation*}
    Also using that $q \in [\gamma, \gamma + \varepsilon']$ and viewing $v$ as a function of $q$, we have
    \begin{equation*}
        v'(q) = \frac{\gamma}{(1-\gamma q)^2} \geq \frac{\gamma}{(1-\gamma^2)^2} = \frac{\gamma}{(1-\gamma)^2 (1+\gamma)^2} \geq \frac{1}{(1-\gamma)^2} \frac{1/2}{(3/2)^2} = \frac{2}{9}\frac{1}{(1-\gamma)^2}
    \end{equation*}
    where we used that $1 - \gamma^2 = (1-\gamma) (1+\gamma)$. Therefore by the mean value theorem,
    \begin{equation*}
        v(\gamma + \varepsilon') - v(q) \geq \frac{2}{9}\frac{1}{(1-\gamma)^2} (\gamma + \varepsilon'-q)
    \end{equation*}
    and if $q \leq \gamma + \frac{\varepsilon'}{2}$ then also
    \begin{equation*}
        v(\gamma + \varepsilon'/2) - v(q) \geq \frac{2}{9}\frac{1}{(1-\gamma)^2} (\gamma + \varepsilon'/2-q).
    \end{equation*}
    If $s^\star = s$, so $s$ contains a planted action, then clearly the planted action $a^\star$ is optimal and we have $V_{\M^{(s^\star, a^\star)}}^\star(s) = v(\gamma + \varepsilon')$ and also
    \begin{equation*}
        \gamma + \varepsilon' - q = \gamma + \varepsilon' - (P^{(s^\star, a^\star)}_{\pi})_{ss} \geq \frac{\varepsilon'}{2} \sum_{a \neq a^\star} \pi(a \mid s),
    \end{equation*}
    so
    \begin{equation*}
        V_{\M^{(s^\star, a^\star)}}^\star(s) - V^\pi_{\M^{(s^\star, a^\star)}}(s) \geq \frac{2}{9}\frac{1}{(1-\gamma)^2} \frac{\varepsilon'}{2} \sum_{a \neq a^\star} \pi(a \mid s) = \frac{\varepsilon}{9} \sum_{a \neq a^\star} \pi(a \mid s).
    \end{equation*}
    Otherwise if $s \neq s^\star, s_\bot$, then action $1$ is optimal in state $s$ and we have $V_{\M^{(s^\star, a^\star)}}^\star(s) = v(\gamma + \frac{\varepsilon'}{2})$. Then $q \leq \gamma + \frac{\varepsilon'}{2}$ and
    \begin{equation*}
        \gamma + \varepsilon'/2 - q = \gamma + \varepsilon'/2 - (P^{(s^\star, a^\star)}_{\pi})_{ss} \geq \frac{\varepsilon'}{2} \sum_{a \neq 1} \pi(a \mid s)
    \end{equation*}
    so
    \begin{equation*}
        v(\gamma + \varepsilon'/2) - v(q) \geq \frac{2}{9}\frac{1}{(1-\gamma)^2} \frac{\varepsilon'}{2} \sum_{a \neq 1} \pi(a \mid s) = \frac{\varepsilon}{9} \sum_{a \neq 1} \pi(a \mid s).
    \end{equation*}

    Therefore, within any MDP $\M^{(s^\star,a^\star)}$ for any $s^\star \neq s_\bot$, $a^\star \neq 1$, if $\pi$ does not place probability mass $\geq \frac{2}{3}$ into the optimal actions of all states for that MDP, the suboptimality of $\pi$ will be at least $\frac{\varepsilon}{27}$ in some state.

    Now we define a decoder $\Phi : \PiSR \to \S_1\times \{2, \dots, A\}$ where $\Phi(\pi)$ is the first (in an arbitrary fixed ordering) state-action pair $(s,a)$ in $\S_1$ such that $a \neq 1$ and $\pi(a \mid s) \geq \frac{2}{3}$, and if no such pair exists, for completeness return $(s_1, 2)$. 
    We also re-index the MDPs as $\M_1, \dots, \M_m$ for $m := (S-1) (A-1) = |\S_1| |\{2, \dots, A\}|$ and write $\P_j$ for the measure over observed samples when drawn from MDP $\M_j$. Define the event $E_j = \{\Phi(\Alg) = j\}$.
    By our above arguments and the first assumption of the theorem we then have that $\P_j(E_j) \geq 1 - \delta$ for all $j = 1, \dots, m$.

    By the same arguments as in Appendix \ref{sec:completing_proof_of_avg_rew_lower_bound}, we have
    \begin{align*}
        m\frac{m-1}{3}\log\left( \frac{m}{2\delta} \right) \leq \sum_j \sum_{k \neq j} \kl{\P_k}{\P_j}
        &\leq \sum_j \sum_{k \neq j} \left(n_j \kl{P^{(k)}_{j}}{P^{(j)}_j} + n_k \kl{P^{(k)}_{k}}{P^{(j)}_k}\right) \\
        &= \sum_j \sum_{k \neq j} \left( n_j \klb(\gamma, \gamma + \varepsilon') + n_k \klb(\gamma + \varepsilon', \gamma) \right).
    \end{align*}

    We can also use Lemma \ref{lem:bernoulli_kl_ub} to bound
    \begin{equation*}
        \klb(\gamma + \varepsilon',\gamma) \leq \frac{\varepsilon'^2}{\gamma (1-\gamma)} \leq \frac{2\varepsilon'^2}{1-\gamma}, \qquad \klb(\gamma, \gamma + \varepsilon') \leq \frac{\varepsilon'^2}{(\gamma + \varepsilon') (1-\gamma - \varepsilon')} \leq \frac{4\varepsilon'^2}{1-\gamma}
    \end{equation*}
    where we used that
    \begin{equation*}
        1-\gamma - \varepsilon' = 1 - \gamma - (1-\gamma)^2 \varepsilon = (1-\gamma) (1-(1-\gamma)\varepsilon) \geq (1-\gamma) \frac{1}{2}
    \end{equation*}
    since $\varepsilon \leq \frac{1}{2}\frac{1}{1-\gamma}$.

    Combining and rearranging, we have that
    \begin{align*}
        &m\frac{m-1}{3}\log\left( \frac{m}{2\delta} \right) \leq \frac{4 \varepsilon'^2}{1-\gamma} \sum_j \sum_{k \neq j} (n_j + n_k) = \frac{4 \varepsilon'^2}{1-\gamma} 2(m-1) \sum_{j=1}^m n_j \\
        \implies & \sum_{j=1}^m n_j \geq \frac{m}{24} \frac{1-\gamma}{\varepsilon'^2} \log\left( \frac{m}{2\delta} \right) = \frac{m}{24} \frac{1}{(1-\gamma)^3 \varepsilon^2} \log\left( \frac{m}{2\delta} \right).
    \end{align*}
    Finally we complete the theorem by noting that $m = (S-1)(A-1) \geq \frac{SA}{4}$.
\end{proof}

\end{document}